\documentclass[10pt,twocolumn,letterpaper]{article}

\usepackage{iccv}
\usepackage{times}
\usepackage{epsfig}
\usepackage{graphicx}
\usepackage{amsmath}
\usepackage{amssymb}

\usepackage{subfigure}

\usepackage{dsfont}	
\usepackage{footnote}
\usepackage{enumitem}

\usepackage{dirtytalk}
\usepackage{multirow}
\usepackage{boldline}
\usepackage{adjustbox}
\usepackage{balance}

\newcommand{\bfsection}[1]{\vspace*{0.1cm}\noindent\textbf{#1.}}

\usepackage[pagebackref=true,breaklinks=true,letterpaper=true,colorlinks,bookmarks=false]{hyperref}
\iccvfinalcopy 

\def\iccvPaperID{6374} 

\ificcvfinal\fi

\begin{document}


\title{DRB-GAN: A Dynamic ResBlock Generative Adversarial Network for\\ Artistic Style Transfer}

\author{Wenju Xu\textsuperscript{1}, 
Chengjiang Long\textsuperscript{2}, 
Ruisheng Wang\textsuperscript{3}, 
Guanghui Wang\textsuperscript{4}\thanks{This work was supervised by Chengjiang Long and Guanghui Wang.}\\
\textsuperscript{1}{OPPO US Research Center, InnoPeak Technology Inc, Palo Alto, CA, USA} \\
\textsuperscript{2}{JD Finance America Corporation, Mountain View, CA, USA} \\
\textsuperscript{3}{Department of Geomatics Engineering, University of Calgary, Alberta, Canada} \\
\textsuperscript{4}{Department of Computer Science, Ryerson University, Toronto, ON, Canada} \\
{\tt\small xuwenju123@gmail.com, cjfykx@gmail.com, ruiswang@ucalgary.ca, wangcs@ryerson.ca}
}

\maketitle
\ificcvfinal\thispagestyle{empty}\fi

\begin{abstract}
The paper proposes a Dynamic ResBlock Generative Adversarial Network (DRB-GAN) for artistic style transfer. The style code is modeled as the shared parameters for Dynamic ResBlocks connecting both the style encoding network and the style transfer network. In the style encoding network, a style class-aware attention mechanism is used to attend the style feature representation for generating the style codes. In the style transfer network, multiple Dynamic ResBlocks are designed to integrate the style code and the extracted CNN semantic feature and then feed into the spatial window Layer-Instance Normalization (SW-LIN) decoder, which enables high-quality synthetic images with artistic style transfer. Moreover, the style collection conditional discriminator is designed to equip our DRB-GAN model with abilities  for both arbitrary style transfer and collection style transfer during the training stage.
No matter for arbitrary style transfer or collection style transfer, extensive experiments strongly demonstrate that our proposed DRB-GAN outperforms state-of-the-art methods and exhibits its superior performance in terms of visual quality and efficiency. Our source
code is available at  \color{magenta}{\url{https://github.com/xuwenju123/DRB-GAN}}.
\end{abstract}

\section{Introduction}
\begin{figure}[t]
\centering
 \includegraphics[width=0.48\textwidth]{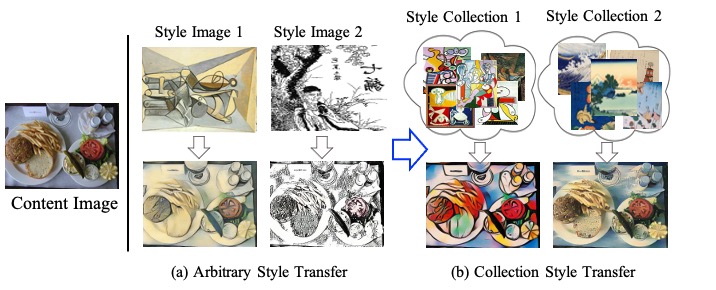}
\vspace{-0.4cm}     
\caption{Examples of two types of artistic style transfer: (a) arbitrary style transfer and (b) collection style transfer. Note that the style image 1 and style collection 1 are from the artist {\em Pablo Picasso}, and the style image 2 and style collection 2 are from the artist {\em Ukiyo-e}. Our proposed DRB-GAN experimentally performs well on both arbitrary style transfer and collection style transfer.}
 \label{fig:introduction}
 \vspace{-0.6cm} 
\end{figure}

Artistic style transfer is to synthesize an image sharing  structure similarity of the content image and reflecting the style of the artistic style. Here, artistic style implies the genre of paintings by the artist, and the artistic images refer to a set of images  created by the same artist, and each image has a unique character. As shown in Figure \ref{fig:introduction}, style image 1 and all the style images in style collection 1 are created by {\em Pablo Picasso}, in which  the style includes color, brushstroke, form, or use of light. Therefore, an ideal artistic style transfer should be able to synthesize images with consistent style genre and also take the diverse artworks of the artist into account. 

To facilitate more efficient artistic style transfer, some prior works have explored arbitrary style transfer \cite{huang2017arbitrary,jing2020dynamic}, which heavily rely on  only one arbitrary style image. Hence, they are not effective to produce a bunch of results that reflect the understanding of artistic style characterized by color, scale, and stroke size of the artistic work set. 
Some recent efforts on generative adversarial networks (GANs)~\cite{CycleGAN2017,sanakoyeu2018style,CycleGAN2017,sanakoyeu2018style,choi2018stargan,liu2019few} have succeeded in collection style transfer, which considers each style image in a style collection as a domain. However, the existing collection style transfer methods only recognize and transfer the domain dominant style clues and thus lack the flexibility of exploring style manifold. 

\begin{figure*}[t]
\vspace{-0.5cm}
\centering
\includegraphics[width=0.95\textwidth]{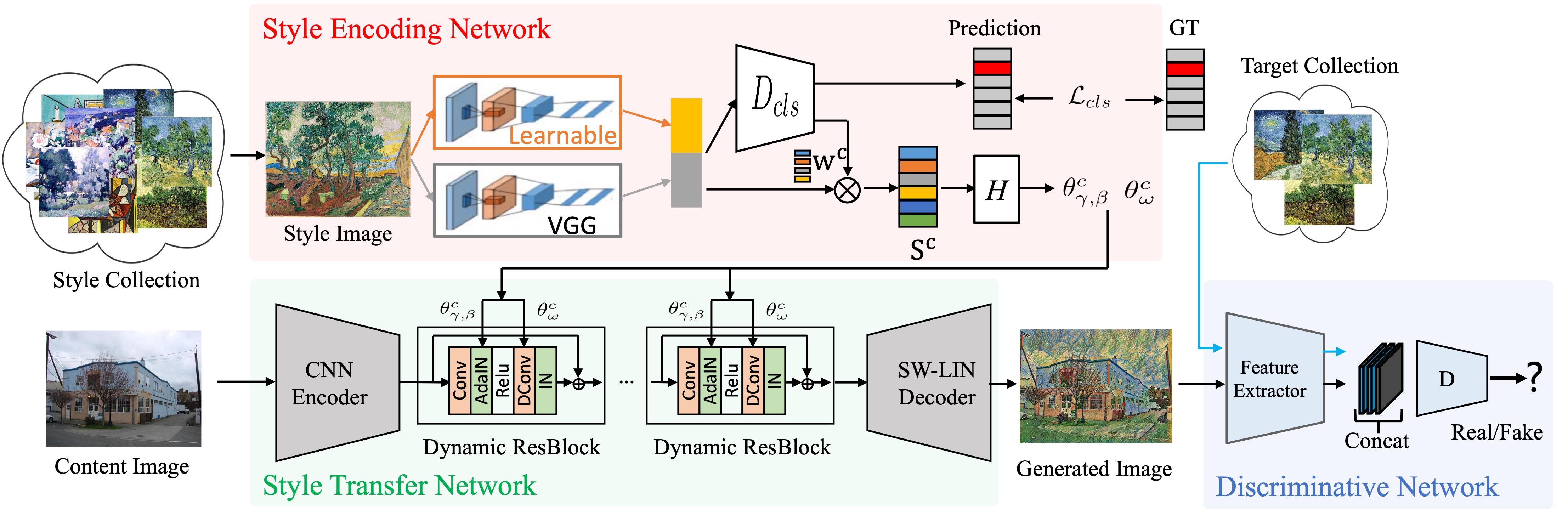}
\caption{An overview of the proposed DRB-GAN, which consists of a style encoding network, a style transfer network, and a discriminative network. The style code modeled as the shared parameter for dynamic ResBlocks is the output from the style encoding network, with a combination of a pre-trained VGG encoder and a learnable encoder as the feature extractor. 
A style class-aware attention mechanism is employed to recalibrate the style code. The final style code  is then fed into the style transfer network  designed as an encoder-decoder structure with multiple well-designed dynamic ResBlocks.
} 
\label{fig:pipeline}
 \vspace{-0.4cm} 
\end{figure*}

In this paper, we propose a Dynamic ResBlock Generative Adversarial Network (DRB-GAN) for artistic style transfer. As illustrated in Figure~\ref{fig:pipeline}, it consists of a style encoding network, a style transfer network, and a style collection discriminative network. In particular, inspired by the ideas of DIN~\cite{jing2020dynamic} and StyleGAN~\cite{karras2019style}, we model the ``{\em style code}" as the shared parameters for Dynamic Convolutions and AdaINs in dynamic ResBlocks, and design multiple Dynamic Residual Blocks (DRBs) at the bottleneck in the style transfer network. Note that each DRB consists of a Convolution Layer, a Dynamic Convolution~\cite{chen2020dynamic} layer, a ReLU layer, an AdaIN~\cite{huang2017arbitrary} layer, and an instance normalization layer with a residual connection. Such treatment is to attentively adjust the shared parameters for Dynamic Convolutions and adaptively adjust affine parameters for AdaINs to ensure the statistic matching in bottleneck feature spaces between content images and style images.


We incorporate a fixed pretrained VGG encoder~\cite{simonyan2014very} and a learnable encoder as feature extractor in the style encoding network to capture style class-aware feature representation. The output style class-aware probabilities can be used as attention weights to attend to the style features for style code recalibration. As the style class-aware attention mechanism is learned via images in style collections, the output ``{\em style code}" captures the underlying discriminative information in the style image collections. Note that our attention mechanism enforces a better clustering of the ``{\em style codes}", which is different from previous class activation mapping based methods \cite{kim2019u,zhou2016learning} that aim at highlighting spatial regions.



With the ``{\em style code}" from the style encoding network, multiple DRBs can adaptively proceed the semantic features extracted from the CNN encoder in the style transfer network then feed them into the spatial window Layer-Instance Normalization (SW-LIN) decoder to generate synthetic images. Specially, we borrow the idea of local feature normalization from~\cite{kotovenko2019content} and design an SW-LIN function that dynamically combines the local channel-wise and layer-wise normalization with a learnable parameter in each decoder block. With the spatial window constraint, our SW-LIN is able to flexibly shift the mean and variance in the feature spaces. As a consequence, our SW-LIN decoder can avoid the possible artifacts and retain the capability to synthesize high-resolution stylization.

 

As the ``{\em style code}" captures the underlying discriminative information in the style encoding network, it is easy to apply the learned style network on each style image in a style collection and get a set of style codes. We can apply the weighted average on the style codes to obtain the ``{\em collection style code}" and then feed it into the style transfer network to conduct the collection style transfer. 
Moreover, our discriminative network takes several style images sampled from the target style collection of the same artist as references to ensure consistency in the feature space. Together with the perception supervision, our well-designed discriminator provides good guidance for our DRB-GAN to own abilities for both arbitrary style transfer and collection style transfer gradually, shrinking their gap at the training stage.  
With extensive experiments, we have demonstrated the effectiveness of our proposed DRB-GAN on both arbitrary style transfer and collection style transfer.

Several aspects distinguish our work from previous style transfer models \cite{sanakoyeu2018style,kotovenko2019content,svobodatwo}. First of all, our DRB-GAN introduces a novel prototype for artistic style transfer, in which ``{\em style code}" is modeled as the shared parameters for Dynamic ResBlocks, connecting both the style encoding network and the style transfer network, to shrink the gap between arbitrary style transfer and collection style transfer in a unified model. Second, we introduce a style class-aware attention mechanism for style code recalibration and then employ well-designed multiple dynamic ResBlocks to integrate the style code and the extracted semantic feature to realize artistic style transfer when generating high-quality synthetic images. Last but not least, the discriminative network makes full use of style images sampled from the target collection as a reference which enforces our DRB-GAN's ability for collection style transfer. Together with perception supervision, the ability for arbitrary style transfer can be well preserved and improved at the training stage. 

Both quantitative and qualitative experiments demonstrate the effectiveness and efficiency of the proposed DRB-GAN, as well as its superior performance in artistic style transfer, regardless of  arbitrary or collection style transfer.

\section{Related Work}
{\noindent\bf{Generative Adversarial Networks (GANs)}}~\cite{goodfellow2014generative} have been successfully applied to visual recognition~\cite{Hu:TIP2021, Long:ICCV2015, Hua:ICCV2013B, Long:IJCV2016, Long:CVPR2017, Hua:TPAMI2018}, object detection~\cite{Islam:CVPR2020}, image generation~\cite{tolstikhin2017wasserstein, gulrajani2017improved,karras2019style,xu2019,XU2019570}, image translation~\cite{zhu2017toward,larsen2016autoencoding,kim2019u}, shadow removal~\cite{Ding:ICCV2019, Wei:CGF2019, Zhang:AAAI2020, Zhang:CGF2020}, image captioning~\cite{chen2019improving, Dong:MM2021}, {\em etc}. 
These GAN models are trained to minimize the discrepancy between distributions of the training data and unobserved generations. Our GAN model is designed with a special discriminator that judges the generated images by taking similar images from the target collection as a reference.

\bfsection{Arbitrary style transfer}  Gatys \etal \cite{gatys2016image} for the first time takes a pre-trained neural network to optimize synthesized images. However, this method inefficiently searches for a numerical solution in pixel space. To address this, recent approaches rely on a learnable neural network to match the statistical information in feature space. The earliest Per-Style-Per-Model (PSPM) algorithms train a single model for one particular style image\cite{johnson2016perceptual}. While the Multiple-Style-Per-Model (MSPM) algorithms are proposed \cite{zhang2017multi,xu2018learning,chen2017stylebank} to use one model for multiple style images. For instance, the StyleBank \cite{chen2017stylebank} uses one single model to incorporate feature representations of multiple style images. Recently,  the Arbitrary-Style-Per-Model (ASPM) algorithms \cite{li2018closed,GDWCT2019,li2019learning} are proposed to transfer arbitrary new styles in one unified model. MetaNet \cite{shen2018neural} introduces a parameter network to generate network parameters based on the style image. AdaIN \cite{huang2017arbitrary} and DIN \cite{jing2020dynamic} methods employ conditional instance normalization to dynamically generate the affine parameter in the instance normalization layer. The CST \cite{svobodatwo} introduces a unified model for conditional style transfer. In a different manner, Li \etal \cite{li2017universal} utilizes the whitening and coloring transforms (WCT) on the style features. However, these style transfer algorithms consider one individual style image as one style. This assumption ignores the concept of the artistic collection of an artist. Instead, our model takes novel Dynamic ResBlocks to efficiently deal with the artistic style transfer task.

\bfsection{Collection style transfer} The collection style transfer methods \cite{dumoulin2016adversarially,choi2018stargan,liu2018unified} work on image collections or domains. They are built upon GANs to map inputs into a different domain. We refer to this type of method as a Per-Domain-Per-Model (PDPM) algorithm. For instance, CycleGAN \cite{CycleGAN2017} takes one generator to translate images into another domain and uses another generator to translate them back for cyclic consistency. The AST \cite{sanakoyeu2018style} extends the GAN-based model for high-resolution artistic style transfer. The CSD \cite{kotovenko2019content} introduces a content transformation block to preserve the content structure in the synthetic images. Recent work \cite{huang2018multimodal,liu2019few} propose to handle the multiple domain translation task. In
contrast, our proposed method models the ``{\em collection style code}" in a dynamic way to facilitate efficiency in handling multiple domains. We, therefore, refer to our mother as a Multiple-Domain-Per-Model (MDPM) method.


\section{Proposed Approach}\label{problem}

As illustrated in Figure \ref{fig:pipeline}, our DRB-GAN consists of three networks, {\em i.e.}, a style encoding network and a style transfer network to formulate the image generator $G$, a discriminative network as the discriminator $D$ to ensure the generated images with desired style consistent with style images in the collection. 
Let us use subscripts $c$ to indicate the $c$-th style. Given a content image $x \in X$ and an arbitrary style image $y^c \in Y$ randomly sampled from $N$ different style image collections, our goal is to transfer the content image with the generator $G$ to produce a desired synthetic image $\tilde{x}^c$,  ensuring a consistent style with the style image $y^c$ via the discriminator $D$.

\subsection{Style Encoding Network with Style Class-Aware Attention}
As shown in Figure~\ref{fig:pipeline}, we model the style code as the shared hyper-parameters 
for Dynamic ResBlocks which is designed to integrate dynamic convolution (DConv) and Adaptive Instance Normalization Normalization (AdaIN)~\cite{huang2017arbitrary} in a residual structure~\cite{he2016deep}. The style encoding network is to generate style code from the style image for the style transfer network on the content image. 

 \bfsection{Attention guided feature extractor}
We introduce an architecture of style encoding by concatenating the features from a pre-trained VGG encoder and a learnable encoder. The parameters in the learnable encoder are updated while those in the VGG encoder are fixed. Since the fixed VGG network is pre-trained on the COCO dataset \cite{lin2014microsoft}, it has seen many images with various textures, thereby it has a global property and strong generalization ability for in-the-wild textures. Considering the gap between the COCO dataset and others, it is difficult for the network with a fixed encoder to fit such a complex model. Therefore we introduce a learnable encoder as complementary to the fixed VGG encoder to extract the subtle variations in style. 

Inspired by the class activation mapping (CAM) \cite{zhou2016learning}, we take a classification weight to recalibrate our encoded style feature, denoted as $F_s$. The attention mechanism is based on an auxiliary classifier $D_{cls}$ trained to predict the style classification probability $w^c$, which is used to the likelihood of the input style image belonging to the $c_{th}$ category. Then, the style encoding is recalibrated as
\begin{equation}
    s^c = \omega^c F_s, \label{eqn:sc}
\end{equation}
The recalibrated style encoded feature is then fed into the weight generation module $H$ designed as multi-layer perceptions (MLPs) to determine parameter values for Dynamic ResBlocks. 

\bfsection{Style code generation for arbitrary style transfer} 
Given the recalibrated style encoding, this module is to generate the parameters as ``{\em style code}" in dynamic ResBlocks, which can be written as:
\begin{align}
\{\theta_{\omega}^c, \theta_{\gamma,\beta}^c\} =\{H_{\omega}(s^c),H_{\gamma,\beta}(s^c)\}, 
\end{align}
\noindent where $s^c$ is the style feature, $H_{\omega}(\cdot)$ is a MLP used to generate filter weights $\theta_{\omega}^c$ for Dynamic Convolution~\cite{chen2020dynamic} Layers. Another MLP $H_{\gamma,\beta}(\cdot)$ creates the affine parameters $\theta_{\gamma,\beta}^c$ in the AdaIN~\cite{huang2017arbitrary} layers.


{\noindent\bf Weighted averaging strategy for collection style transfer}.
We introduce a weighted averaging strategy to extend arbitrary style encoding for collection style transfer. Specifically, we calculate the ``{\em collection style code}" as a weighted  mean of the ``{\em style codes}" 
based on several representative paintings of the same artist and the corresponding weights ${\bf \pi}$. For the $k$-th style image in the collection, the weight $\pi_k$ is determined by the similarity between the style image and the query content image. Therefore, we can formulate the ``{\em collection style code}" as
\begin{align}
\label{eqx}
\begin{split}
\{\bar{\theta}_{\omega}^c,~ \bar{\theta}_{\gamma,\beta}^c\} = \{\frac{1}{K}\sum_{k=0}^{K} \pi_k \theta_{\omega_k}^{c},~ \frac{1}{K}\sum_{k=0}^{K} \pi_k \theta_{\gamma_k,\beta_k}^{c} |c \sim N\},
\nonumber
\end{split}
\end{align}
\noindent where $K$ is the number of style images used to calculate the mean of the generated weights at test stage, and $c$ indicates the target style domain. Our experimental results show that our weighted averaging strategy produces impressive results on collection style transfer task. 

\subsection{Style Transfer Network with Dynamic ResBlocks}
Our style transfer network contains a CNN Encoder to down-sample the input, multiple dynamic residual blocks, and a spatial window Layer-Instance Normalization (SW-LIN) decoder to up-sample the output. 
With different ``{\em style code}" parameters, the style transfer network converts the content image into different styles. 

\bfsection{Dynamic ResBlock}
As the core of the transfer network, each dynamic ResBlock is composed of  a Convolution Layer, a Dynamic Convolution~\cite{chen2020dynamic} layer, a ReLU layer, an AdaIN~\cite{huang2017arbitrary} layer, and an instance normalization layer with a residual structure. 
It is designed to integrate the advantages of both dynamic convolution and adaptive instance normalization layer in the residual block. Note that all of the parameters in layers are dynamically generated from the style encoding network. 

\bfsection{SW-LIN Decoder}
The GAN model tends to produce artifacts in the generated samples \cite{Karras_2020_CVPR}, which significantly degrades its applications. Many researchers attempt to replace the instance normalization function with the layer normalization function in the decoder modules to remove the artifacts. After studying these normalization operations, we observe that instance normalization normalizes each feature map separately, thereby potentially destroying any information found in the magnitudes of the features relative to each other. While layer normalization operation normalizes the feature map together, thereby potentially demolishing each feature map as the representation of the style. Therefore, we equip the decoder blocks with spatial window Layer-Instance Normalization (SW-LIN) function which dynamically combine these normalization functions with learnable parameter $\rho$:
\begin{align}
\text{SW-LIN}(\gamma,\beta,\rho) = \gamma (\rho \phi_{sw}^c +(1-\rho) \phi_{sw}^l) + \beta ,
\end{align}
\noindent where $\gamma, \beta$ are learnable parameters, and $\phi_{sw}^c,\phi_{sw}^l$ are channel-wise, layer-wise normalized features, respectively. Specially, the statistic mean and variance are obtained across a window of spatial location instead of the whole input tensor ${\bf h}$.  Formally, we get
\begin{align}
\phi_{sw} = \frac{{\bf h}-E_{x_i \in sw}[{\bf h}(x_i)]}{\sqrt{Var_{x_i \in sw}[{\bf h}(x_i)]}} .
\end{align}

The $\text{SW-LIN}$ function helps our decoder to flexibly normalize the features. As a result, without modifying the model architecture or hyper-parameters, our SW-LIN decoder can remove the artifacts and retain the capability to synthesize high-resolution stylization. 




\subsection{Discriminative Network for Arbitrary and Collection Style Transfer} 
Simply answering a real or fake question is not enough to provide correct supervision to the generator which aims at both individual style and collection style. To address this issue, we introduce a novel conditional discriminator with a collection of style images, which encourages the generated images to maintain the texture in any style genre. 

As illustrated in Figure~\ref{fig:pipeline}, our collection discriminator takes the generated images and several style images sampled from the target style collection as input. The feature extraction part generates a feature map for each image and we concatenate them channel-wisely. Then a small network with three Convolution layers is used to assess the quality based on the concatenated feature map.
Unlike the discriminator of conditional GANs taking a category label as an additional input besides the generated image, our discriminator takes a collection of style images as a reference instead to ensure the style consistency at the feature space. 


Working together with the perceptual supervision (see Equation~\ref{eqn:perception_loss}) between a selected style image and the corresponding generated stylization image, our discriminator provides good guidance to train the generator for both arbitrary and collection style transfer. Consequently, the gap between the arbitrary and collection style transfer has been shrunk smoothly at the training stage.

\subsection{Objective Functions}
The objective loss function $\mathcal{L}$ is formulated with 
adversarial loss $\mathcal{L}_{adv}$, perceptual loss $\mathcal{L}_{per}$, and style classification loss $\mathcal{L}_{cls}$ as flollows, 
\begin{align}
\begin{aligned}
\mathcal{L} = \mathcal{L}_{adv}  + \lambda_{per} \mathcal{L}_{per} + \lambda_{cls} \mathcal{L}_{cls}
\end{aligned}
\label{eqn:overall_loss}
\end{align}
where $\lambda_{per}$ and $\lambda_{cls}$ are the weight parameters.

{\noindent\bf Adversarial loss} $\mathcal{L}_{adv}$ is designed to distinguish between the patches from synthesized images and those from a group of style images belonging to the same style collection, {\em i.e.},
\begin{align}
\begin{aligned}
&\mathcal{L}_{adv}=E_{y^c,y_i^c \sim Y, c\sim N}[-\log D(y^c,\{y_i^c\}_{i=0}^M)]\\&+E_{\tilde{x}^c \sim G(x), y_j^c \sim Y, c\sim N}[-\log(1-D(\tilde{x}^c,\{y_j^c\}_{j=0}^M))],
\end{aligned}
\end{align}
where $M$ indicates the number of style images used by the collection discriminator at each iteration. in our experiment, we find setting $M=2$ is enough to obtain decent performance. 

{\noindent\bf Perceptual loss} $\mathcal{L}_{per}$ is employed to compute style losses at multiple levels and content loss between the style image and the generated stylization image with a pretrained VGG network in a way similar to the prior work~\cite{li2017demystifying,fu2019edit}, {\em i.e.},
\begin{equation}
    \mathcal{L}_{per} =  \lambda_{c}\mathcal{L}_c +  \lambda_{s}\mathcal{L}_s,
    \label{eqn:perception_loss}
\end{equation} 
where the style loss is calculated by matching the mean and standard deviation of the style features:
\begin{align}
\label{eqx}
\begin{split}
\mathcal{L}_{s} = \mathbb{E}_{l\sim N_l} ((\mu_{y^c}^{l} - \mu_{\tilde{x}^c}^{l} )^2 + (Gram_{y^c}^{l} - Gram_{\tilde{x}^c}^{l})^2 ),
\end{split}
\nonumber
\end{align}
\noindent where $N_L$ is the number of involved layers (in this work, we use the $Relu_{12}$, $Relu_{22}$, $Relu_{33}$, $Relu_{43}$ and $Relu_{51}$ layers in the VGG network and that of feature maps in the $l_{th}$ layer). In addition, $\mu$ and is the mean and Gram represents the Gram matrix of the corresponding feature map.
We take advantage of the content loss to preserve the structural similarity between the content image and the synthesized image. The content loss is the Euclidean distance between the target features and the features of the output image.
\begin{equation}
    \mathcal{L}_{c} = E_{x\sim X,c\sim N}||\phi(x)-\phi(\tilde{x}^c)||,
    \label{eqx}
\end{equation}
where $\phi$ represents the features of $Relu_{41}$ layers. At the training stage, the style image is randomly sampled with the target domain label. We then transfer the content image to the target style domain so as to learn all the mappings across multiple artistic style domains. 

{\noindent\bf Style classification loss} $\mathcal{L}_{cls}$ is exploited for the auxiliary classification $D_{cls}$ in the style encoding network to ensure that the style class prediction is correct.
\begin{equation}
    \mathcal{L}_{cls} = E_{y \sim Y, c\sim N}(-\log D_{cls}(c|y^c)).
\end{equation}

\section{Experiment}

\bfsection{Implementation details} Our model is implemented by PyTorch \cite{paszke2017automatic}. It is trained on 624,777 content images from the Place365 \cite{zhou2014learning} dataset, and eleven artistic painting collections from WikiArt \cite{karayev2013recognizing} database. We adopt Adam \cite{kingma2014adam} as the optimization solver. The learning rate is set to 0.0001. We train our model for $600,000$ iterations with a batch size of 1. The training images are augmented by random rotation and horizontal flipping, and then resized and randomly cropped to $768\times768$ resolution. Note that
during testing, our model is able to work on images of arbitrary size. We empirically set $\lambda_c = 1$, $\lambda_s = 0.02$, $\lambda_{cls} = 1$, and $\lambda_{vgg} = 1$.

\bfsection{Baseline methods}
The goal of our DRB-GAN model is to generate high-resolution artistic stylizations.
We compare our DRB-GAN with state-of-the-art methods, {\em i.e.}, instance style transfer method like Gatys \etal~\cite{gatys2016image}, arbitrary style transfer methods including AdaIN \cite{huang2017arbitrary}, CST \cite{svobodatwo} and MetaNet\cite{shen2018neural}, and collection style transfer methods like AST \cite{sanakoyeu2018style} and CycleGAN \cite{CycleGAN2017}). For a fair comparison, we deploy all competing methods for style transfer on images of size $768\times 768$, unless otherwise specified. Note that we use the public released code by authors and train the models on the same training data.


\bfsection{Evaluation metrics} We use deception rate to evaluate how well the target style characteristics are transferred to generated images. The deception rate is calculated as the percentage predicted by a pretrained artist classification network for the correct artist. 
We also conduct human perceptual study to assess the quality of the stylization results in terms of content preservation and style consistency to the target style.  

\begin{figure*}[ht!]
\vspace{-0.5cm}
\centering
 \includegraphics[height=0.078\textwidth,width=0.095\textwidth]{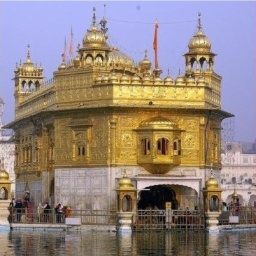}
  \hspace{-4pt}
  \includegraphics[height=0.078\textwidth,width=0.095\textwidth]{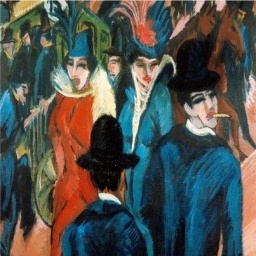}
  \hspace{-4pt}
  \includegraphics[height=0.078\textwidth,width=0.095\textwidth]{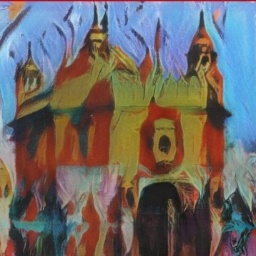} 
  \hspace{-4pt}
  \includegraphics[height=0.078\textwidth,width=0.095\textwidth]{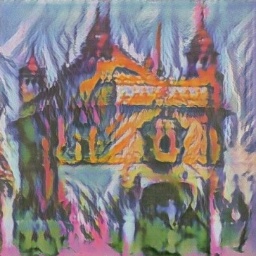}  
 \hspace{-4pt}
  \includegraphics[height=0.078\textwidth,width=0.095\textwidth]{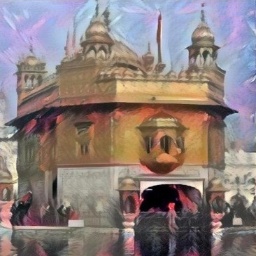}
  \hspace{-4pt}
  \includegraphics[height=0.078\textwidth,width=0.095\textwidth]{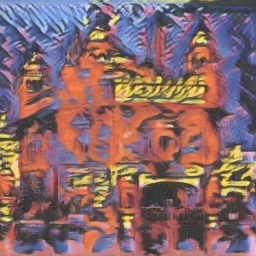}
  \hspace{-4pt}  
  \includegraphics[height=0.078\textwidth,width=0.095\textwidth]{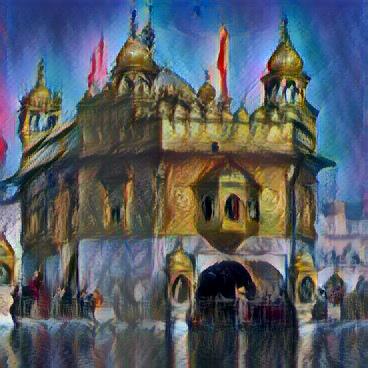}  
  \hspace{-4pt}
  \includegraphics[height=0.078\textwidth,width=0.095\textwidth]{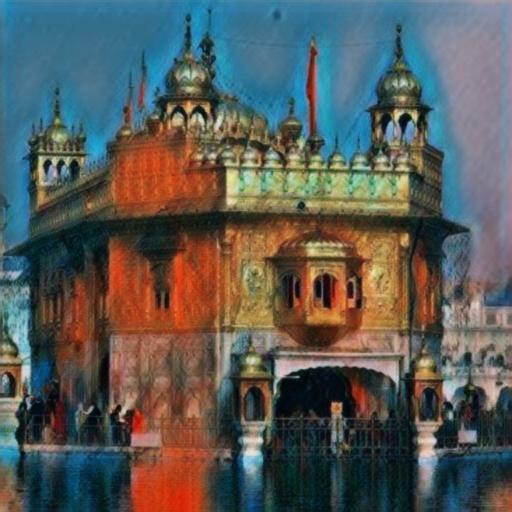}  
  \hspace{-4pt}
  \includegraphics[height=0.078\textwidth,width=0.095\textwidth]{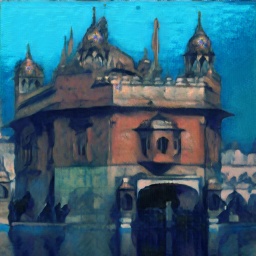}  
  \hspace{-4pt}
  \includegraphics[height=0.078\textwidth,width=0.095\textwidth]{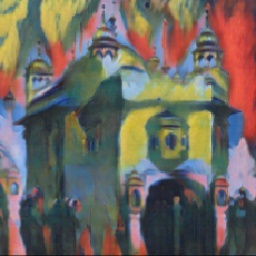} \\
 \includegraphics[height=0.078\textwidth,width=0.095\textwidth]{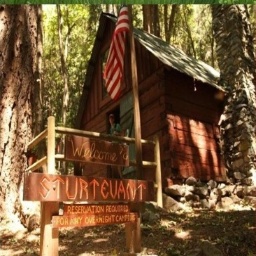}
  \hspace{-4pt}
  \includegraphics[height=0.078\textwidth,width=0.095\textwidth]{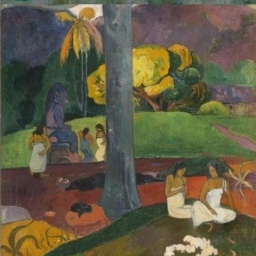}
  \hspace{-4pt}
  \includegraphics[height=0.078\textwidth,width=0.095\textwidth]{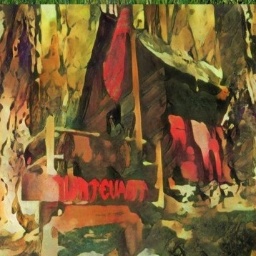} 
  \hspace{-4pt}
  \includegraphics[height=0.078\textwidth,width=0.095\textwidth]{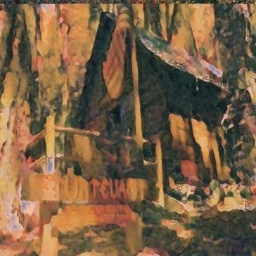}  
 \hspace{-4pt}
  \includegraphics[height=0.078\textwidth,width=0.095\textwidth]{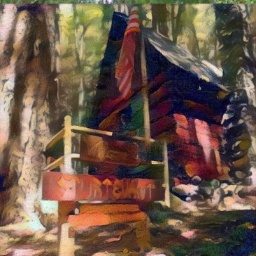}
  \hspace{-4pt}
  \includegraphics[height=0.078\textwidth,width=0.095\textwidth]{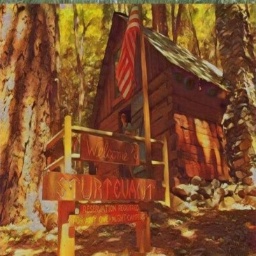}
  \hspace{-4pt}   
  \includegraphics[height=0.078\textwidth,width=0.095\textwidth]{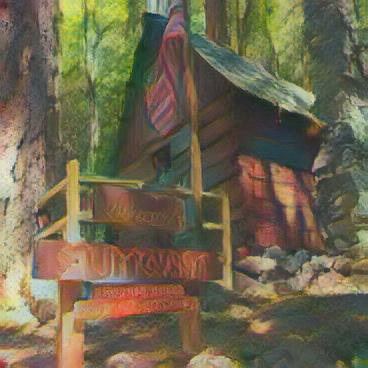}  
  \hspace{-4pt}
  \includegraphics[height=0.078\textwidth,width=0.095\textwidth]{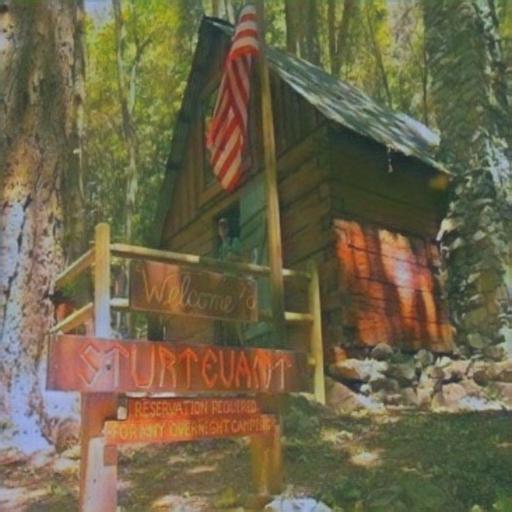}  
  \hspace{-4pt}
  \includegraphics[height=0.078\textwidth,width=0.095\textwidth]{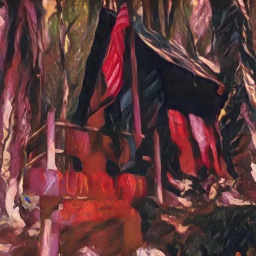} 
  \hspace{-4pt}
  \includegraphics[height=0.078\textwidth,width=0.095\textwidth]{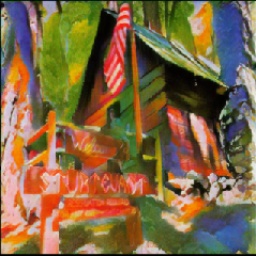} \\
 \vspace{-5.pt}
 \subfigure[]{\includegraphics[height=0.078\textwidth,width=0.095\textwidth]{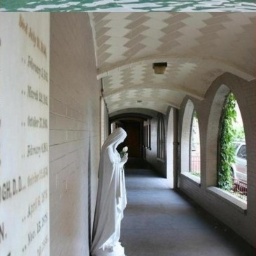}} 
  \hspace{-4pt}
  \subfigure[]{\includegraphics[height=0.078\textwidth,width=0.095\textwidth]{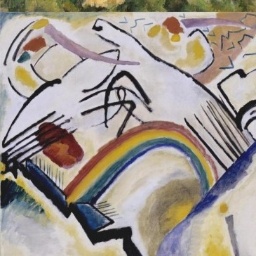}}
  \hspace{-4pt}  
  \subfigure[]{\includegraphics[height=0.078\textwidth,width=0.095\textwidth]{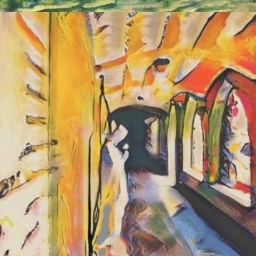}} 
  \hspace{-4pt}  
  \subfigure[]{\includegraphics[height=0.078\textwidth,width=0.095\textwidth]{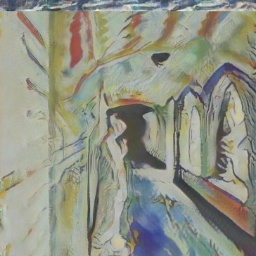}} 
  \hspace{-4pt}
  \subfigure[]{\includegraphics[height=0.078\textwidth,width=0.095\textwidth]{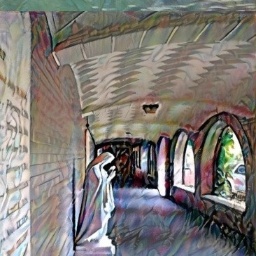}} 
  \hspace{-4pt}  
  \subfigure[]{\includegraphics[height=0.078\textwidth,width=0.095\textwidth]{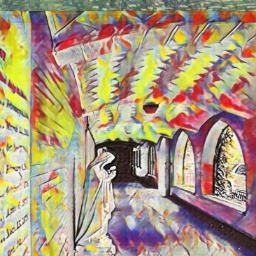}}
  \hspace{-4pt}  
  \subfigure[]{\includegraphics[height=0.078\textwidth,width=0.095\textwidth]{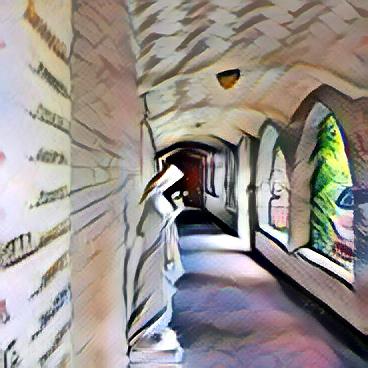}}
  \hspace{-4pt}  
  \subfigure[]{\includegraphics[height=0.078\textwidth,width=0.095\textwidth]{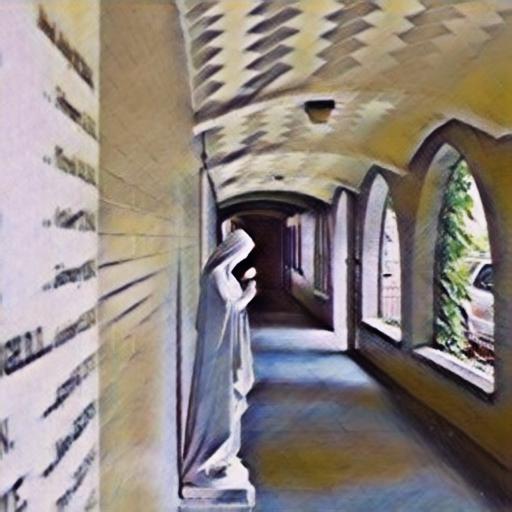}}
  \hspace{-4pt}  
  \subfigure[]{\includegraphics[height=0.078\textwidth,width=0.095\textwidth]{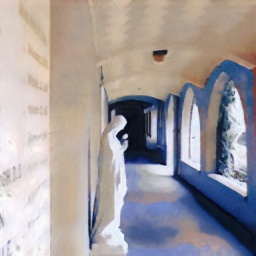}}
  \hspace{-4pt}  
  \subfigure[]{\includegraphics[height=0.078\textwidth,width=0.095\textwidth]{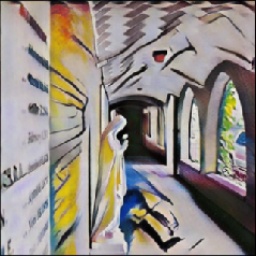}}  \\  
  \vspace{-0.1cm}
\caption{Performance comparison on stylized results from different models. From left to right are (a) content image, (b) style image, the results of (c) CSD, (d) AST, (e) Gatys, (f) CycleGAN, (g) AdaIN, (h) MetaNet, (i) CST, and (j) DRB-GAN, respectively. 
}
\label{fig:comp_arbitrary_style_transfer}
\vspace{-0.4cm} 
\end{figure*}


\subsection{Arbitrary Style Transfer}


\subsubsection{Qualitative Evaluation}
\bfsection{Image stylization}
We present qualitative results of different style transfer methods in Figure~\ref{fig:comp_arbitrary_style_transfer} for comparison. All the results are obtained on images of high resolution. The comparison shows the outperformance of our DRB-GAN in terms of visual quality. These images in Figure~\ref{fig:comp_arbitrary_style_transfer} (j) contain no artifacts in the regions, and most importantly, they preserve the structural similarity of the content images. On the contrary, the algorithms AdaIN and MetaNet fail to generate sharp details and fine strokes. We also observe non-negligible artificial structures in those images obtained by AdaIN, Gatys, and CycleGAN. And the models CSD  \cite{kotovenko2019content} and AST \cite{sanakoyeu2018style} fail to transfer the Kandinsky style (third row) to the content image. 


\begin{figure}[ht!]
\centering
  \includegraphics[height=0.06\textwidth,width=0.08\textwidth]{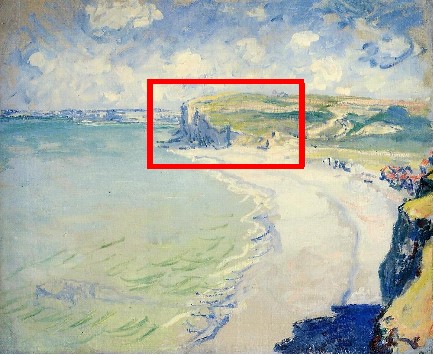} 
  \hspace{-4pt}
  \includegraphics[height=0.06\textwidth,width=0.076\textwidth]{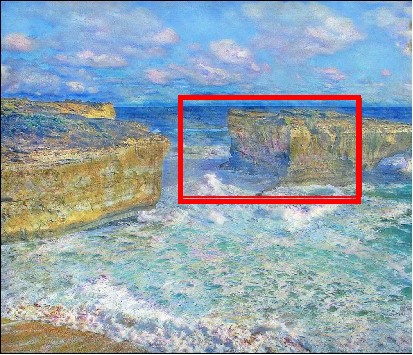} 
  \hspace{-4pt}  
  \includegraphics[height=0.06\textwidth,width=0.076\textwidth]{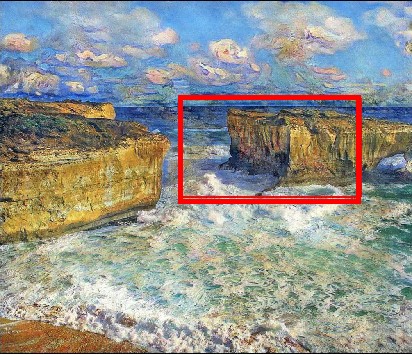}  
 \hspace{-4pt}
  \includegraphics[height=0.06\textwidth,width=0.076\textwidth]{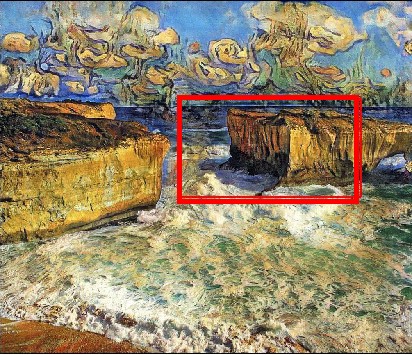}
 \hspace{-4pt}  
  \includegraphics[height=0.06\textwidth,width=0.076\textwidth]{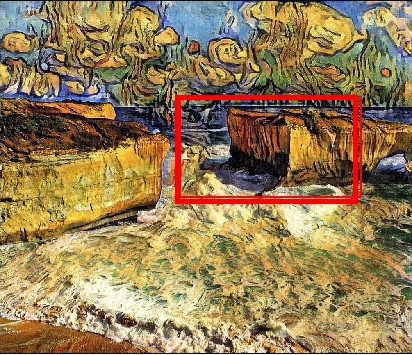}
 \hspace{-4pt}  
  \includegraphics[height=0.06\textwidth,width=0.08\textwidth]{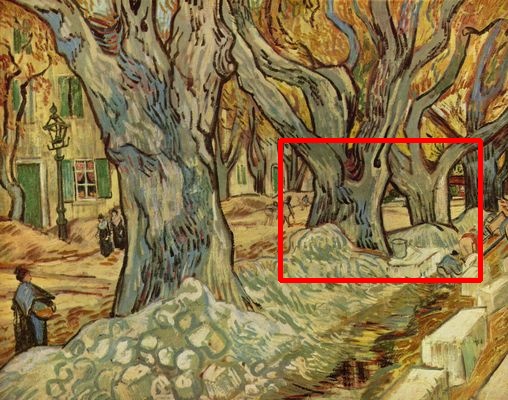}  \\
  \includegraphics[height=0.04\textwidth,width=0.08\textwidth]{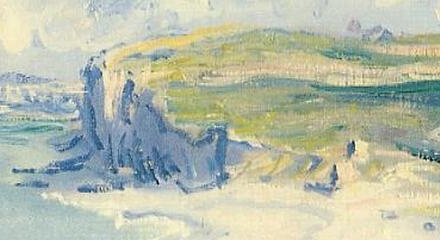} 
  \hspace{-4pt}
  \includegraphics[height=0.04\textwidth,width=0.076\textwidth]{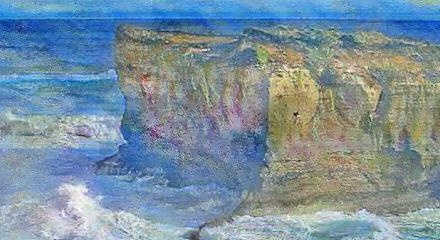} 
  \hspace{-4pt}  
  \includegraphics[height=0.04\textwidth,width=0.076\textwidth]{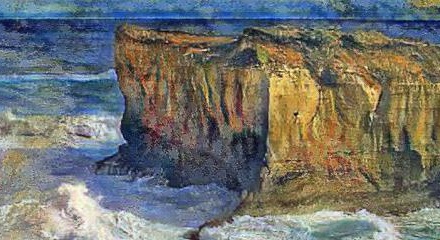}  
 \hspace{-4pt}
  \includegraphics[height=0.04\textwidth,width=0.076\textwidth]{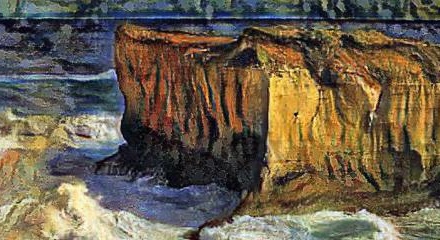}  
 \hspace{-4pt}
  \includegraphics[height=0.04\textwidth,width=0.076\textwidth]{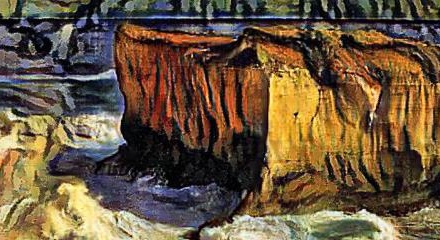}  
 \hspace{-4pt} 
  \includegraphics[height=0.04\textwidth,width=0.08\textwidth]{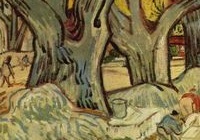}  \\
\caption{Interpolation results between given style samples of Monet (first column) and van Gogh (last column ). Magnified regions (row 2) show that our method mimics not only colors but also contours and textures specific to the style.}
\label{interpolation}
\vspace{-0.4cm} 
\end{figure}



\bfsection{Style interpolation} To make smooth transitions between different style images, we operate linear interpolation on dynamically generated weights. The interpolated weights $\{\tilde{\theta}_{\omega},~ \tilde{\theta}_{\gamma,\beta}\}$ are obtained as
\begin{align}
\label{eqx}
\begin{split}
\{\tilde{\theta}_{\omega},~ \tilde{\theta}_{\gamma,\beta}\} &= \{\alpha \theta_{\omega}^{c_1} + (1-\alpha) \theta_{\omega}^{c_2} ,~ \alpha \theta_{\gamma,\beta}^{c_1} \\&+ (1-\alpha) \theta_{\gamma,\beta}^{c_2}|c_1,c_2 \sim N\}
\end{split}
\end{align}
\noindent where $\alpha$ is the interpolation factor between 0 and 1, $c_1$ and $c_2$ represent style domains. Figure \ref{interpolation} demonstrates a smooth transition between Monet and van Gogh style with magnified details. Our model captures subtle variations between these two styles. 


\subsubsection{Quantitative Evaluation}

\bfsection{Style transfer deception rate}
To quantitatively measure the performances of different models, we take the deception rate metric used by AST \cite{sanakoyeu2018style}. The deception rate is the correct rate of stylized images that were recognized by a pretrained network as the target styles. We take the same way used in CSD \cite{kotovenko2019content2} to measure the style transfer deception rate and report the mean deception rate in Table \ref{fig:score}. As we can see, our method DRB-GAN achieves 0.573, which significantly outperforms other baseline methods. As a comparison, the mean accuracy of the network on real images of the artists from Wikiart is 0.626.
\setlength{\tabcolsep}{2.5pt}
\begin{table}[ht!]
\small 
\centering
\caption{\ Quantitative comparison with state-of-the-art methods. Average inference time and GPU memory consumption, measured
on a Titan XP GPU, for different methods with a batch size of 1 and an input image of $768 \times 768$. The column ``model" is for the category of the style transfer method. 
For the content and style score, higher values indicate better performance. Scores are averaged over ten different styles.}\label{fig:score}
\vspace{2mm}
    \begin{tabular}{c||ccc| c c c}
	\hlineB{2} 
	 &      & GPU  &    & & \multicolumn{2}{c}{Human studies}    \\
	\cline{6-7} 
	Method&  Time &memory & Model & Deception&Content&Style \\
	& (sec) &(MiB) &  &rate &score&score\\
	\hline \hline
	Wikiart test&&&&0.626&-& -\\			       
	\hline
	Gatys \etal  &200&3887&PSPM &0.251&67.1\%&0.127 \\
    AdaIN &0.16&8872&ASPM &0.061&43.6\%&0.019\\
    WCT & 5.22&10720&ASPM &0.023&39.2\%&0.013\\
    PatchBased &8.70&4159 &ASPM&0.063&53.4\%&0.043\\
    Johnson &0.06 &671&ASPM&0.080&38.5\%&0.021\\
    CycleGAN &0.07&1391 & PDPM&0.130&43.2\%&0.012\\
    AST &0.07&1043 & PDPM &0.450&63.9\%&0.312\\
    DRB-GAN&0.08&1324& MDPM&{\bf 0.573}& {\bf 72.2\%}&{\bf 0.453}  \\
    \hlineB{2} 
	\end{tabular}
 \vspace{-1cm} 	
\end{table}

\begin{figure*}[ht!]
\vspace{-0.75cm}
\centering
 \includegraphics[height=0.078\textwidth,width=0.103\textwidth]{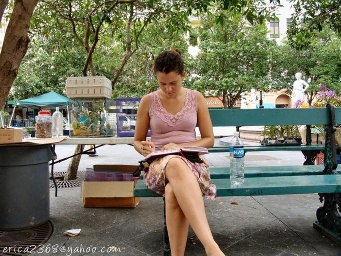}
  \hspace{-4pt}
  \includegraphics[height=0.078\textwidth,width=0.068\textwidth]{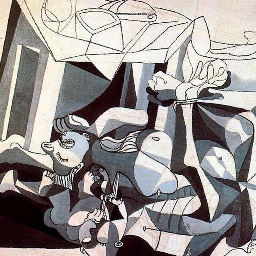}
  \hspace{-4pt}  
  \includegraphics[height=0.078\textwidth,width=0.103\textwidth]{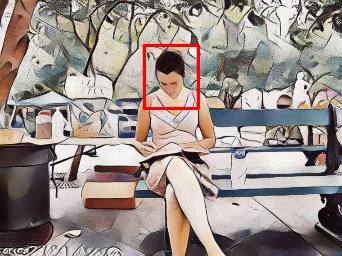}  
  \hspace{-4pt}
  \includegraphics[height=0.078\textwidth,width=0.103\textwidth]{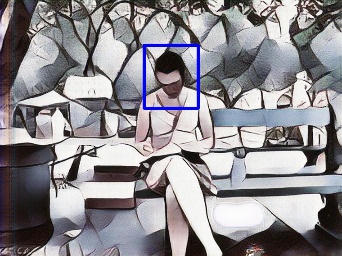} 
  \hspace{-4pt}
  \includegraphics[height=0.078\textwidth,width=0.103\textwidth]{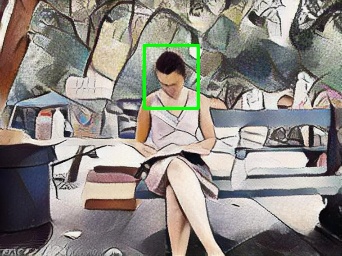}  
 \hspace{-4pt}
  \includegraphics[height=0.078\textwidth,width=0.103\textwidth]{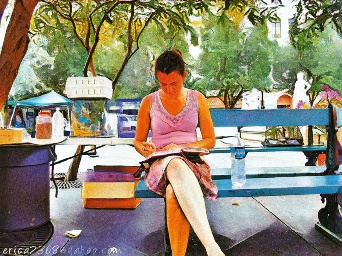}
  \hspace{-4pt}
  \includegraphics[height=0.078\textwidth,width=0.103\textwidth]{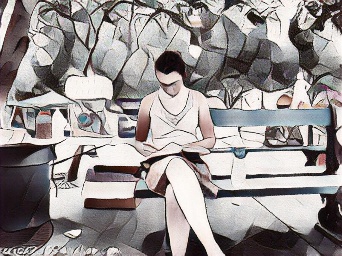}
  \hspace{-4pt}  
  \includegraphics[height=0.078\textwidth,width=0.103\textwidth]{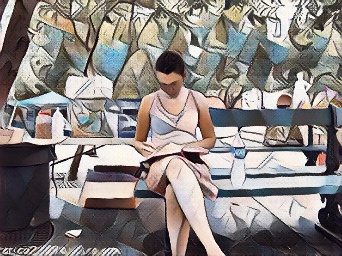} 
  \hspace{-4pt}
  \includegraphics[height=0.078\textwidth,width=0.103\textwidth]{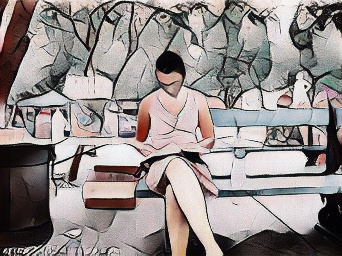} 
  \hspace{-4pt}  
  \includegraphics[height=0.078\textwidth,width=0.078\textwidth]{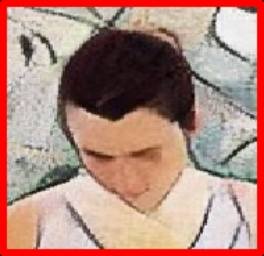} \\
 \includegraphics[height=0.078\textwidth,width=0.103\textwidth]{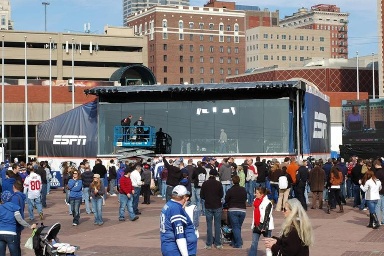}
  \hspace{-4pt}
  \includegraphics[height=0.078\textwidth,width=0.068\textwidth]{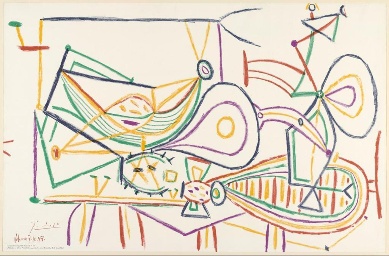}
  \hspace{-4pt}  
  \includegraphics[height=0.078\textwidth,width=0.103\textwidth]{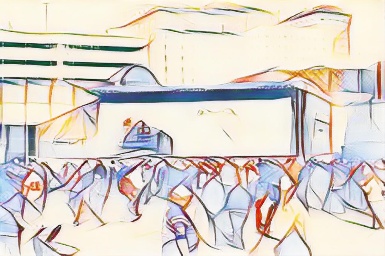}   
  \hspace{-4pt}
  \includegraphics[height=0.078\textwidth,width=0.103\textwidth]{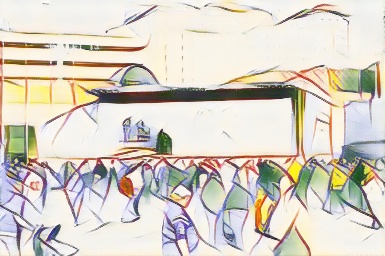} 
  \hspace{-4pt}
  \includegraphics[height=0.078\textwidth,width=0.103\textwidth]{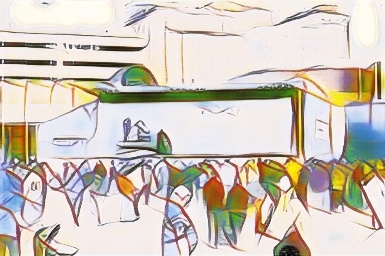}  
 \hspace{-4pt}
  \includegraphics[height=0.078\textwidth,width=0.103\textwidth]{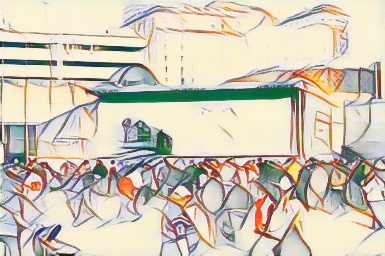}
  \hspace{-4pt}
  \includegraphics[height=0.078\textwidth,width=0.103\textwidth]{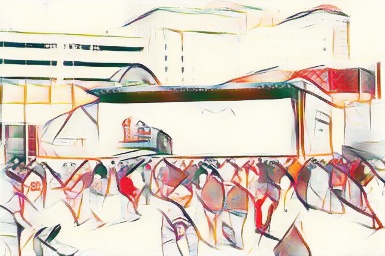}
  \hspace{-4pt}  
  \includegraphics[height=0.078\textwidth,width=0.103\textwidth]{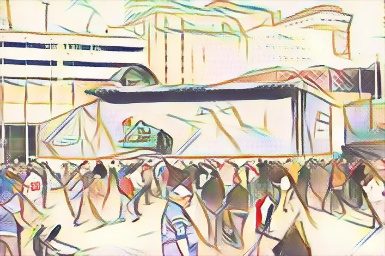} 
  \hspace{-4pt}
  \includegraphics[height=0.078\textwidth,width=0.103\textwidth]{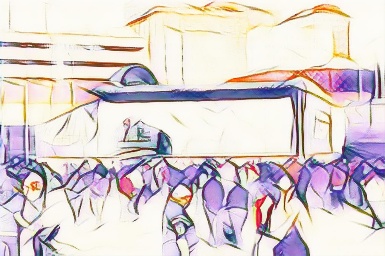} 
  \hspace{-4pt}  
  \includegraphics[height=0.078\textwidth,width=0.078\textwidth]{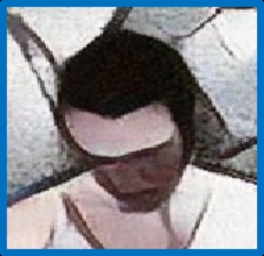}\\
  \vspace{-5pt}
 \subfigure[]{\includegraphics[height=0.078\textwidth,width=0.103\textwidth]{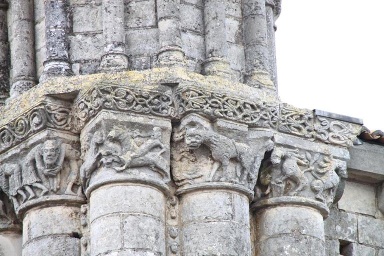}} 
  \hspace{-4pt}   
  \subfigure[]{\includegraphics[height=0.078\textwidth,width=0.068\textwidth]{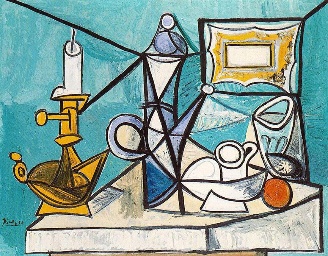}}
  \hspace{-4pt}   
  \subfigure[]{\includegraphics[height=0.078\textwidth,width=0.103\textwidth]{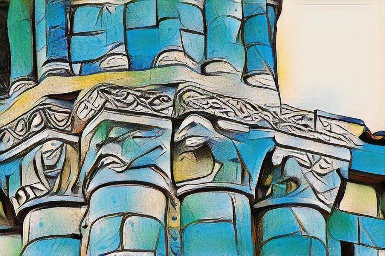}}  
  \hspace{-4pt}  
  \subfigure[]{\includegraphics[height=0.078\textwidth,width=0.103\textwidth]{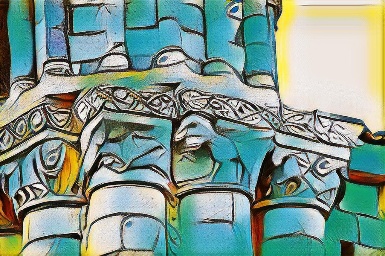}} 
  \hspace{-4pt}  
  \subfigure[]{\includegraphics[height=0.078\textwidth,width=0.103\textwidth]{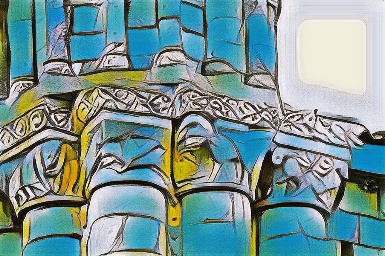}} 
  \hspace{-4pt}   
  \subfigure[]{\includegraphics[height=0.078\textwidth,width=0.103\textwidth]{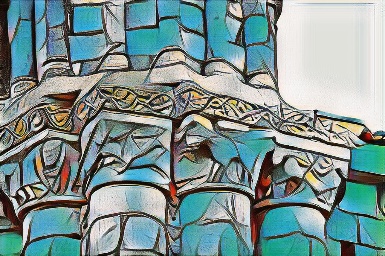}} 
  \hspace{-4pt}  
  \subfigure[]{\includegraphics[height=0.078\textwidth,width=0.103\textwidth]{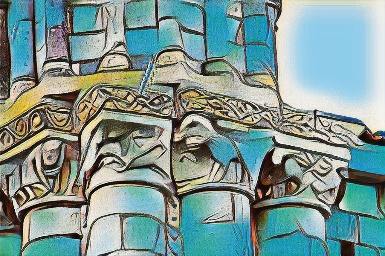}}
  \hspace{-4pt}  
  \subfigure[]{\includegraphics[height=0.078\textwidth,width=0.103\textwidth]{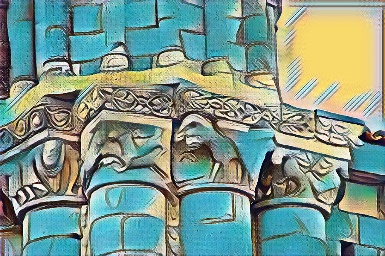}} 
  \hspace{-4pt}   
  \subfigure[]{\includegraphics[height=0.078\textwidth,width=0.103\textwidth]{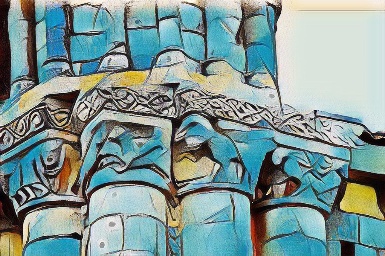}} 
  \hspace{-4pt}   
  \subfigure[]{\includegraphics[height=0.078\textwidth,width=0.078\textwidth]{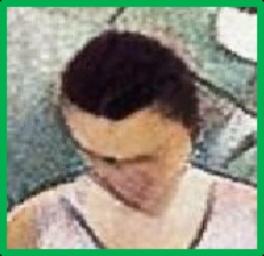}} \\ 
  \vspace{-0.1cm}
\caption{Ablation study of our DRB-GAN. From left to right are (a) content image, (b) style image, the results of (c) DRB-GAN, (d) w/ layer normalization layer in decoder, (e) w/ instance normalization layer in decoder, (f) w/o VGG encoder; (g) w/o $\mathcal{L}_{cls}$; (h) w/o $\mathcal{L}_{adv}$; (i) w/ AdaIN ResBlk, instead of our Dynamic ResBlk, and (j) are the zoom-in details from three regions marked in the top row, respectively.}
\label{fig:Ablation}
\vspace{-0.4cm} 
\end{figure*}

\bfsection{Human perceptual study} We also operate human study on the performance of different approaches. Specifically, we show each participant with 700 groups of images. Each group consists of stylized images generated by different methods based on the same content and style images. We ask the participants to choose one image that most realistically reflects the target style. The style score is computed as the frequency and  
a specific method is chosen as the best in the group. The content score is provided by the participants to evaluate the structural similarity of the content image. Human perceptual studies are reported in Table. \ref{fig:score}. We can see that our method obtains the best scores, proving the superior performance of our model in terms of both style transfer and structure preservation.

\bfsection{Speed and memory}
The comparison on time and memory consumption are also listed in Table. \ref{fig:score}. We observe that our approach has comparable speed and modest demand on GPU memory.  Thanks to the weight generation module that creates a smooth manifold structure, our model can perform flexible style transfer via interpolation and model averaging, which significantly improves the efficiency of our model as a Multiple-Domain-Per-Mode (MDPM) algorithm. On the contrary, other models lack the capability of performing diverse or collection artistic style transfer. 

\begin{table}[ht!]
\centering
\caption{\small Quantitative comparison of different methods. SD stands for style distance metric; DS represents deception score.}
\begin{tabular}{c||c||c|c|c|c}
\hlineB{2} 
\multirow{2}{*}{Setting}&Arbitrary&\multicolumn{4}{c}{Collection style (DS$\uparrow$)}\\
\cline{3-6}
&Style (SD$\downarrow$) &K=2&5&10&20 \\
\hline \hline
 AdaIN &263.4 &0.066& 0.045& 0.013 & 0.011\\  
 MetaNet&271.8  &0.032& 0.026& 0.023 & 0.020\\ 
 DRB-GAN &{\bf 241.2}&{\bf 0.576}& {\bf 0.580}& {\bf 0.581} & {\bf 0.583} \\
\hlineB{2} 
\end{tabular}
\vspace{-0.4cm}	
\label{tab:collection_style_average}
\end{table}


\begin{figure}[ht!]
\centering
  \subfigure[]{
  \includegraphics[height=0.15\linewidth]{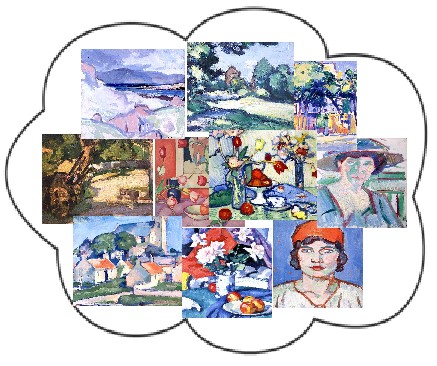} }
  \hspace{-8pt}
  \subfigure[]{  
  \includegraphics[height=0.125\linewidth]{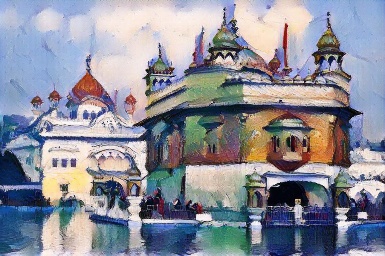} }
  \hspace{-8pt}
  \subfigure[]{
  \includegraphics[height=0.125\linewidth]{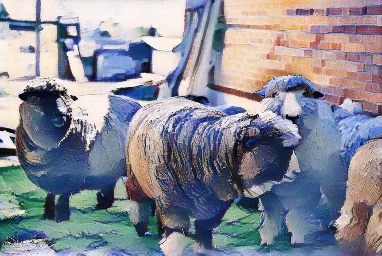} }
  \hspace{-8pt}
  \subfigure[]{
  \includegraphics[height=0.125\linewidth]{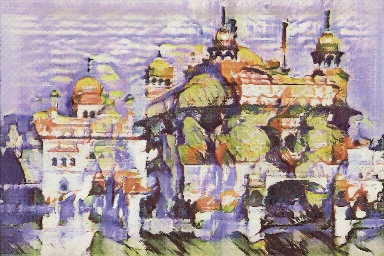} }
  \hspace{-8pt}
  \subfigure[]{
  \includegraphics[height=0.125\linewidth]{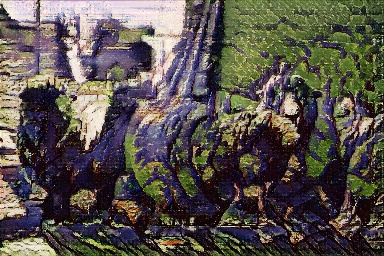}}\\
  \vspace{-0.1cm}
\caption{Performance comparison in terms of collection style transfer. (a) Style collection, (b) and (c) are the result of our DRB-GAN, and (d) and (e) are the results of AST. }
\label{fig:Comp_collection_style}
\vspace{-0.3cm} 
\end{figure}

\begin{figure}[ht!]
\centering
  \includegraphics[height=0.08\textwidth,width=0.1\textwidth]{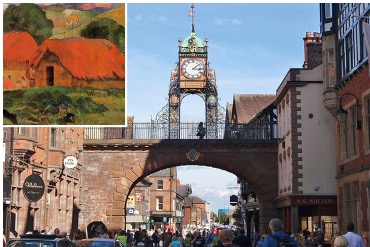} 
  \hspace{-4pt}
  \includegraphics[height=0.08\textwidth,width=0.1\textwidth]{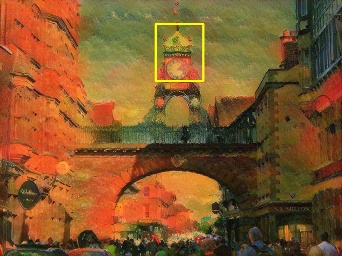} 
  \hspace{-4pt}  
  \includegraphics[height=0.08\textwidth,width=0.1\textwidth]{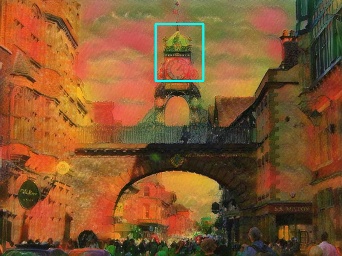}  
 \hspace{-4pt}
  \includegraphics[height=0.08\textwidth,width=0.1\textwidth]{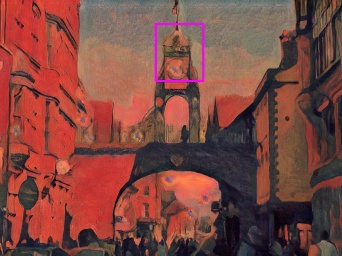}  
 \hspace{-4pt}  
  \includegraphics[height=0.08\textwidth,width=0.06\textwidth]{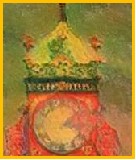} \\
  \includegraphics[height=0.08\textwidth,width=0.1\textwidth]{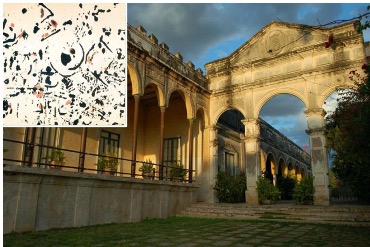} 
  \hspace{-4pt}
  \includegraphics[height=0.08\textwidth,width=0.1\textwidth]{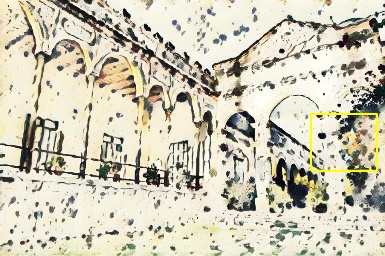} 
  \hspace{-4pt}  
  \includegraphics[height=0.08\textwidth,width=0.1\textwidth]{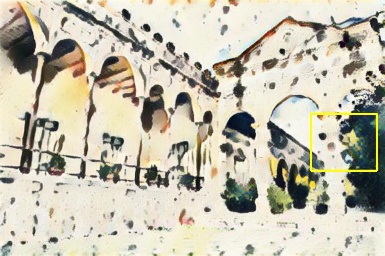}  
 \hspace{-4pt}
  \includegraphics[height=0.08\textwidth,width=0.1\textwidth]{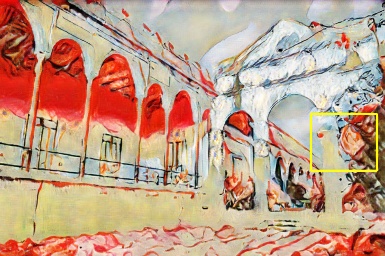}  
 \hspace{-4pt} 
  \includegraphics[height=0.08\textwidth,width=0.06\textwidth]{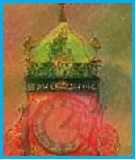} \\
  \vspace{-5.pt}
  \subfigure[]{\includegraphics[height=0.08\textwidth,width=0.1\textwidth]{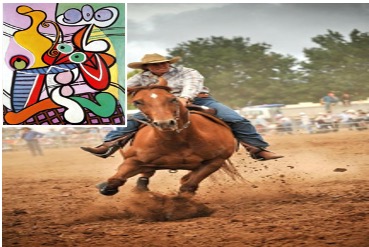}} 
  \hspace{-4pt} 
  \subfigure[]{\includegraphics[height=0.08\textwidth,width=0.1\textwidth]{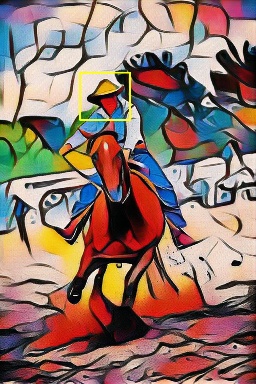}} 
  \hspace{-4pt} 
  \subfigure[]{\includegraphics[height=0.08\textwidth,width=0.1\textwidth]{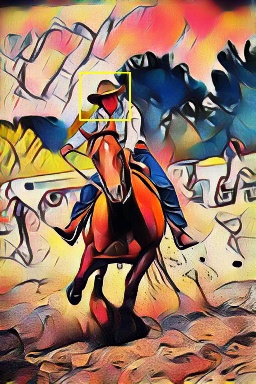}} 
  \hspace{-4pt}
  \subfigure[]{\includegraphics[height=0.08\textwidth,width=0.1\textwidth]{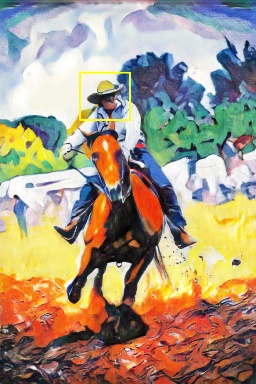}} 
  \hspace{-4pt}  
  \subfigure[]{\includegraphics[height=0.08\textwidth,width=0.06\textwidth]{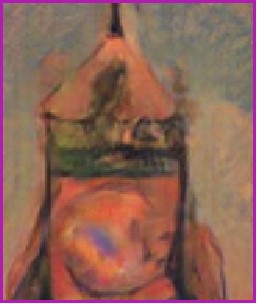}}   \\  
  \vspace{-0.1cm}
\caption{Performance comparison in terms of collection discriminator. (a) content/style images, (b) DRB-GAN w/ collection discriminator, (c) DRB-GAN w/ conditional discriminator, (d) CST and (e) are the zoom-in details from three regions marked in the top row, respectively. }
\label{fig::collection_dis}
\vspace{-0.5cm} 
\end{figure}

\begin{figure*}[ht!]
\vspace{-0.5cm}
\centering
 \includegraphics[height=0.078\textwidth,width=0.103\textwidth]{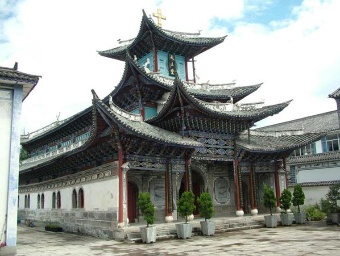}
  \hspace{-4pt}
  \includegraphics[height=0.078\textwidth]{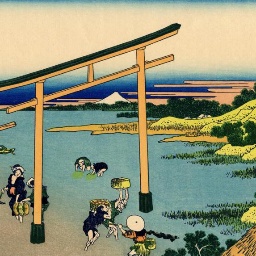}
  \hspace{-4pt}
  \includegraphics[height=0.078\textwidth,width=0.103\textwidth]{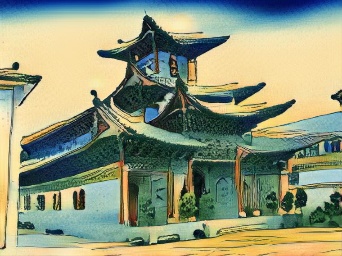} 
  \hspace{-4pt}
  \includegraphics[height=0.078\textwidth,width=0.103\textwidth]{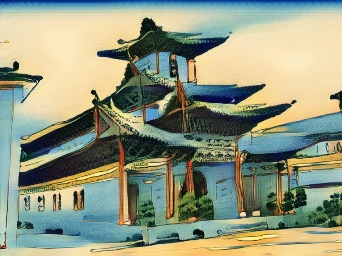}  
 \hspace{-4pt}
  \includegraphics[height=0.078\textwidth,width=0.103\textwidth]{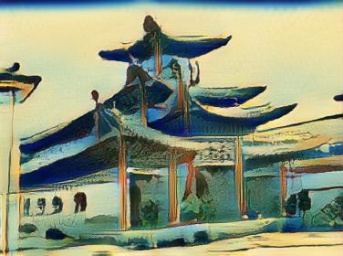}
  \hspace{-4pt}
  \includegraphics[height=0.078\textwidth,width=0.103\textwidth]{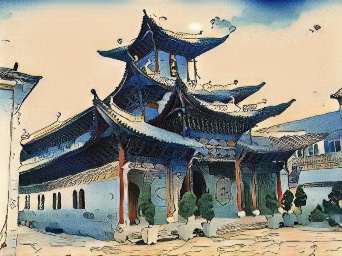}
  \hspace{-4pt}  
  \includegraphics[height=0.078\textwidth,width=0.103\textwidth]{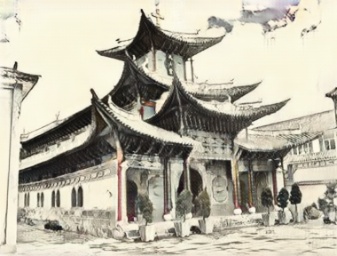} 
  \hspace{-4pt}
  \includegraphics[height=0.078\textwidth,width=0.103\textwidth]{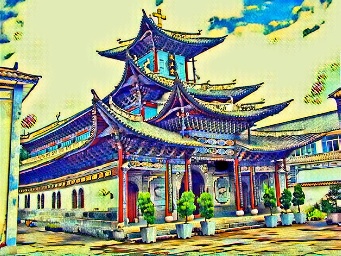} 
  \hspace{-4pt}  
  \includegraphics[height=0.078\textwidth,width=0.103\textwidth]{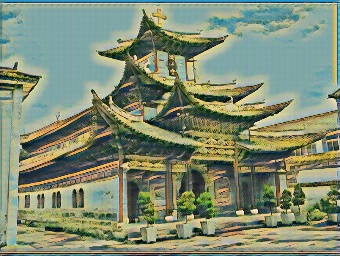}
  \hspace{-4pt}  
  \includegraphics[height=0.078\textwidth]{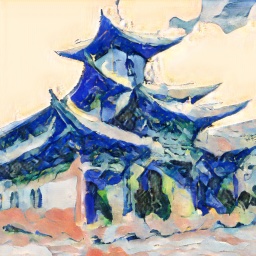} \\
 \includegraphics[height=0.078\textwidth,width=0.103\textwidth]{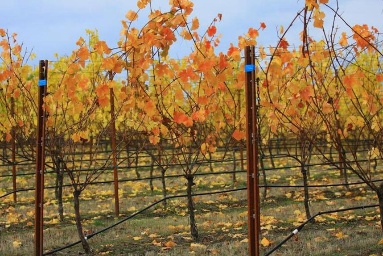}
  \hspace{-4pt}
  \includegraphics[height=0.078\textwidth]{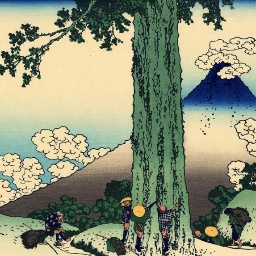}
  \hspace{-4pt}
  \includegraphics[height=0.078\textwidth,width=0.103\textwidth]{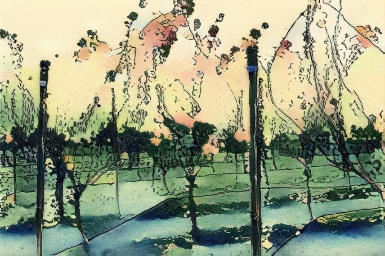} 
  \hspace{-4pt}
  \includegraphics[height=0.078\textwidth,width=0.103\textwidth]{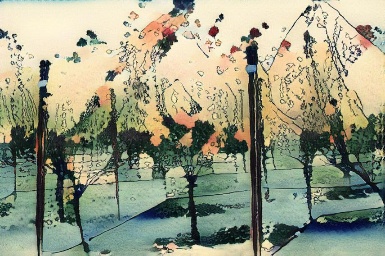}  
 \hspace{-4pt}
  \includegraphics[height=0.078\textwidth,width=0.103\textwidth]{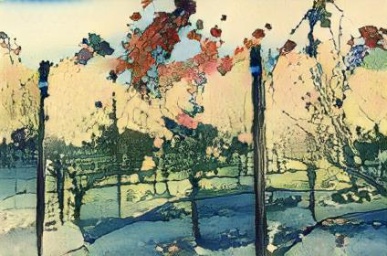}
  \hspace{-4pt}
  \includegraphics[height=0.078\textwidth,width=0.103\textwidth]{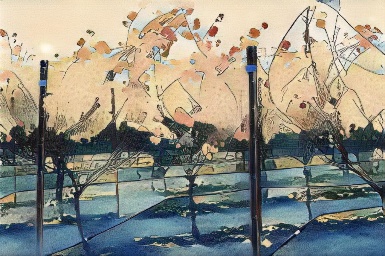}
  \hspace{-4pt}  
  \includegraphics[height=0.078\textwidth,width=0.103\textwidth]{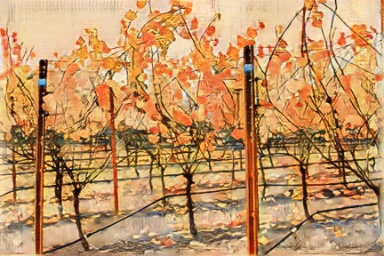} 
  \hspace{-4pt}
  \includegraphics[height=0.078\textwidth,width=0.103\textwidth]{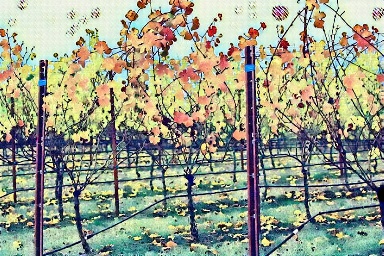} 
  \hspace{-4pt}  
  \includegraphics[height=0.078\textwidth,width=0.103\textwidth]{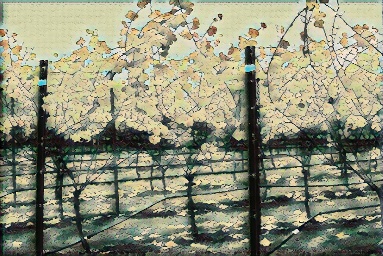}
  \hspace{-4pt}  
  \includegraphics[height=0.078\textwidth]{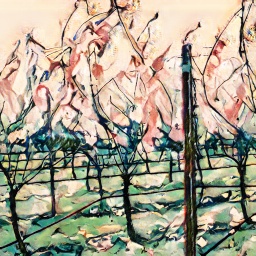} \\
  \vspace{-5pt}
 \subfigure[]{\includegraphics[height=0.078\textwidth,width=0.103\textwidth]{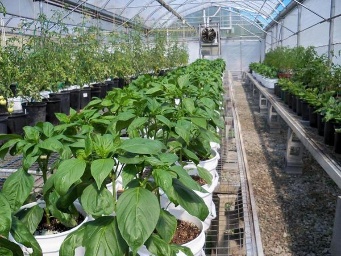}} 
  \hspace{-4pt}   
  \subfigure[]{\includegraphics[height=0.078\textwidth]{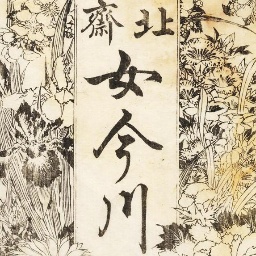}}
  \hspace{-4pt}  
  \subfigure[]{\includegraphics[height=0.078\textwidth,width=0.103\textwidth]{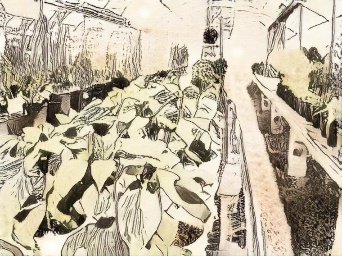}} 
  \hspace{-4pt}  
  \subfigure[]{\includegraphics[height=0.078\textwidth,width=0.103\textwidth]{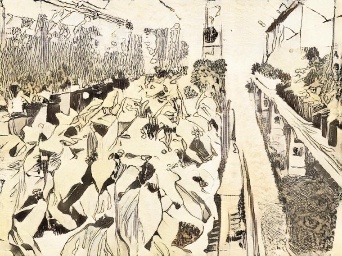}} 
  \hspace{-4pt}   
  \subfigure[]{\includegraphics[height=0.078\textwidth,width=0.103\textwidth]{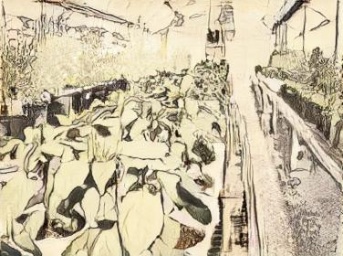}} 
  \hspace{-4pt}  
  \subfigure[]{\includegraphics[height=0.078\textwidth,width=0.103\textwidth]{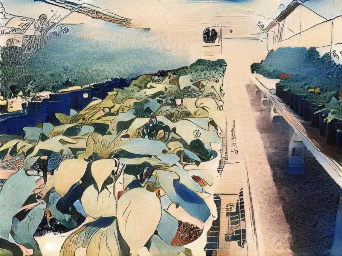}}
  \hspace{-4pt}  
  \subfigure[]{\includegraphics[height=0.078\textwidth,width=0.103\textwidth]{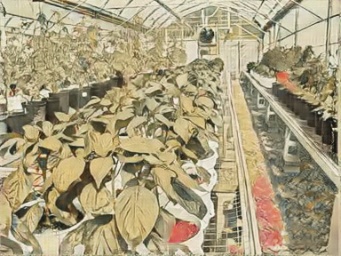}} 
  \hspace{-4pt}   
  \subfigure[]{\includegraphics[height=0.078\textwidth,width=0.103\textwidth]{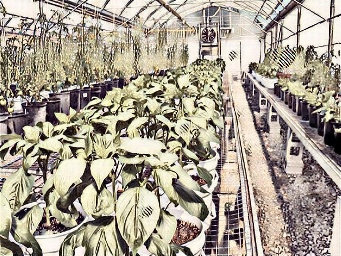}} 
  \hspace{-4pt}   
  \subfigure[]{\includegraphics[height=0.078\textwidth,width=0.103\textwidth]{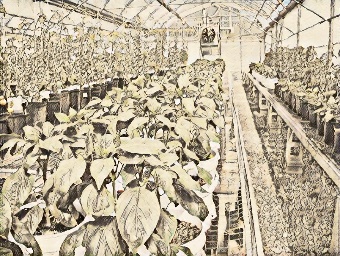}}
  \hspace{-4pt}   
  \subfigure[]{\includegraphics[height=0.078\textwidth]{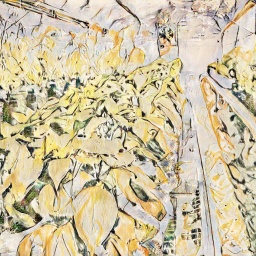}} \\  
  \vspace{-0.1cm}
\caption{Performance comparison on stylized results from different models trained on images with different resolutions. From left to right are (a) content image, (b) style image, the results of (c) DRB-GAN{\color{blue}$\times$768}, (d) DRB-GAN{\color{blue}$\times$512}, (e) DRB-GAN{\color{blue}$\times$256}, (f) DRB-GAN@20{\color{blue}$\times$512}, (g) CycleGAN{\color{blue}$\times$256}, (h) MetaNet{\color{blue}$\times$512}, (i) StyleBank{\color{blue}$\times$512}, and (j) AST{\color{blue}$\times$512},  respectively. 
The number in {\color{blue}blue} indicates the size of the smaller edge of the original image.
}
\label{fig:U}
\vspace{-0.3cm} 
\end{figure*}

\subsection{Collection Style Transfer} 
We denote our DRB-GAN model for collection style transfer as DRB-GAN@K, where $K$ is the number of painting images from the same artist used to create the averaged transformer network. We test a range of $K$ values, including 2, 5, 10, and 20. Especially, DRB-GAN is equivalent to DRB-GAN@1 for arbitrary style transfer. In Table \ref{tab:collection_style_average}, we demonstrate the deception score obtained with these different $K$ values. A larger $K$ leads to more improvements, and the largest performance boost comes from $K = 2$ to $K = 5$. The stylizations of DRB-GAN@20 is provided in Figure~ \ref{fig:Comp_collection_style}. As we can see, in comparison to the AST, the results of DRB-GAN@20 are better with sharper details and reflect the dominant clue of the artistic style. This also can be observed from Figure \ref{fig:U} (f) and (j).

\begin{figure*}[ht!]
\centering
 \includegraphics[width=0.096\textwidth, height=0.065\textwidth]{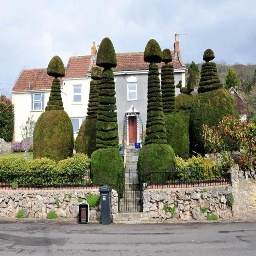}
  \includegraphics[width=0.096\textwidth, height=0.065\textwidth]{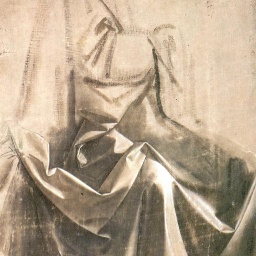}
  \hspace{-4pt}
  \includegraphics[width=0.096\textwidth, height=0.065\textwidth]{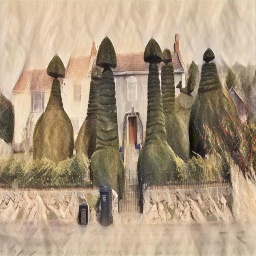} 
  \hspace{-4pt}
  \includegraphics[width=0.096\textwidth, height=0.065\textwidth]{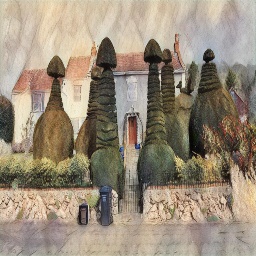}  
  \includegraphics[width=0.096\textwidth, height=0.065\textwidth]{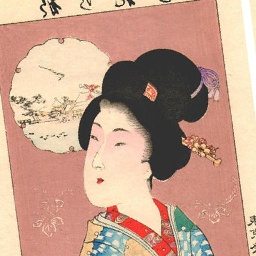}
  \hspace{-4pt}
  \includegraphics[width=0.096\textwidth, height=0.065\textwidth]{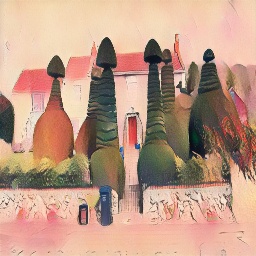}
  \hspace{-4pt}  
  \includegraphics[width=0.096\textwidth, height=0.065\textwidth]{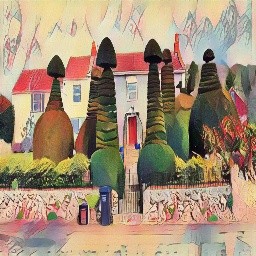} 
  \includegraphics[width=0.096\textwidth, height=0.065\textwidth]{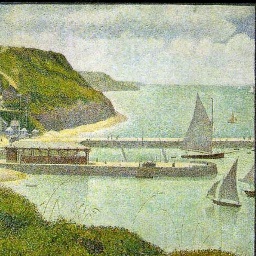} 
  \hspace{-4pt}  
  \includegraphics[width=0.096\textwidth, height=0.065\textwidth]{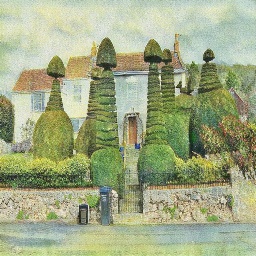}
  \hspace{-4pt}  
  \includegraphics[width=0.096\textwidth, height=0.065\textwidth]{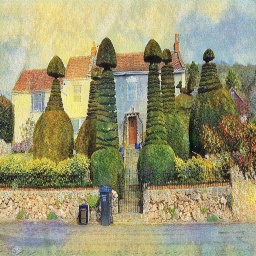} \\
 \includegraphics[width=0.096\textwidth, height=0.065\textwidth]{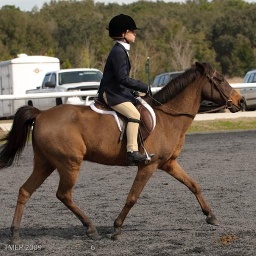}
  \includegraphics[width=0.096\textwidth, height=0.065\textwidth]{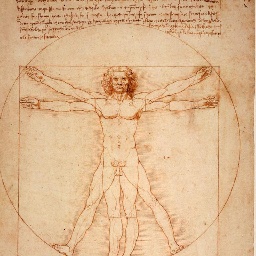}
  \hspace{-4pt}
  \includegraphics[width=0.096\textwidth, height=0.065\textwidth]{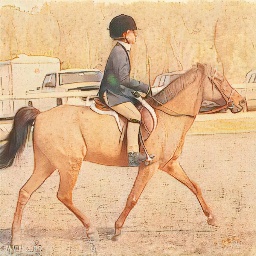} 
  \hspace{-4pt}
  \includegraphics[width=0.096\textwidth, height=0.065\textwidth]{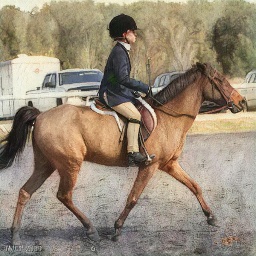}  
  \includegraphics[width=0.096\textwidth, height=0.065\textwidth]{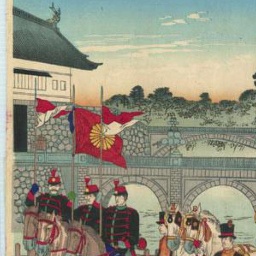}
  \hspace{-4pt}
  \includegraphics[width=0.096\textwidth, height=0.065\textwidth]{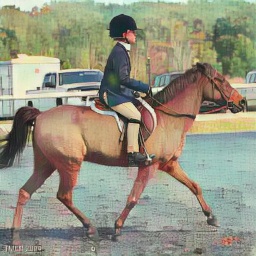}
  \hspace{-4pt}  
  \includegraphics[width=0.096\textwidth, height=0.065\textwidth]{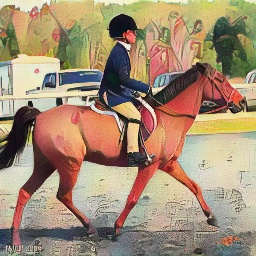} 
  \includegraphics[width=0.096\textwidth, height=0.065\textwidth]{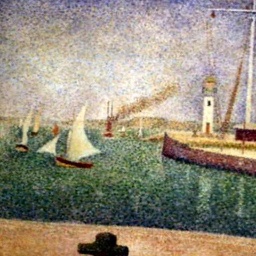} 
  \hspace{-4pt}  
  \includegraphics[width=0.096\textwidth, height=0.065\textwidth]{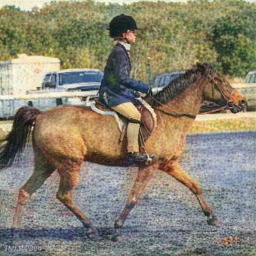}
  \hspace{-4pt}  
  \includegraphics[width=0.096\textwidth, height=0.065\textwidth]{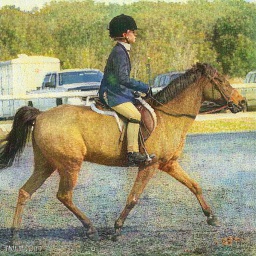} \\
 \includegraphics[width=0.096\textwidth, height=0.065\textwidth]{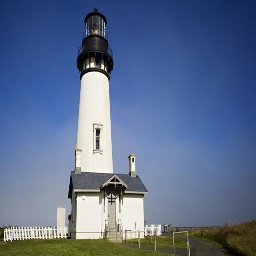}
  \includegraphics[width=0.096\textwidth, height=0.065\textwidth]{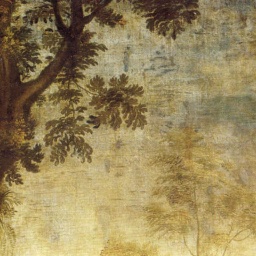}
  \hspace{-4pt}
  \includegraphics[width=0.096\textwidth, height=0.065\textwidth]{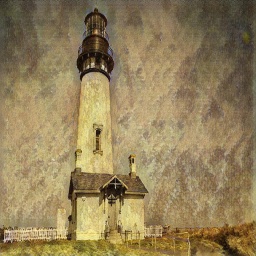} 
  \hspace{-4pt}
  \includegraphics[width=0.096\textwidth, height=0.065\textwidth]{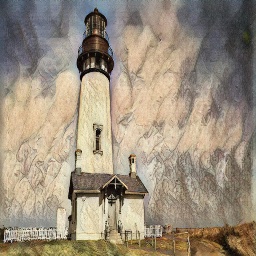}  
  \includegraphics[width=0.096\textwidth, height=0.065\textwidth]{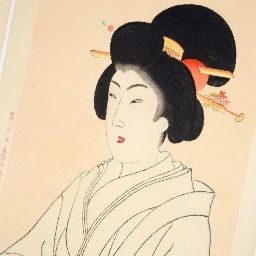}
  \hspace{-4pt}
  \includegraphics[width=0.096\textwidth, height=0.065\textwidth]{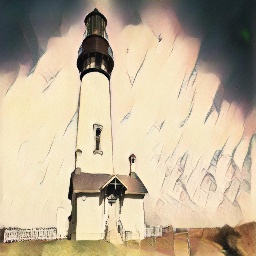}
  \hspace{-4pt}  
  \includegraphics[width=0.096\textwidth, height=0.065\textwidth]{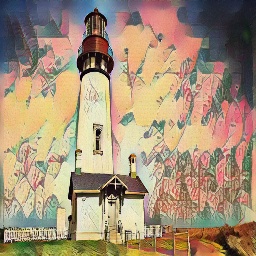} 
  \includegraphics[width=0.096\textwidth, height=0.065\textwidth]{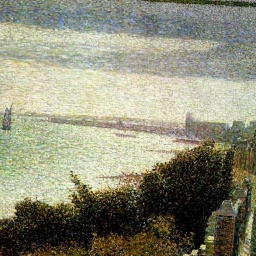} 
  \hspace{-4pt}  
  \includegraphics[width=0.096\textwidth, height=0.065\textwidth]{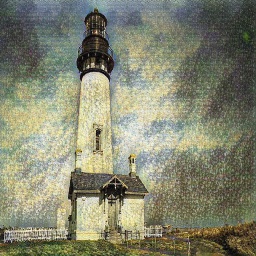}
  \hspace{-4pt}  
  \includegraphics[width=0.096\textwidth, height=0.065\textwidth]{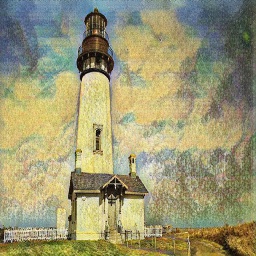} \\
 \includegraphics[width=0.096\textwidth, height=0.065\textwidth]{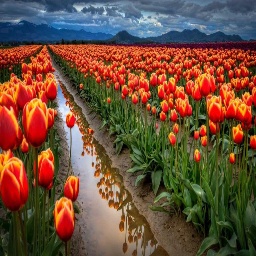}
  \includegraphics[width=0.096\textwidth, height=0.065\textwidth]{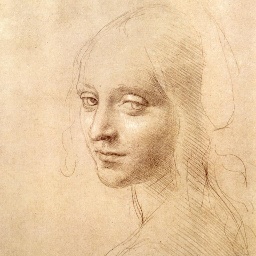}
  \hspace{-4pt}
  \includegraphics[width=0.096\textwidth, height=0.065\textwidth]{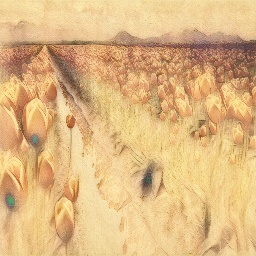} 
  \hspace{-4pt}
  \includegraphics[width=0.096\textwidth, height=0.065\textwidth]{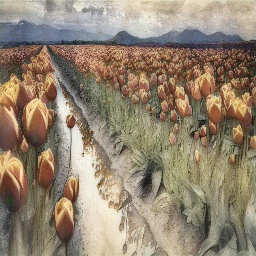}  
  \includegraphics[width=0.096\textwidth, height=0.065\textwidth]{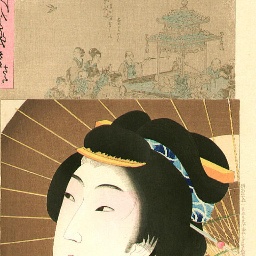}
  \hspace{-4pt}
  \includegraphics[width=0.096\textwidth, height=0.065\textwidth]{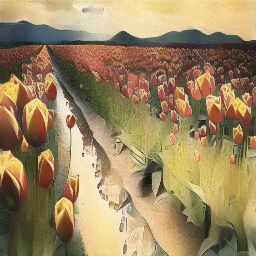}
  \hspace{-4pt}  
  \includegraphics[width=0.096\textwidth, height=0.065\textwidth]{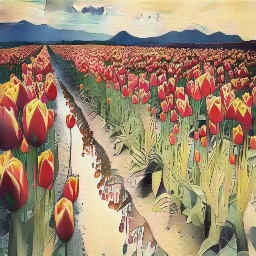} 
  \includegraphics[width=0.096\textwidth, height=0.065\textwidth]{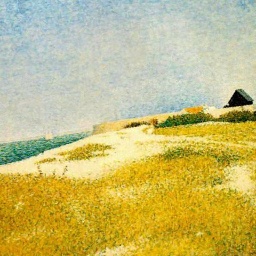} 
  \hspace{-4pt}  
  \includegraphics[width=0.096\textwidth, height=0.065\textwidth]{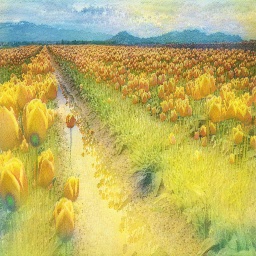}
  \hspace{-4pt}  
  \includegraphics[width=0.096\textwidth, height=0.065\textwidth]{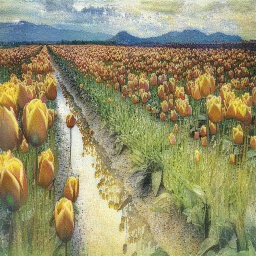} \\
  \vspace{-4.4pt}
 \subfigure[Content]{\includegraphics[width=0.096\textwidth, height=0.065\textwidth]{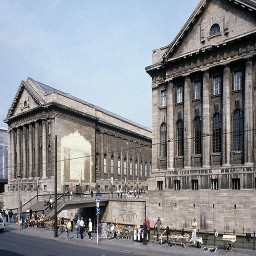}} 
  \subfigure[DaVinci]{\includegraphics[width=0.096\textwidth, height=0.065\textwidth]{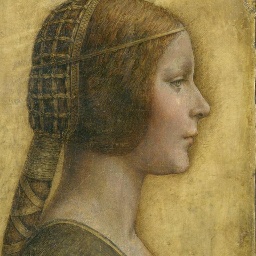}}
  \hspace{-4pt}  
  \subfigure[]{\includegraphics[width=0.096\textwidth, height=0.065\textwidth]{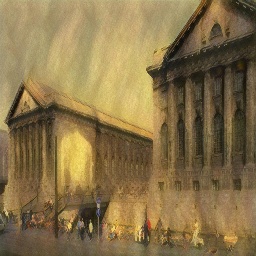}} 
  \hspace{-4pt}  
  \subfigure[]{\includegraphics[width=0.096\textwidth, height=0.065\textwidth]{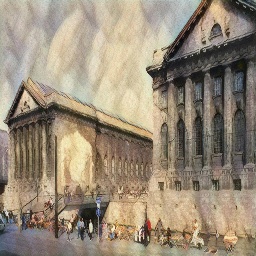}} 
  \subfigure[Chikanobu]{\includegraphics[width=0.096\textwidth, height=0.065\textwidth]{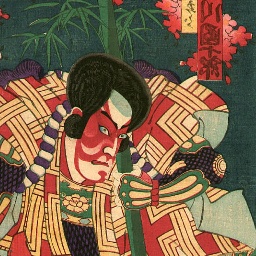}} 
  \hspace{-4pt}  
  \subfigure[]{\includegraphics[width=0.096\textwidth, height=0.065\textwidth]{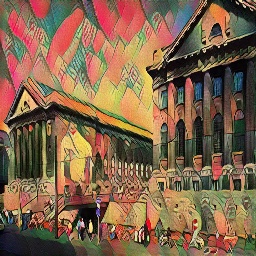}}
  \hspace{-4pt}  
  \subfigure[]{\includegraphics[width=0.096\textwidth, height=0.065\textwidth]{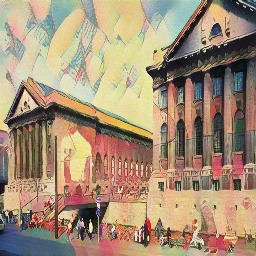}} 
  \subfigure[Seurat]{\includegraphics[width=0.096\textwidth, height=0.065\textwidth]{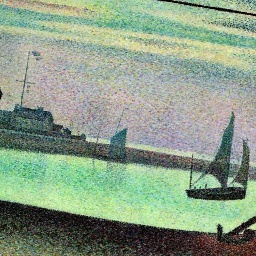}} 
  \hspace{-4pt}   
  \subfigure[]{\includegraphics[width=0.096\textwidth, height=0.065\textwidth]{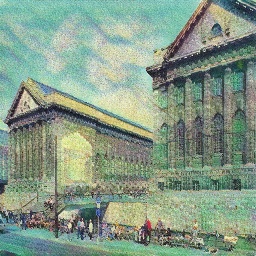}}
  \hspace{-4pt}   
  \subfigure[]{\includegraphics[width=0.096\textwidth, height=0.065\textwidth]{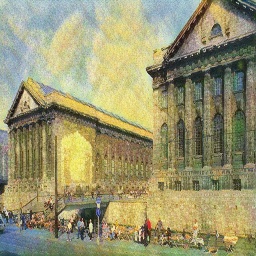}} \\  
  \vspace{-0.1cm}
\caption{Qualitative evaluation of our method for previously unseen styles. It can be observed that the generated images are consistent with the provided target style, (c),(f),(i), showing the good generalization capabilities of the approach. And it can also be observed that our model shows good performance on collection style transfer, (d),(g),(j). }
\label{fig:nonseestyle}
\vspace{-0.5cm} 
\end{figure*}

\subsection{Ablation Study}
On arbitrary style transfer task, we perform the ablation study by removing or replacing each part of the network to validate the effectiveness of the proposed network and summarize the results in Figure~\ref{fig:Ablation} and Table~\ref{tab:ablation_study}. 

As we can see, without using the layer-instance normalization function in the decoder, the model either produce degraded stylizations when the layer normalization function is used in the decoder (see Figure \ref{fig:Ablation} (d)), or create artifacts in the stylizations when using instance normalization (see Figure \ref{fig:Ablation} (e)). The results validate our claims that instance normalization normalizes each feature map separately, thereby potentially destroying the information in feature maps. 
Note that layer normalization operation normalizes the feature map together, thereby potentially demolishing each feature map as the representation of the style. Training without VGG encoder opts to capture the dominant style clues (see Figure \ref{fig:Ablation} (f)) without subtle details. 

\setlength{\tabcolsep}{8pt}
\begin{table}[ht!]
\small 
\centering
\vspace{-0.2cm}
\caption{\small Quantitative analysis of ablation study. DS: deception score, CS: content score, and SS: style score.}
\begin{tabular}{l||c|c|c}
\hlineB{2} 
Method &  DS$\uparrow$ &  CS$\uparrow$ &  SS$\uparrow$ \\
\hline
DRB-GAN w/ IN Decoder& 0.561 &42.1\%&0.007 \\  
DRB-GAN w/ LN Decoder& 0.568 &54.7\%&0.102 \\  
DRB-GAN w/ AdaIN ResBlk& 0.552&36.2\%&0.006 \\
DRB-GAN w/o VGG Encoder& 0.570& 66.3\%&0.228 \\ 
DRB-GAN w/o $\mathcal{L}_{cls}$ & 0.571 &70.2\%&0.273 \\
DRB-GAN w/o $\mathcal{L}_{adv}$ &0.542&32.2\%&0.001\\
DRB-GAN & {\bf 0.573}&{\bf 72.2\%}&{\bf 0.383} \\ 
\hlineB{2} 
\end{tabular}
\label{tab:ablation_study}
 \vspace{-0.4cm} 
\end{table}

Moreover, we observe that training without the attention recalibration module ($D_{cls}$) causes slight degradation on stroke size variations. One possible reason is that the attention recalibration module works to shorten the gap between the specific style pattern of one individual style image and the overall style clues of the domain (see Figure \ref{fig:Ablation} (d)). Adversarial training is essential to improve the visual quality of the generated images (see Figure \ref{fig:Ablation} (e)). We successfully combine adversarial and perceptual supervision to obtain high-quality style transfer. Finally, we demonstrate the advantage of our Dynamic Resblk over the AdaIN Resblk (see Figure \ref{fig:Ablation} (f)). 

\bfsection{Effect of collection discriminator} In Figure \ref{fig::collection_dis}, we demonstrate the effect of our collection discriminator. Comparing to conditional GAN whose discriminator takes a category label as an additional input, our DRB-GAN produces better stylized images. 
However, the CST fails to maintain the style consistency between the synthesized images and target style images.  

\subsection{Discussions}
{\noindent\bf Influence of image resolution}. Our DRB-GAN is efficient in high-resolution image style transfer. It is also robust to perform style transfer on images with different resolutions. To illustrate the influence of image resolutions, we show the qualitative comparison in Figure \ref{fig:U}. It demonstrates that our model creates consistent stylizations on different image resolutions with slight variations. The results are much better than other baseline models on the same image resolution. 

{\noindent\bf Effectiveness of unseen styles}. We apply our DRB-GAN to handle the unseen styles for both arbitrary style transfer and collection style transfer tasks. The visualization results are provided in Figure~\ref{fig:nonseestyle}. Apparently, these \if promising \fi results strongly demonstrate the robustness of our proposed method.
\section{Conclusion}
\if In this paper,\fi We have presented the DRB-GAN for artistic style transfer. In our model, ``{\em style codes}" is modeled as the shared parameters, for Dynamic ResBlocks connecting both the style encoding network and the style transfer network to shrink the gap between arbitrary style transfer and collection style transfer in one single model. The proposed attention mechanism and discriminative network make full use of style information in target style images and thus encourage our model's ability for artistic style transfer. Extensive experimental results clearly demonstrate the remarkable performance of our proposed DRB-GAN model in generating synthetic style images with better quality than the state-of-the-art. 
{\small
\bibliographystyle{ieee_fullname}
\bibliography{egbib}
}
\balance

\section{Appendix}

\subsection{Configuration of Our DRB-GAN Networks}
The Dynamic Convolution layer is a $3\times3$ convolution layer where the weights are produced by the hyper-network. A basic Conv layer contains a $3\times3$ convolution operation, a batch-normalization, and a ReLU activation sequentially. The learnable encoder is a CNN consisting of 5 Conv layers with stride size 2. The Content Encoder is a CNN having 2 Conv layers with stride size 2. The SW-LIN decoder is a CNN decoder using the SW-LIN to modulate the activations in normalization layers. And the spatial window is to be a center region of size $(H-h,W-w,C)$ given its corresponding feature map of size $(H,W,C)$, where H, W are the spatial resolution and C is the channel dimension. The style code is a concatenation of outputs from fixed VGG and learnable encoder. In our collection discriminator, the feature extractor contains 2 Conv layers with stride size 2 and $D$ network consists of 3 Conv layers with stride size 2.

\subsection{Effectiveness of Dynamic Blocks}
\begin{figure*}[ht!]
\centering
\includegraphics[width=1.0\textwidth]{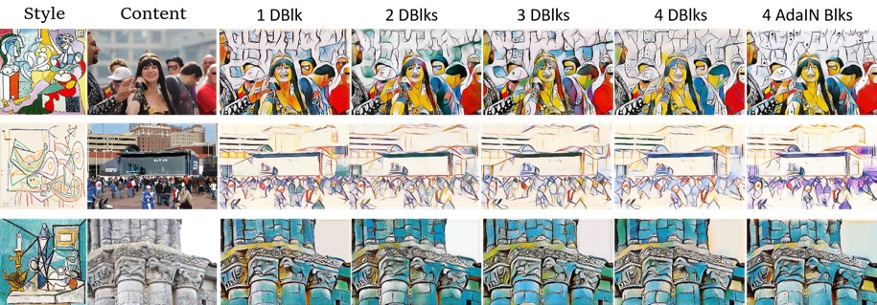}
\caption{The effect of the Dynamic Blocks}
\label{fig:Dblk}
\end{figure*}

The proposed dynamic module is an important network design. We further conduct two experiments. One is to show the
advantage of the progressive generation strategy by varying the number of the
proposed dynamic blocks, and the other is to explore the advantage of the dynamic
block by replacing it with the AdaIN residual block. Quantitative and qualitative results are shown in Table \ref{tab:Dblk} and Figure \ref{fig:Dblk}. We observe that the proposed model with 4 blocks works the best.
However, only using one dynamic block could outperform the model using AdaIN blocks. This could be attributed to the proposed dynamic block. Therefore, we adopt 4 dynamic blocks as default in our experiments.

\setlength{\tabcolsep}{4pt}
\begin{table}[h]
\footnotesize
\centering
\caption{\footnotesize Quantitative results of using different numbers of dynamic blocks.}
\begin{tabular}{c||c|c|c|c|c}
\hlineB{2} 
Setting &1 DBk&2 DBlks&3 DBlks&4 Dblks&4 AdaIN Blks \\
\hline \hline
 Score &0.567 &0.570 &0.571& 0.573& 0.558  \\ 
 GPU Memory (MB)&1084&1164&1244&1324&1149\\ 
 
\hlineB{2} 
\end{tabular}
\label{tab:Dblk}
\end{table}

\subsection{Consistency on Collection Style Transfer}
\begin{figure*}[h]
\begin{center}
    \centering
    \includegraphics[width=0.98\textwidth]{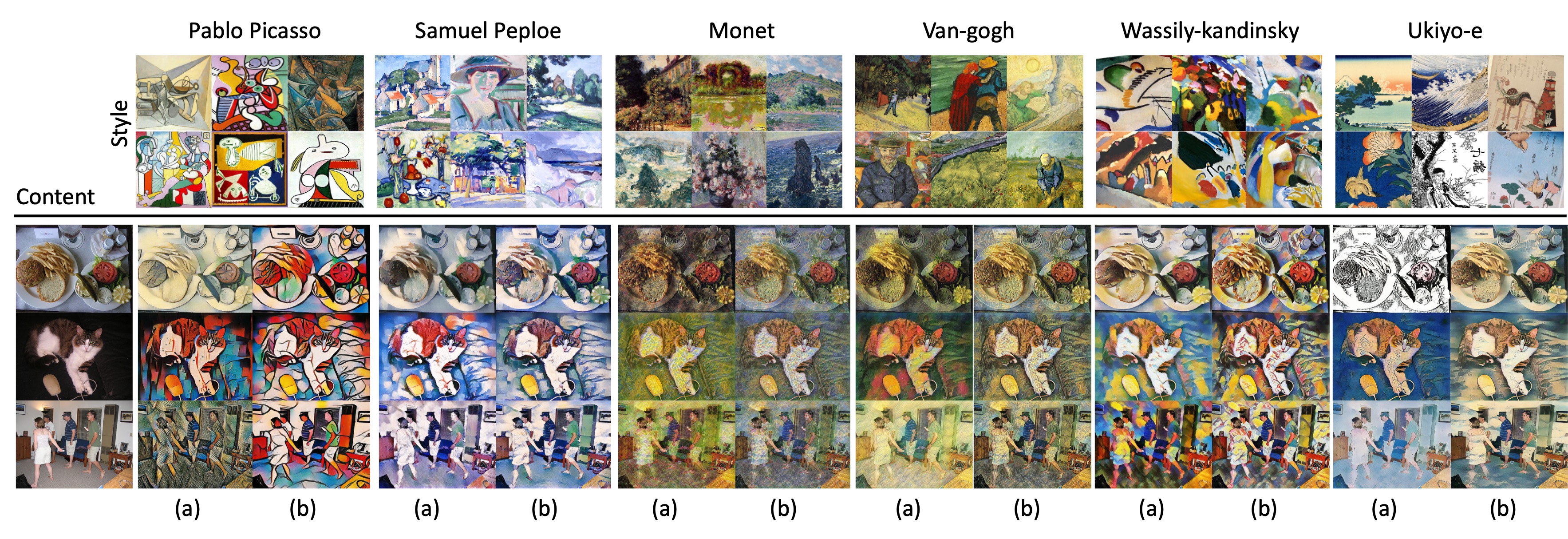}
    \caption{Examples of two types of artistic style transfer: (a) arbitrary style transfer and (b) collection style transfer. Our proposed is able to conduct arbitrary style transfer well and even can extend to handle collection style transfer.}
\vspace{-0.2cm}     
\end{center}%
\label{fig:avg}
\end{figure*}

Our DRB-GAN model achieves both arbitrary style transfer and collection style transfer. Note that we adopt the DRB-GAN@20 for collection style transfer. The stylizations are listed in Figure \ref{fig:avg}. We observe that the results of collection style transfer are consistent in stroke size and color, which further demonstrates that our DRB-GAN model has a good capability to transfer domain-level style with very few blocks.

\subsection{More Visualization Results}
\begin{figure*}[h]
\begin{center}
\includegraphics[width=1.0\linewidth]{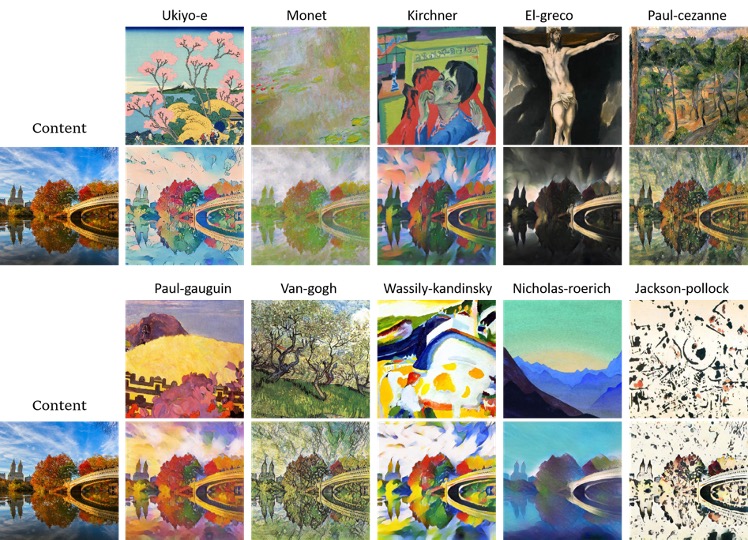}
\end{center}
\vspace{-0.5cm}
\caption{ Qualitative performance of one unified model on 10 different styles  unified model .}
\label{fig:dem_10}
\end{figure*}

{\noindent\bf Style Transfer Involving Multiple Domains.}
To further understand the advantage of our proposed DRB-GAN, we present additional visualization results in Figure \ref{fig:dem_10}. As we can see, our model can efficiently transfer the content image into 10 different styles by only using one unified model.

\begin{figure}[b!]
\hspace{0.066\textwidth}
  \hspace{-4pt}   
 \includegraphics[height=0.066\textwidth]{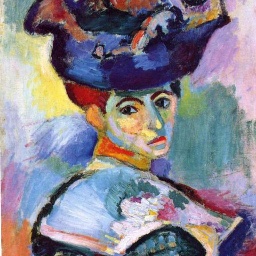}
  \hspace{-4pt}
 \includegraphics[height=0.066\textwidth]{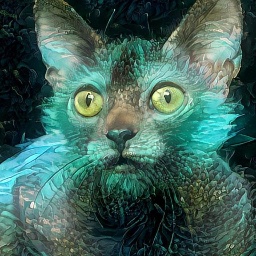}
  \hspace{-4pt}
 \includegraphics[height=0.066\textwidth]{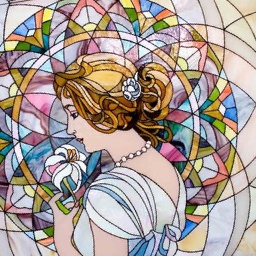}
  \hspace{-4pt}
 \includegraphics[height=0.066\textwidth]{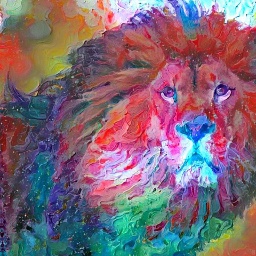} 
 \hspace{-4pt}
 \includegraphics[height=0.066\textwidth]{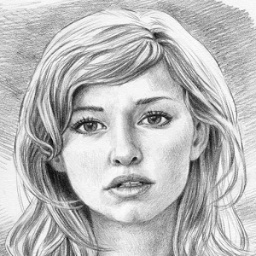}
  \hspace{-4pt}
 \includegraphics[height=0.066\textwidth]{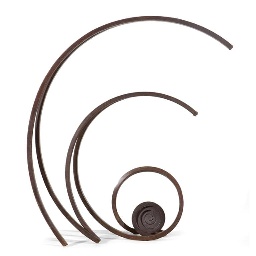}
 
 \includegraphics[height=0.066\textwidth]{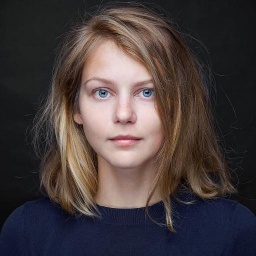} 
  \hspace{-4pt}   
 \includegraphics[height=0.066\textwidth]{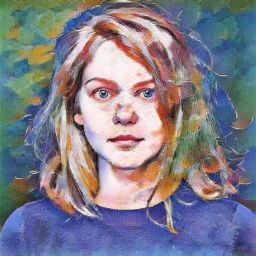}
  \hspace{-4pt}
 \includegraphics[height=0.066\textwidth]{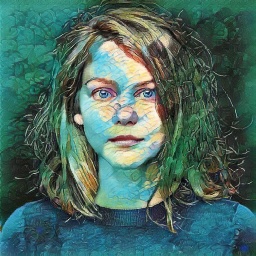}
  \hspace{-4pt}
 \includegraphics[height=0.066\textwidth]{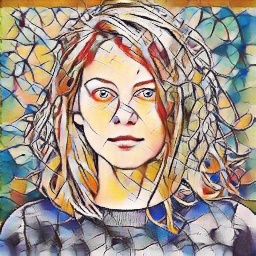}
  \hspace{-4pt}
 \includegraphics[height=0.066\textwidth]{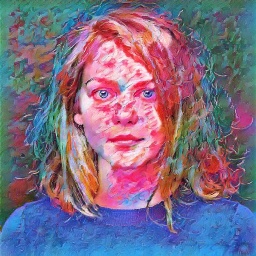} 
 \hspace{-4pt}
 \includegraphics[height=0.066\textwidth]{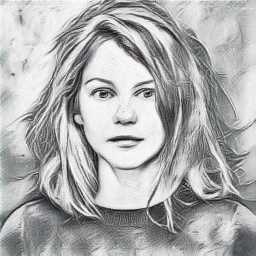}
  \hspace{-4pt}
 \includegraphics[height=0.066\textwidth]{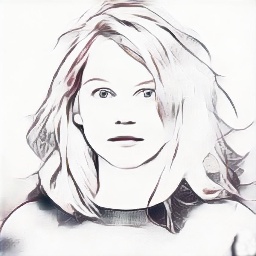}
 
 \includegraphics[height=0.066\textwidth]{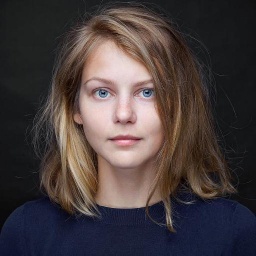} 
  \hspace{-4pt}   
 \includegraphics[height=0.066\textwidth]{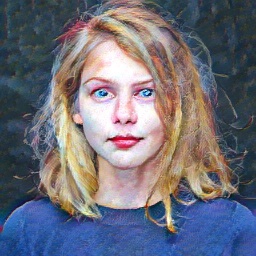}
  \hspace{-4pt}
 \includegraphics[height=0.066\textwidth]{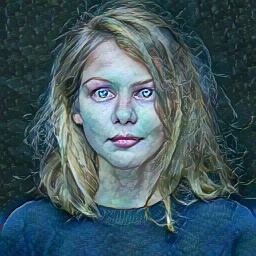}
  \hspace{-4pt}
 \includegraphics[height=0.066\textwidth]{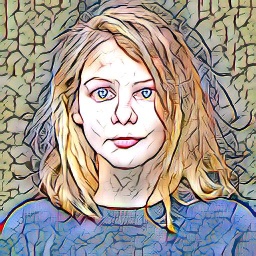}
  \hspace{-4pt}
 \includegraphics[height=0.066\textwidth]{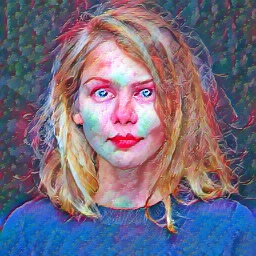} 
 \hspace{-4pt}
 \includegraphics[height=0.066\textwidth]{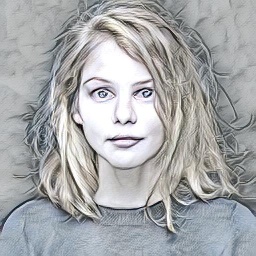}
  \hspace{-4pt}
 \includegraphics[height=0.066\textwidth]{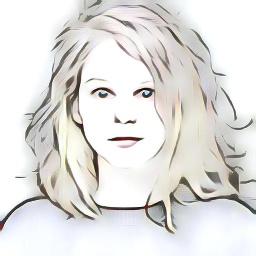}  
\caption{\footnotesize Results on arbitrary new styles. The images in the first column are the content images. Images in the first row are the style images. Images in the second row are the outputs of our DRB-GAN. Images in the third row are the outputs of AdaIN. 
}
\label{fig:new}
\vspace{-4pt} 
\end{figure}

{\noindent\bf Performance on Arbitrary New Styles.}
To clarify, we have evaluated our DRB-GAN model on images in arbitrary new styles. As shown in Figure \ref{fig:new}, our DRB-GAN model generalizes well for these new styles. It achieves better qualitative performance than AdaIN.

\begin{figure*}[ht!]
\centering
 \includegraphics[width=0.108\textwidth,height=0.07\textwidth]{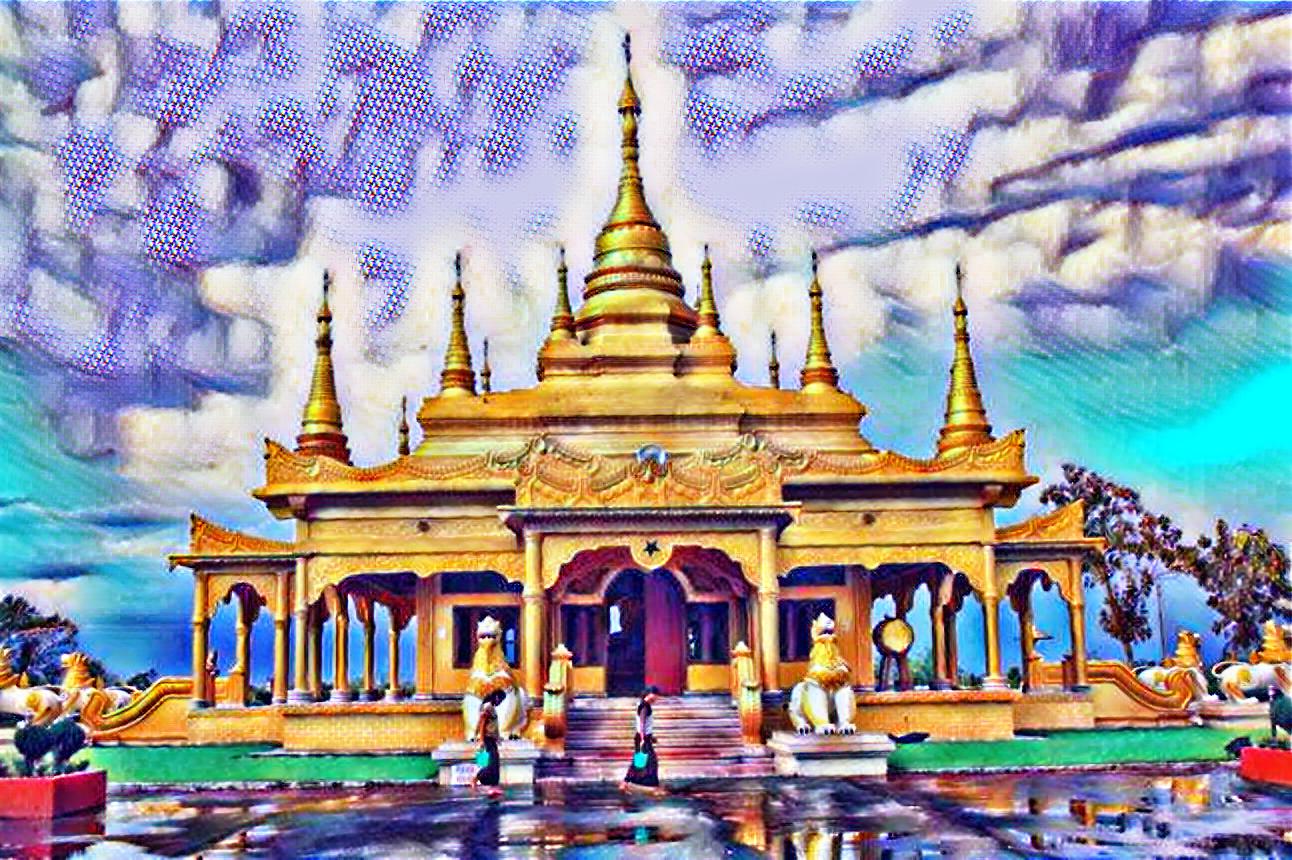}
  \hspace{-4pt}
 \includegraphics[width=0.108\textwidth,height=0.07\textwidth]{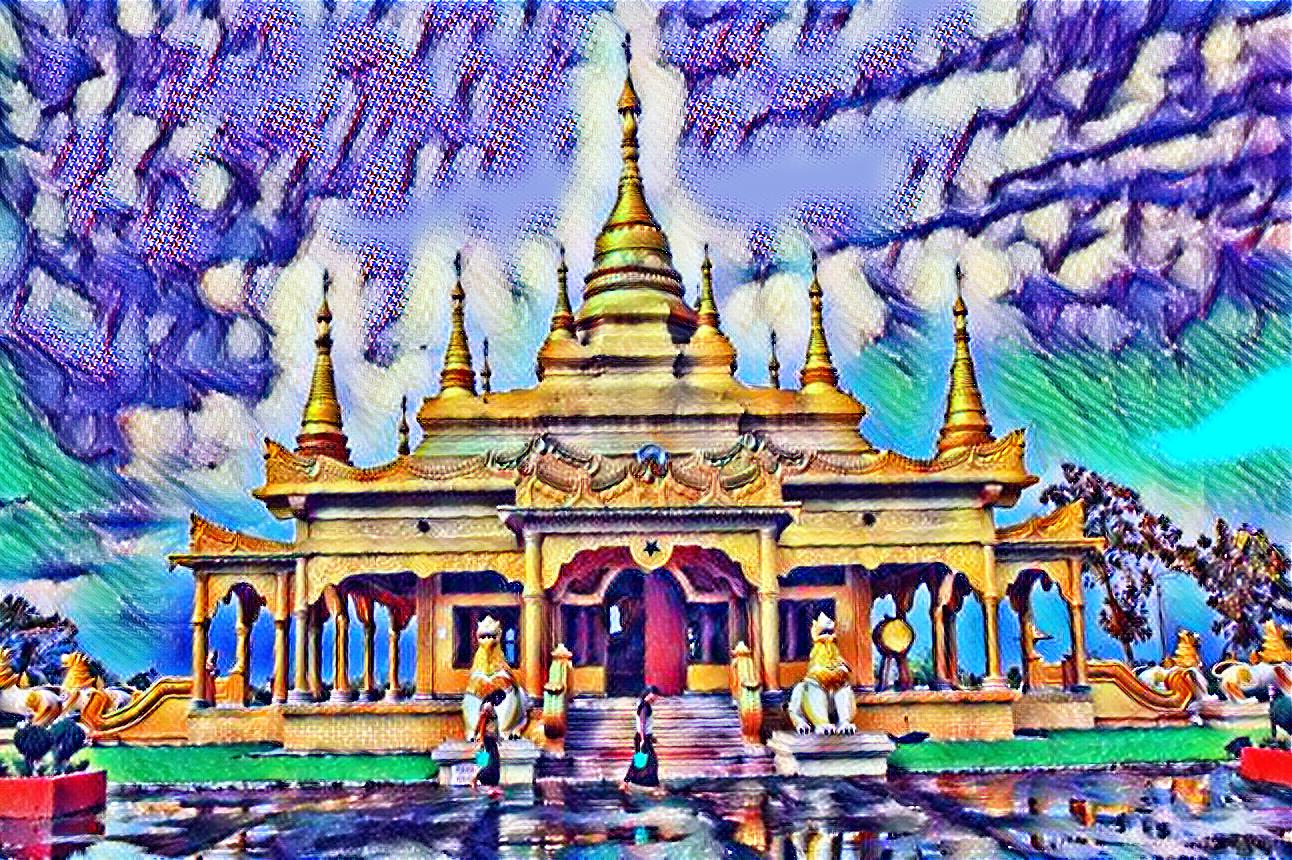}
  \hspace{-4pt}
 \includegraphics[width=0.108\textwidth,height=0.07\textwidth]{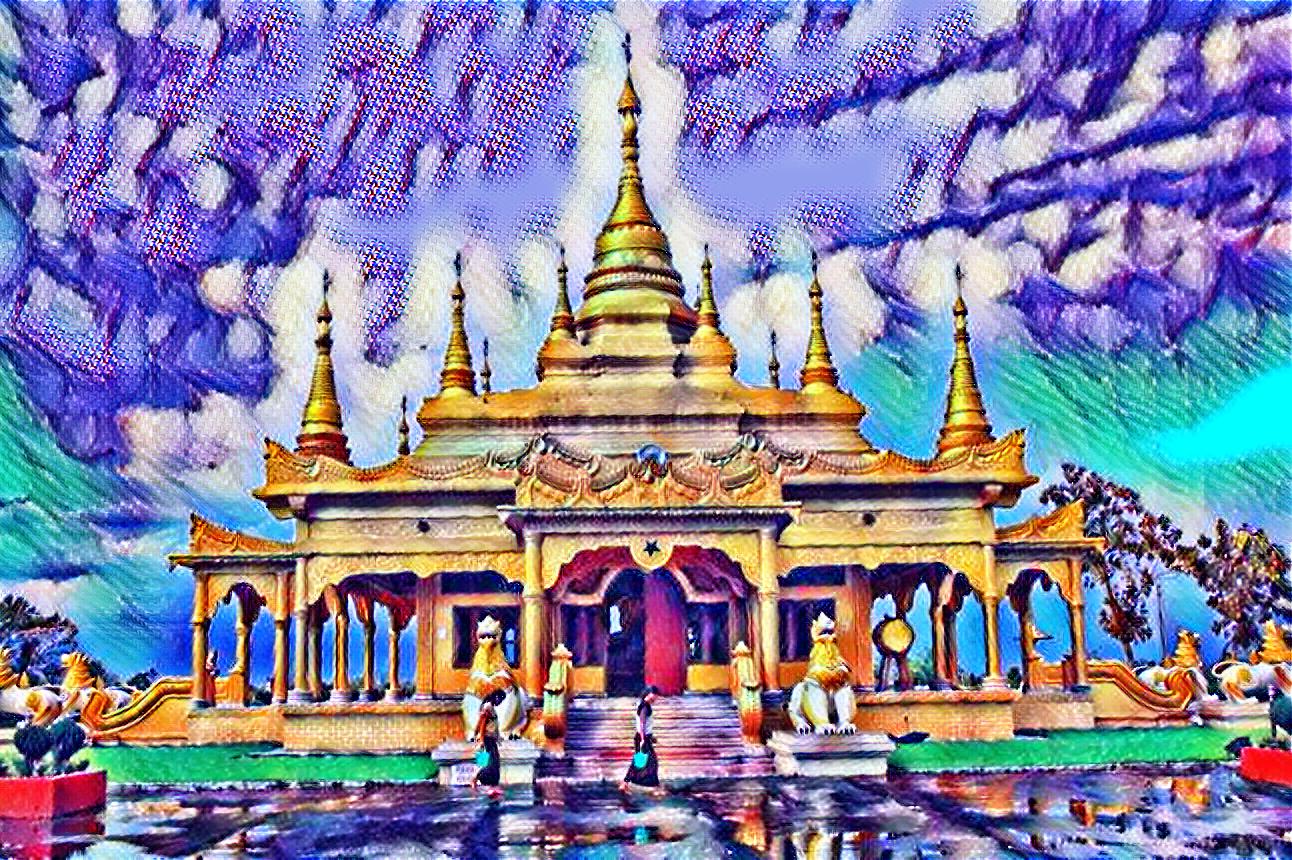}
  \hspace{-3pt}
 \includegraphics[width=0.108\textwidth,height=0.07\textwidth]{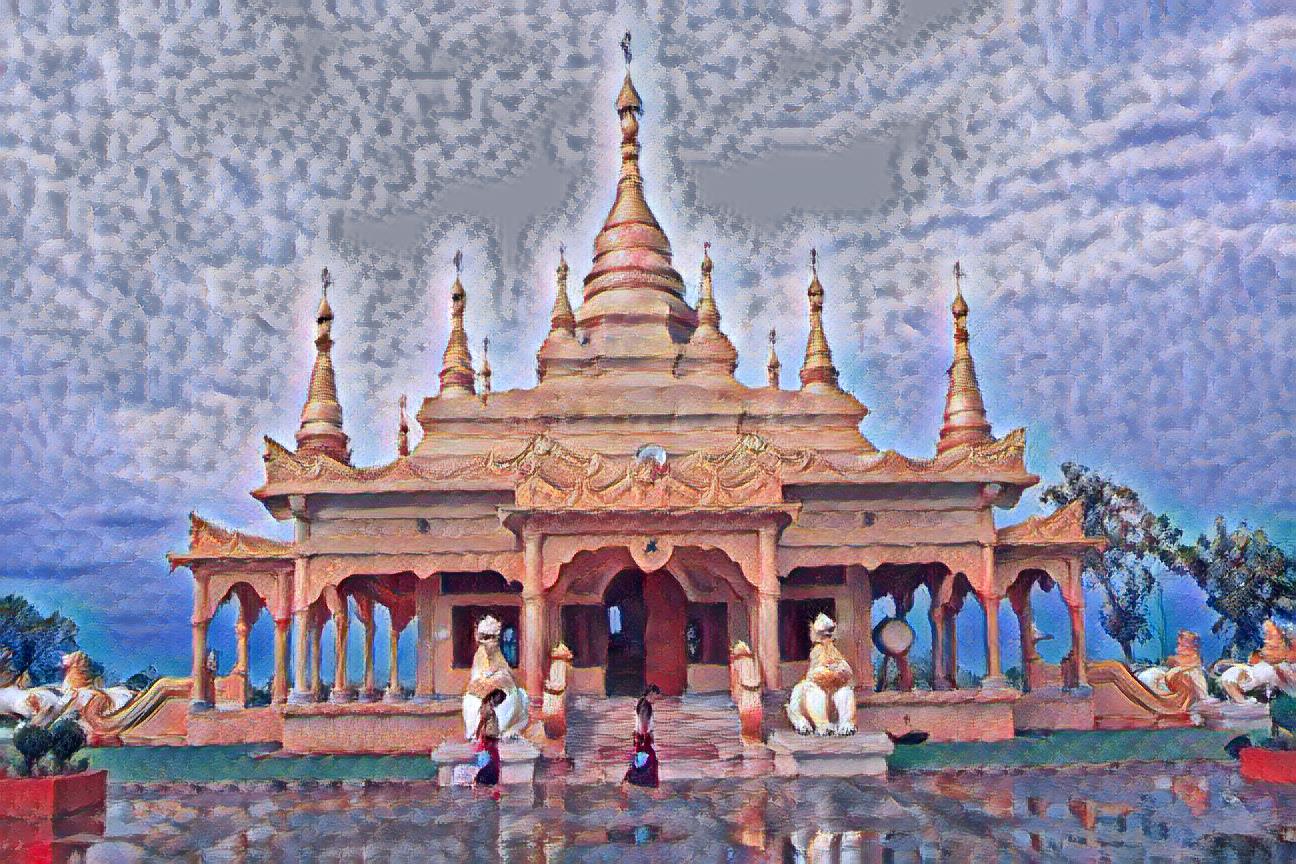} 
 \hspace{-4pt}
 \includegraphics[width=0.108\textwidth,height=0.07\textwidth]{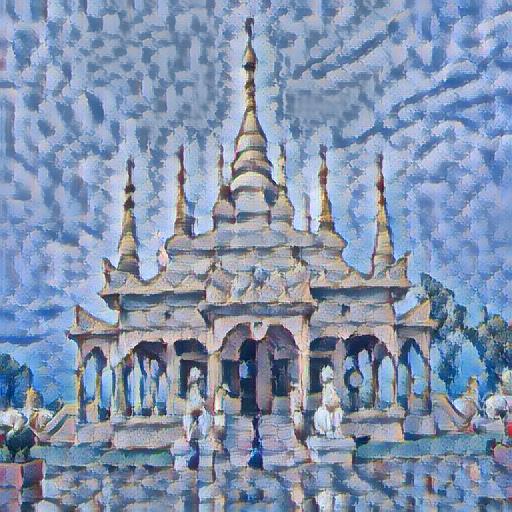}
 \hspace{-4pt}
 \includegraphics[width=0.108\textwidth,height=0.07\textwidth]{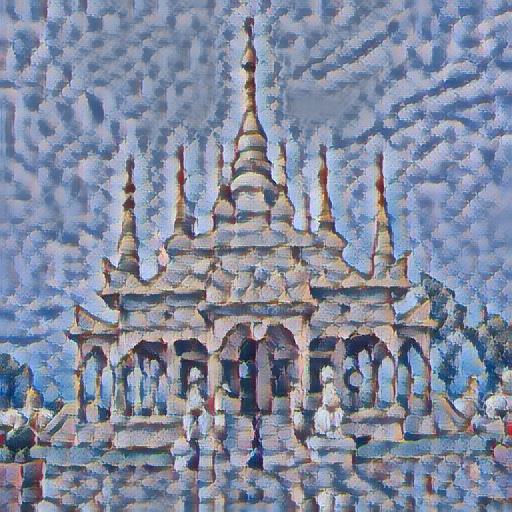}
 \hspace{-3pt}
 \includegraphics[width=0.108\textwidth,height=0.07\textwidth]{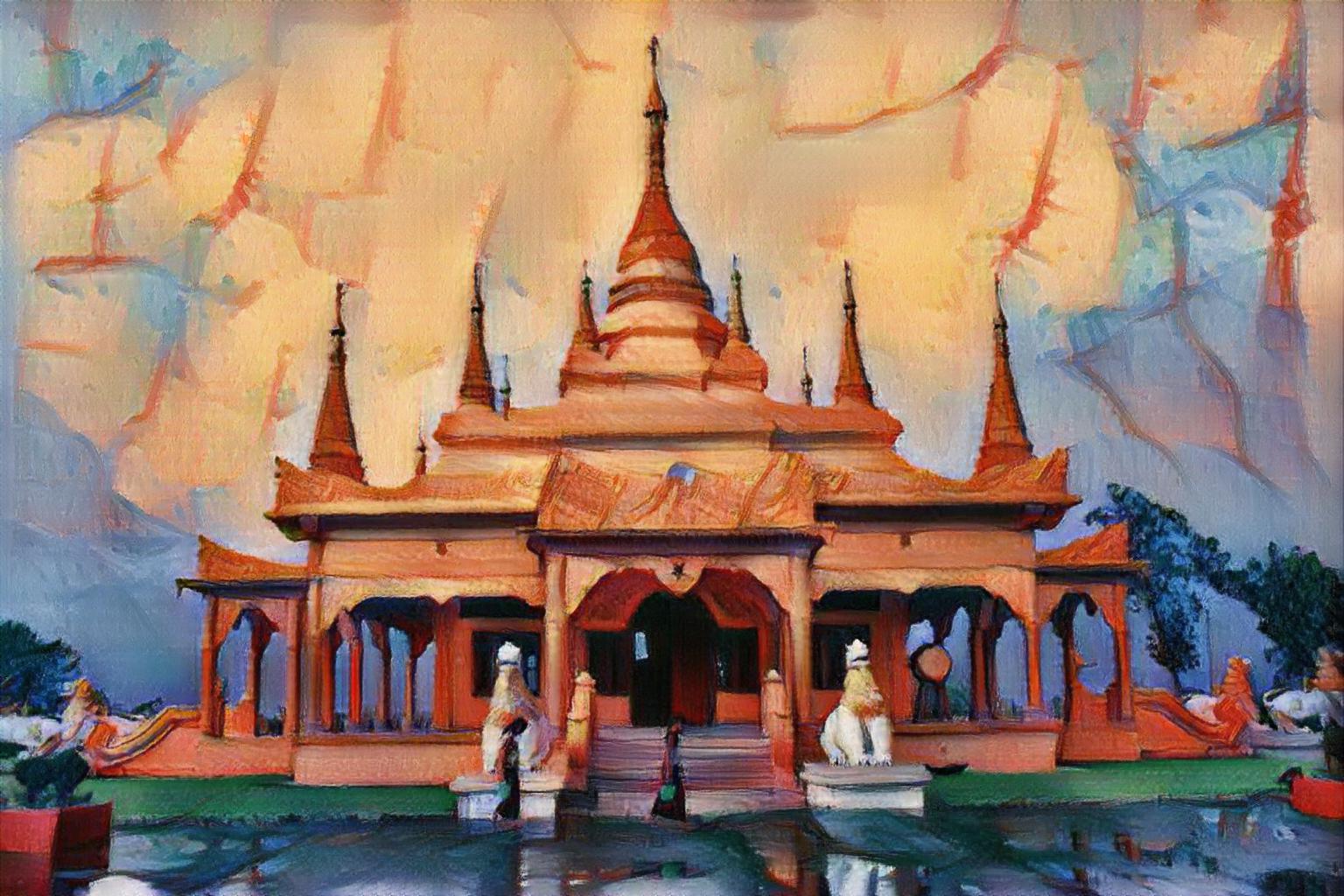}
 \hspace{-4pt}
 \includegraphics[width=0.108\textwidth,height=0.07\textwidth]{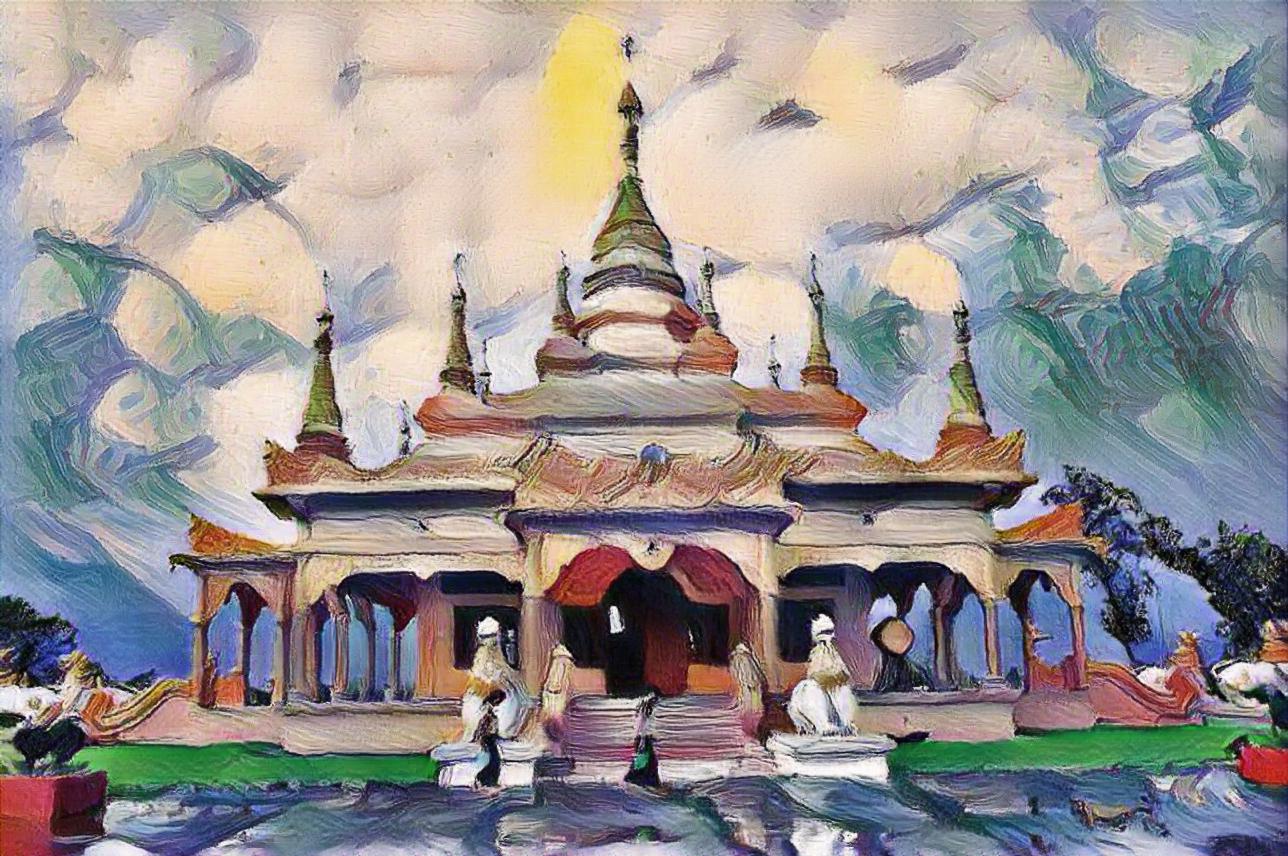} 
  \hspace{-4pt}
 \includegraphics[width=0.108\textwidth,height=0.07\textwidth]{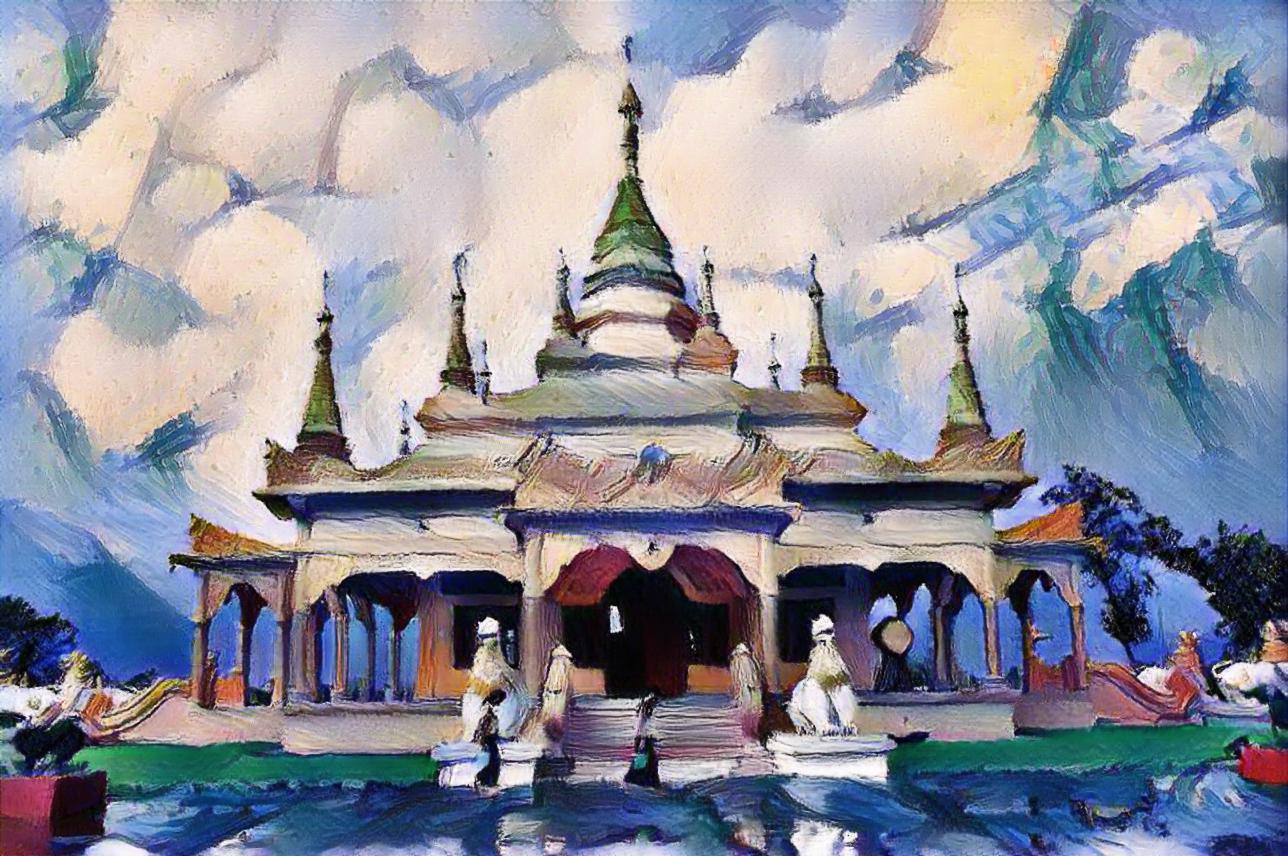}\\
 \includegraphics[width=0.108\textwidth,height=0.07\textwidth]{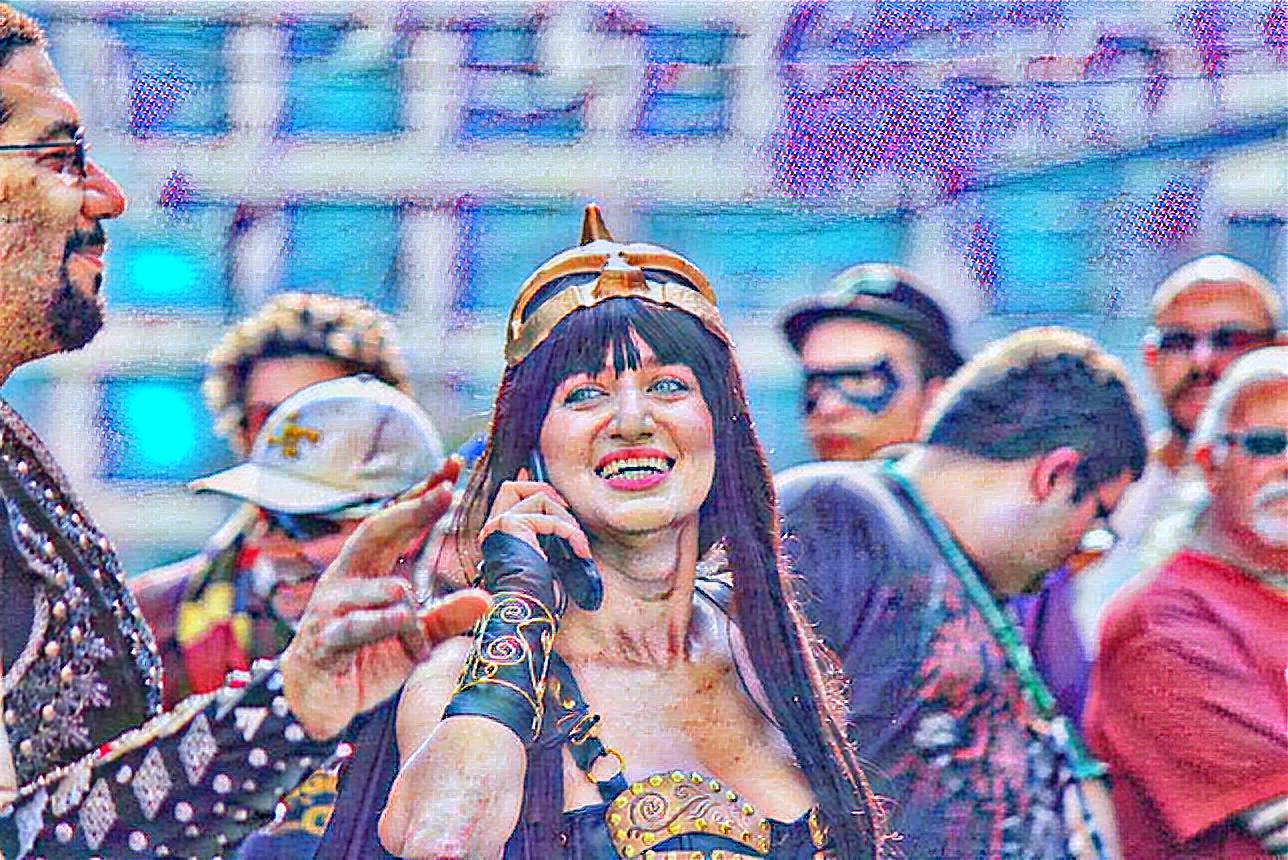}
  \hspace{-4pt}
 \includegraphics[width=0.108\textwidth,height=0.07\textwidth]{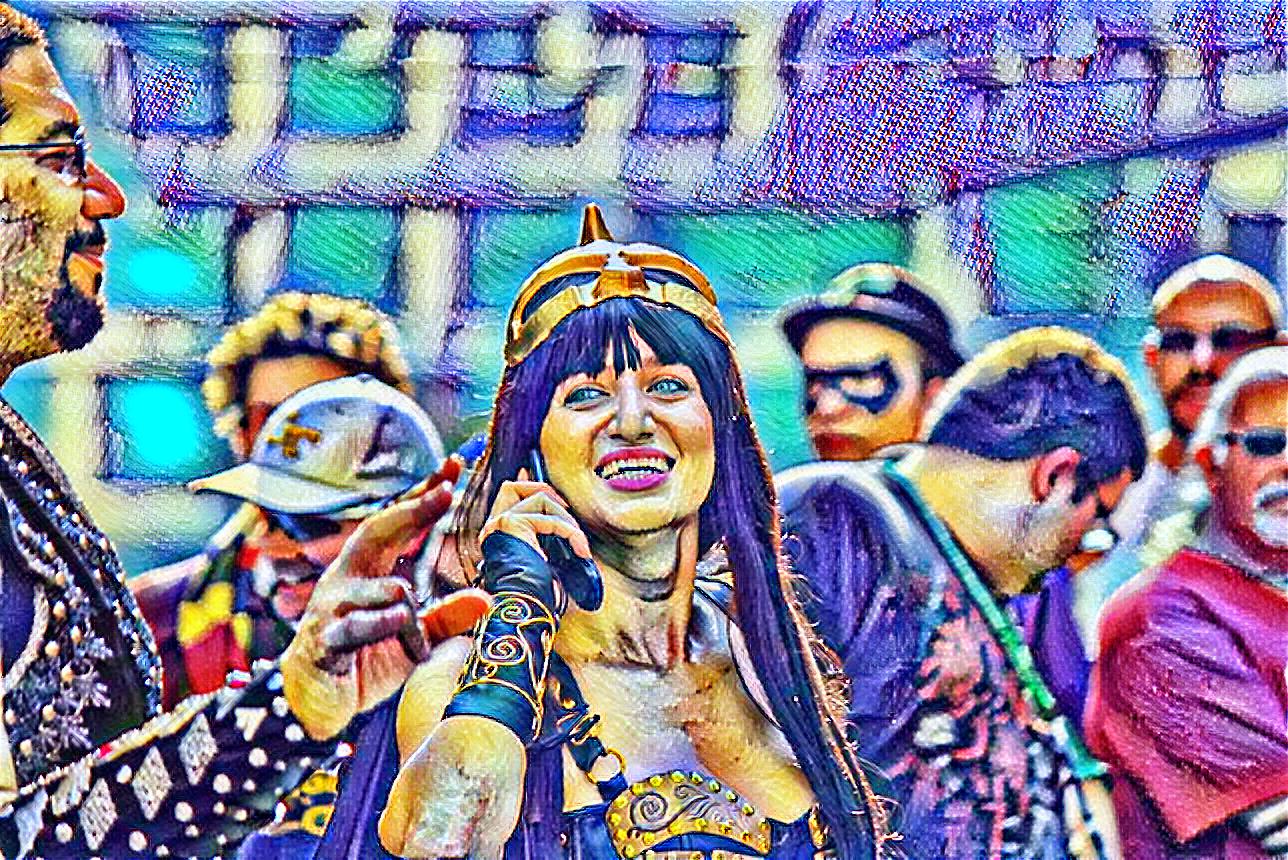}
  \hspace{-4pt}
 \includegraphics[width=0.108\textwidth,height=0.07\textwidth]{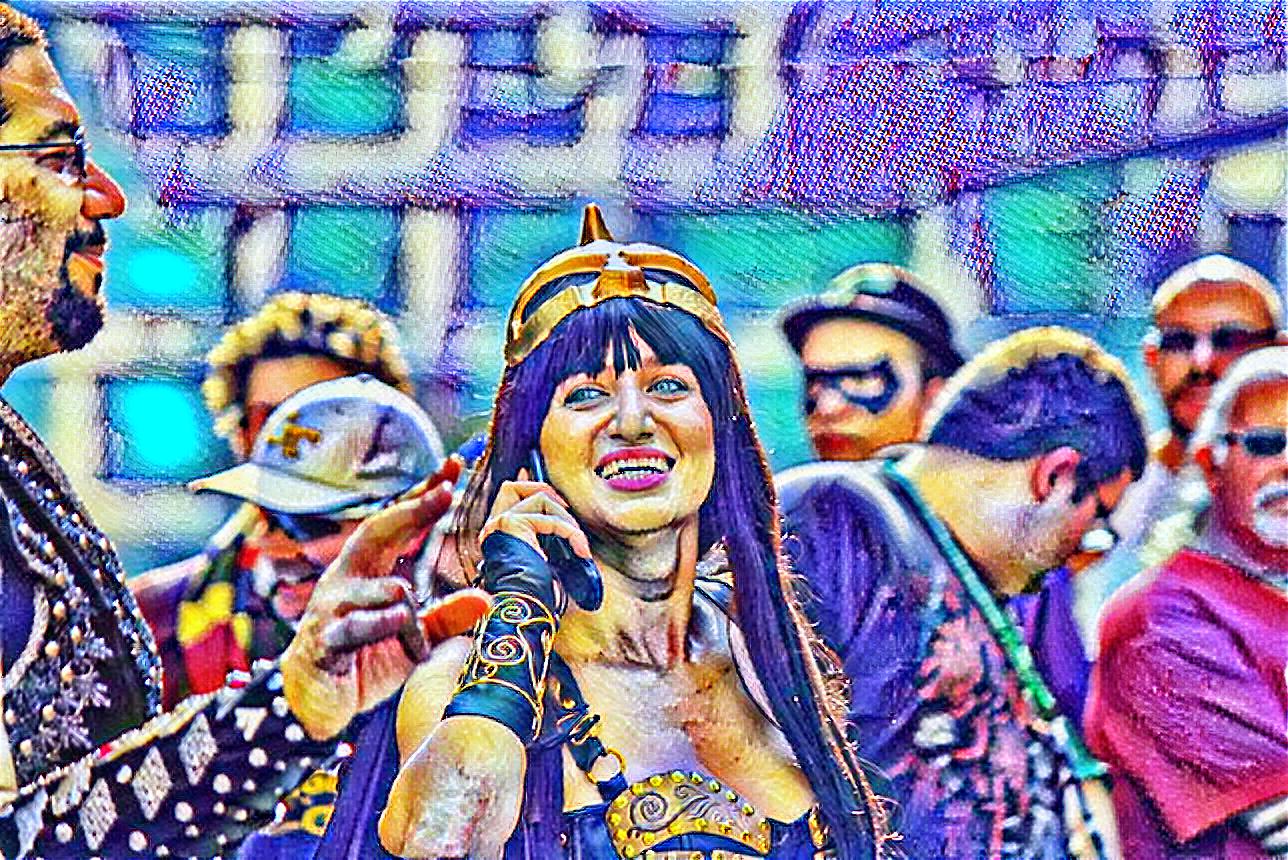}
  \hspace{-3pt}
 \includegraphics[width=0.108\textwidth,height=0.07\textwidth]{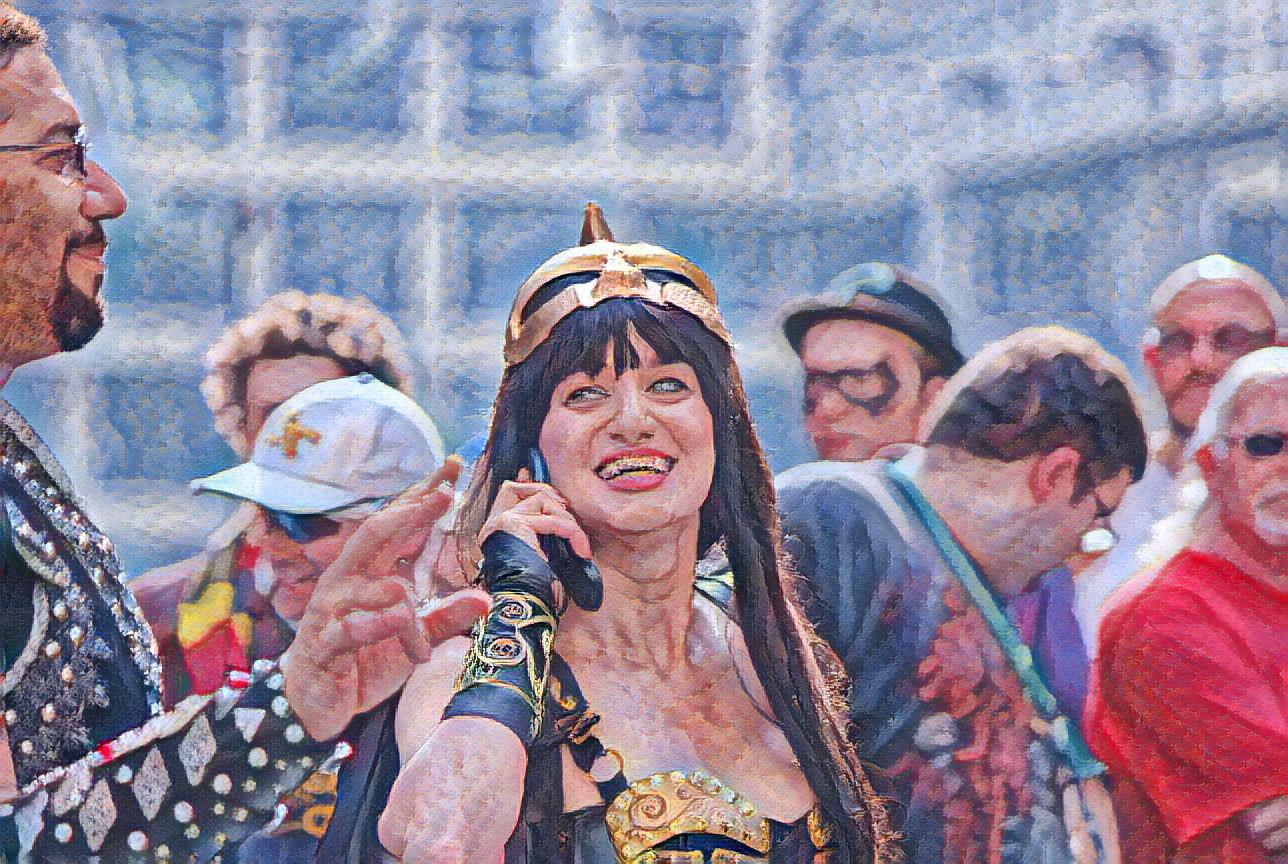} 
 \hspace{-4pt}
 \includegraphics[width=0.108\textwidth,height=0.07\textwidth]{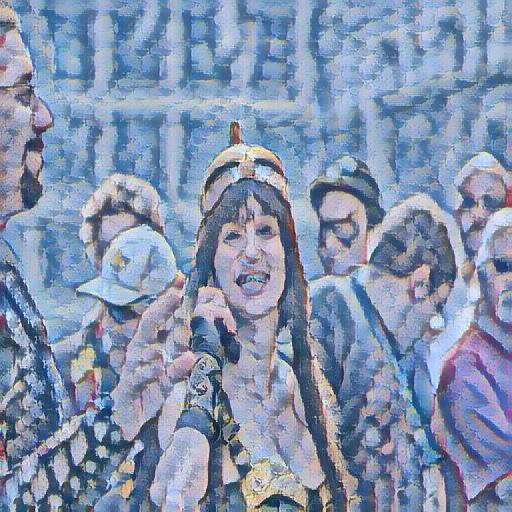}
 \hspace{-4pt}
 \includegraphics[width=0.108\textwidth,height=0.07\textwidth]{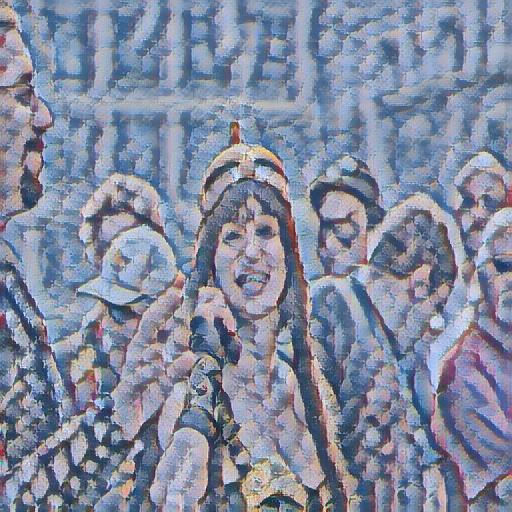}
 \hspace{-3pt}
 \includegraphics[width=0.108\textwidth,height=0.07\textwidth]{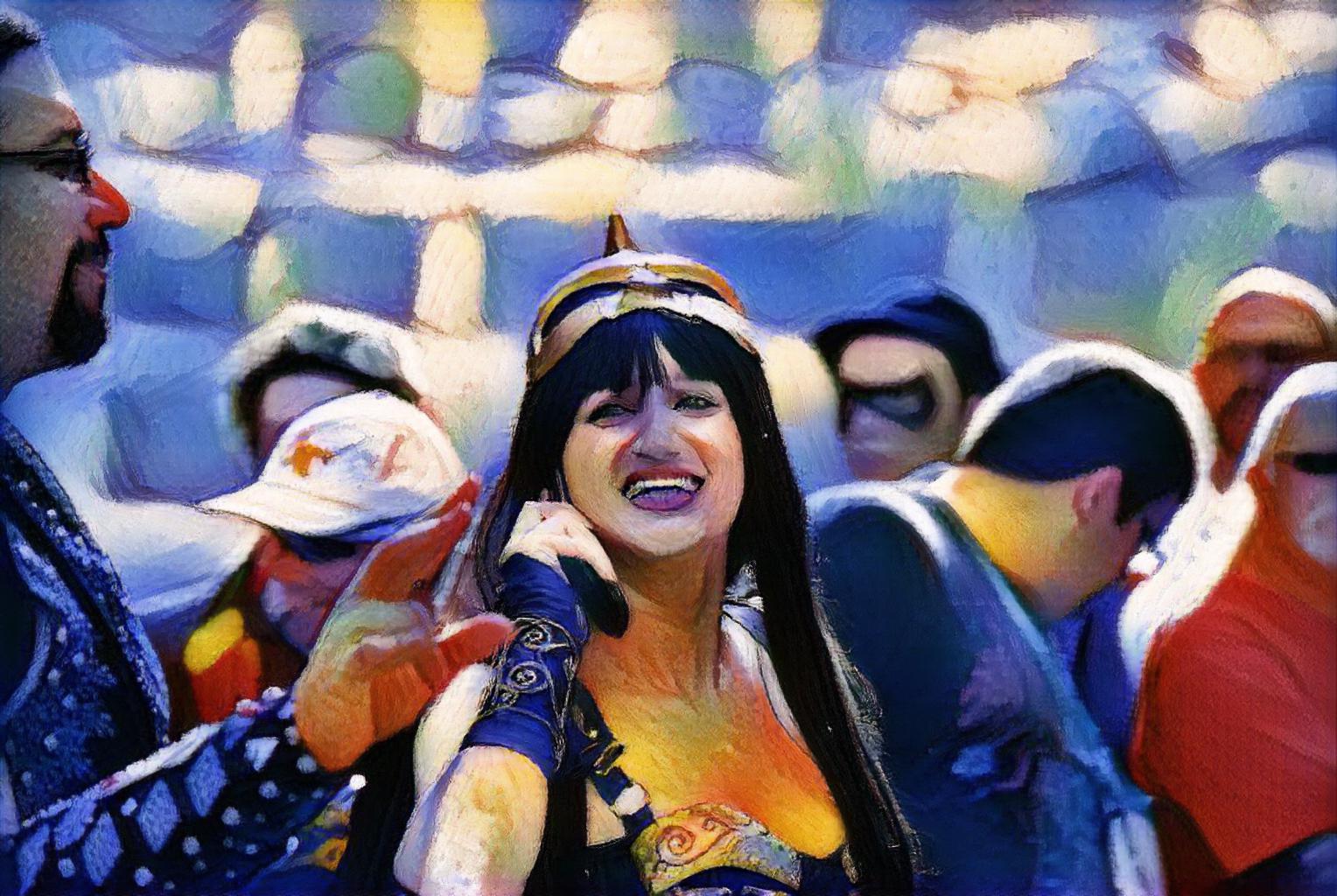}
 \hspace{-4pt}
 \includegraphics[width=0.108\textwidth,height=0.07\textwidth]{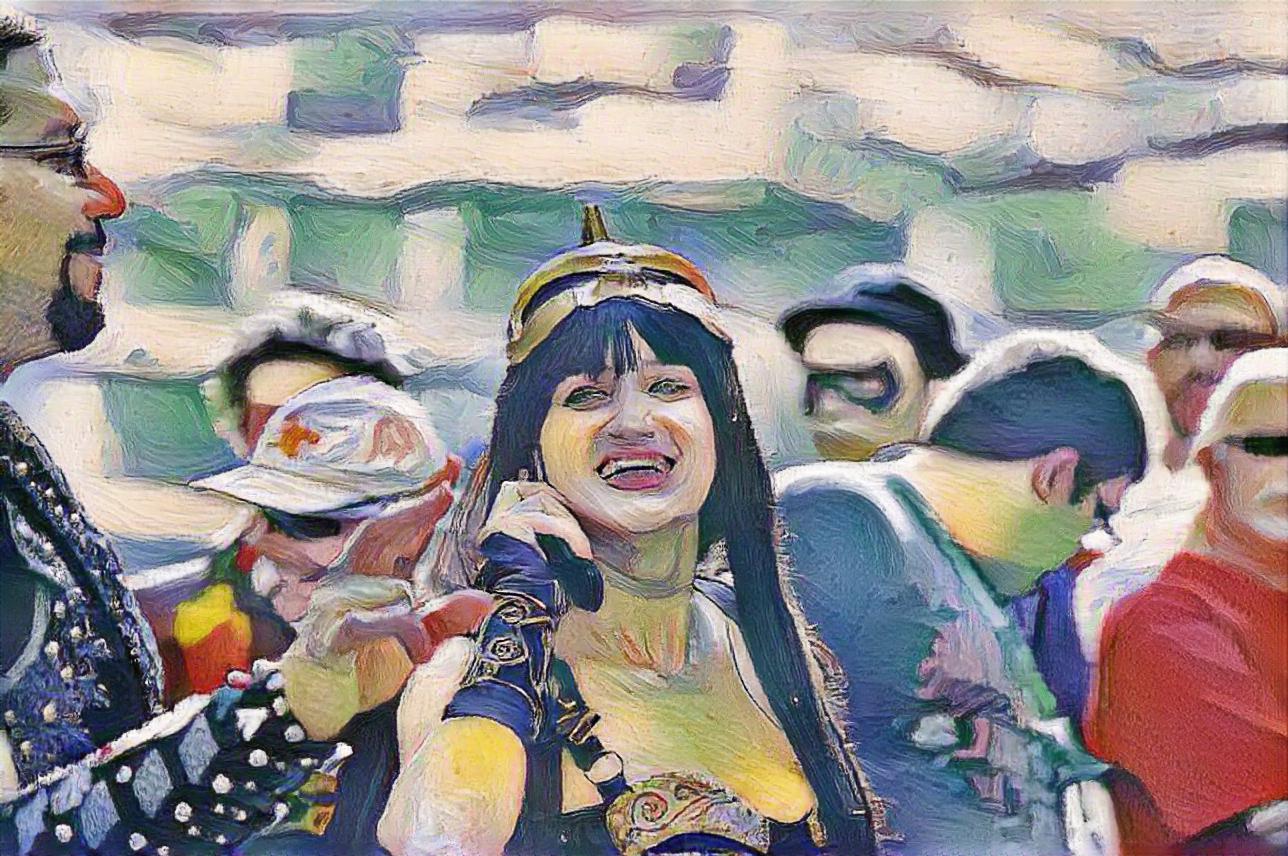} 
  \hspace{-4pt}
 \includegraphics[width=0.108\textwidth,height=0.07\textwidth]{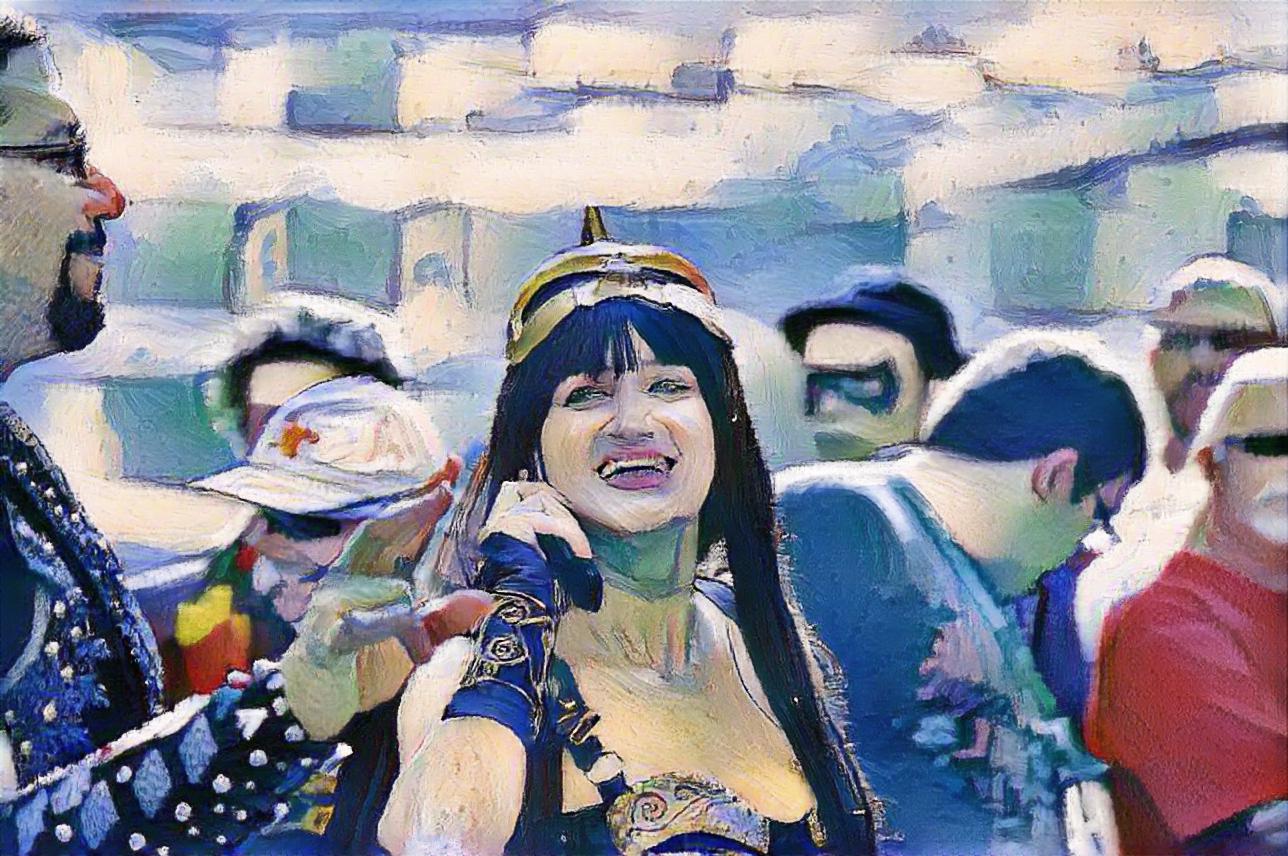}\\
  \vspace{-5.pt}
  \subfigure[]{\includegraphics[width=0.108\textwidth,height=0.07\textwidth]{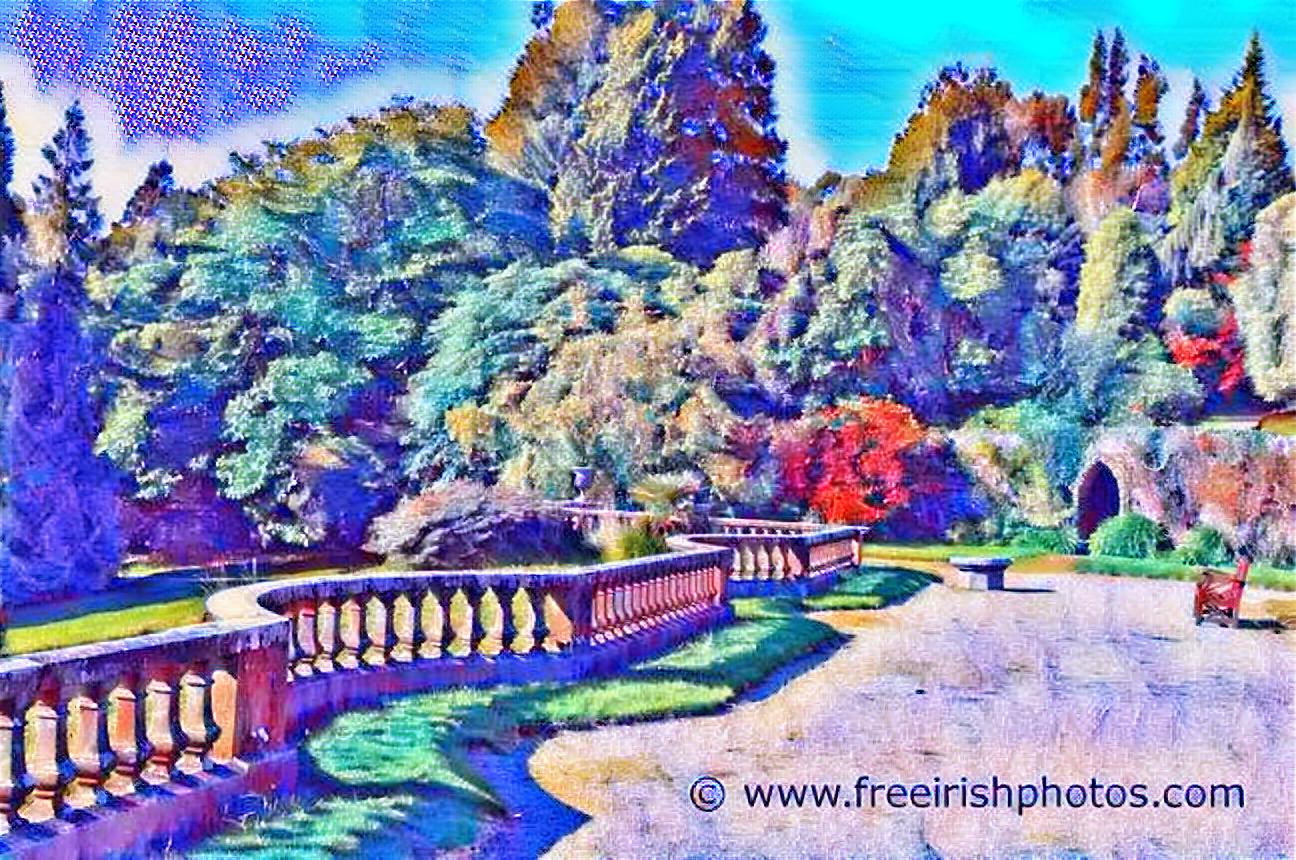}} 
  \hspace{-4pt} 
  \subfigure[]{\includegraphics[width=0.108\textwidth,height=0.07\textwidth]{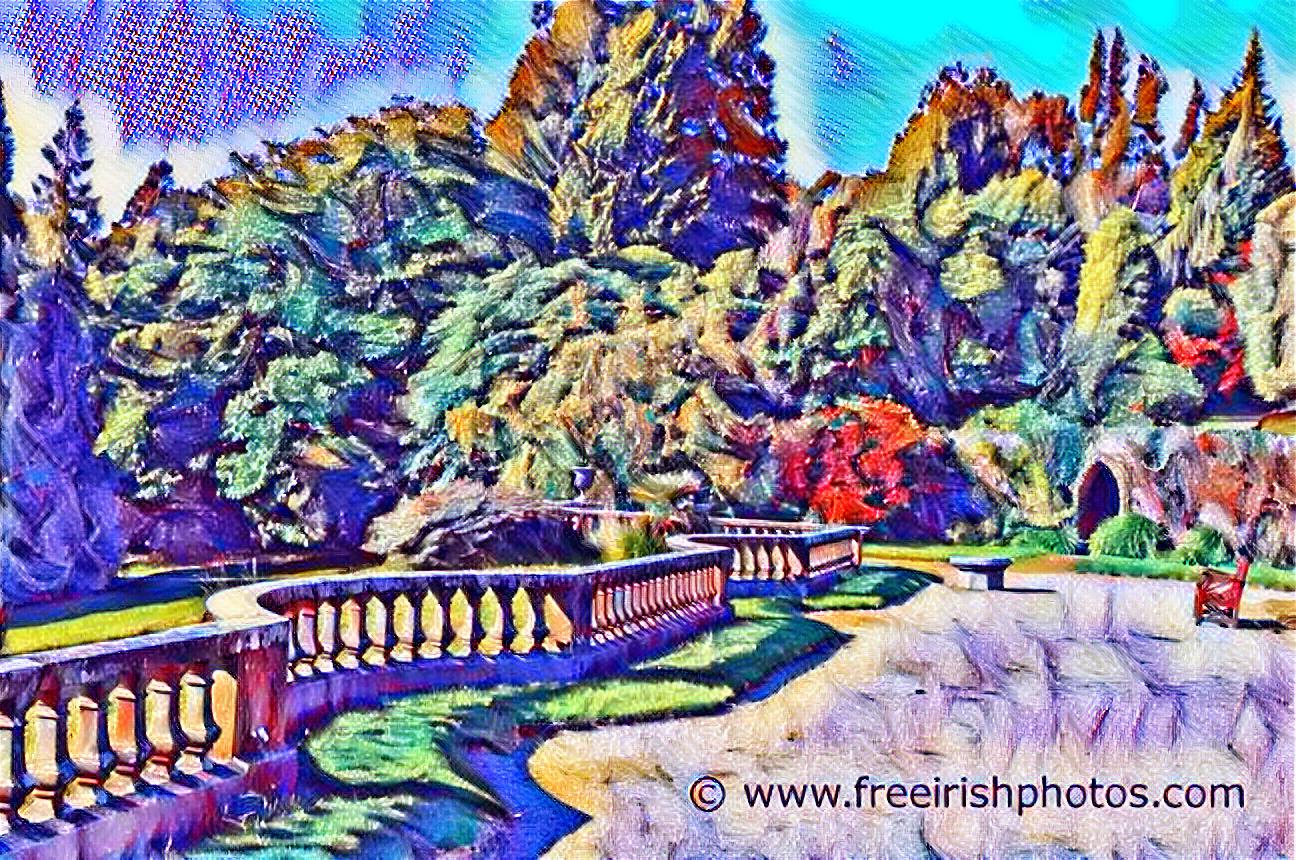}} 
  \hspace{-4pt}  
  \subfigure[]{\includegraphics[width=0.108\textwidth,height=0.07\textwidth]{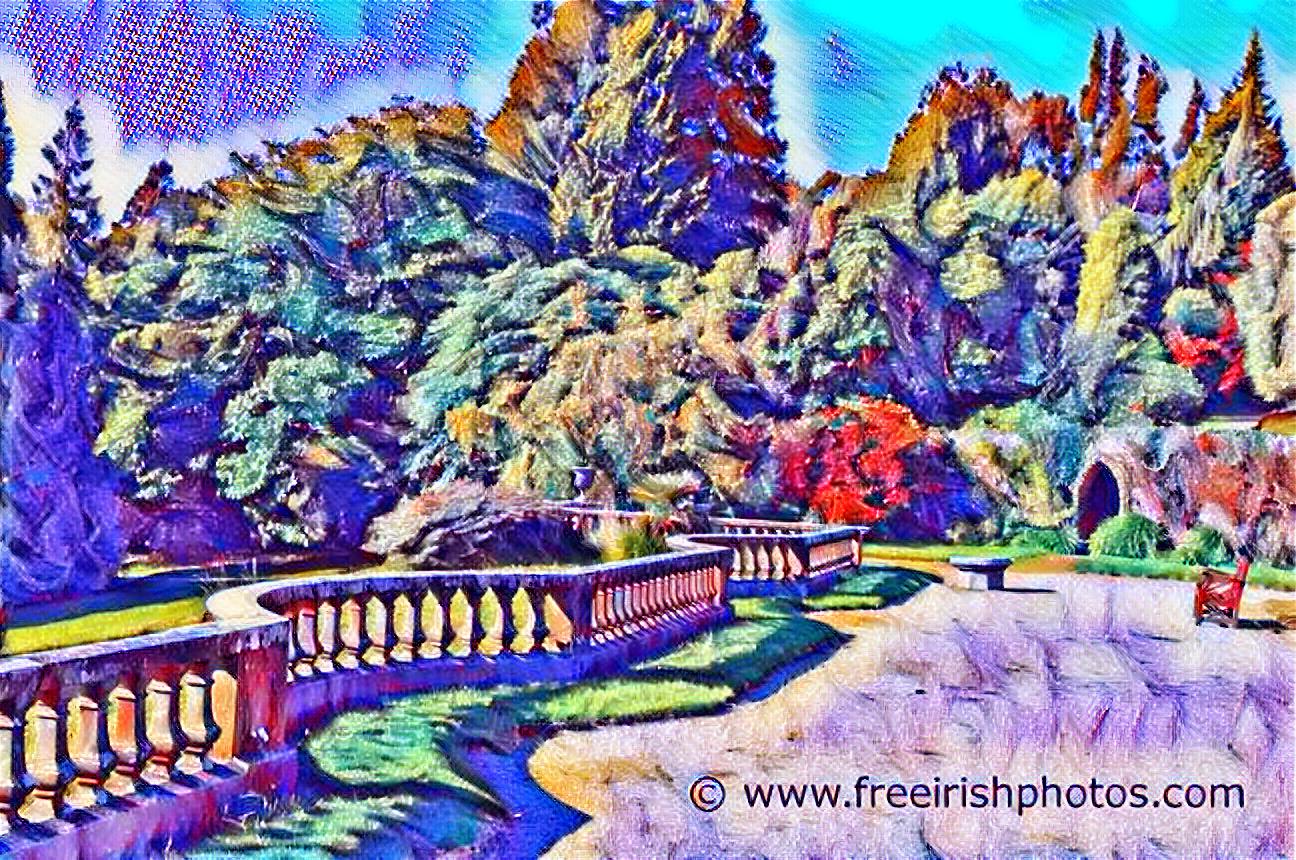}} 
  \hspace{-3pt}
  \subfigure[]{\includegraphics[width=0.108\textwidth,height=0.07\textwidth]{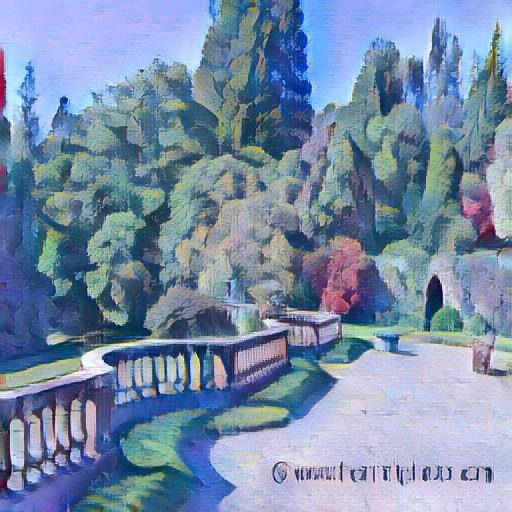}} 
  \hspace{-4pt}  
  \subfigure[]{\includegraphics[width=0.108\textwidth,height=0.07\textwidth]{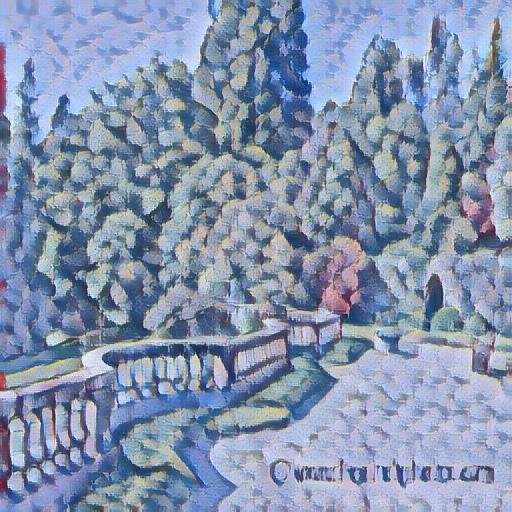}}
  \hspace{-4pt}
  \subfigure[]{\includegraphics[width=0.108\textwidth,height=0.07\textwidth]{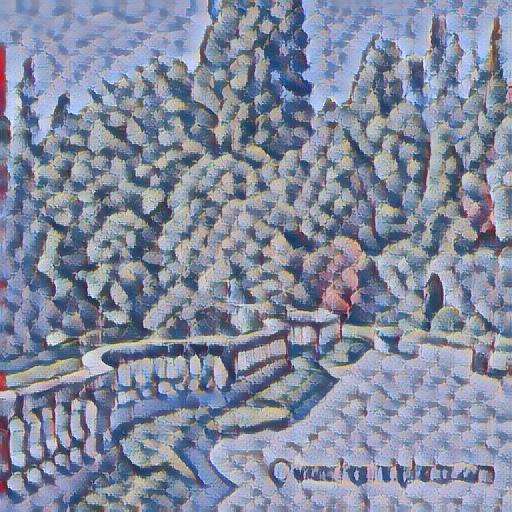}}
  \hspace{-3pt}
  \subfigure[]{\includegraphics[width=0.108\textwidth,height=0.07\textwidth]{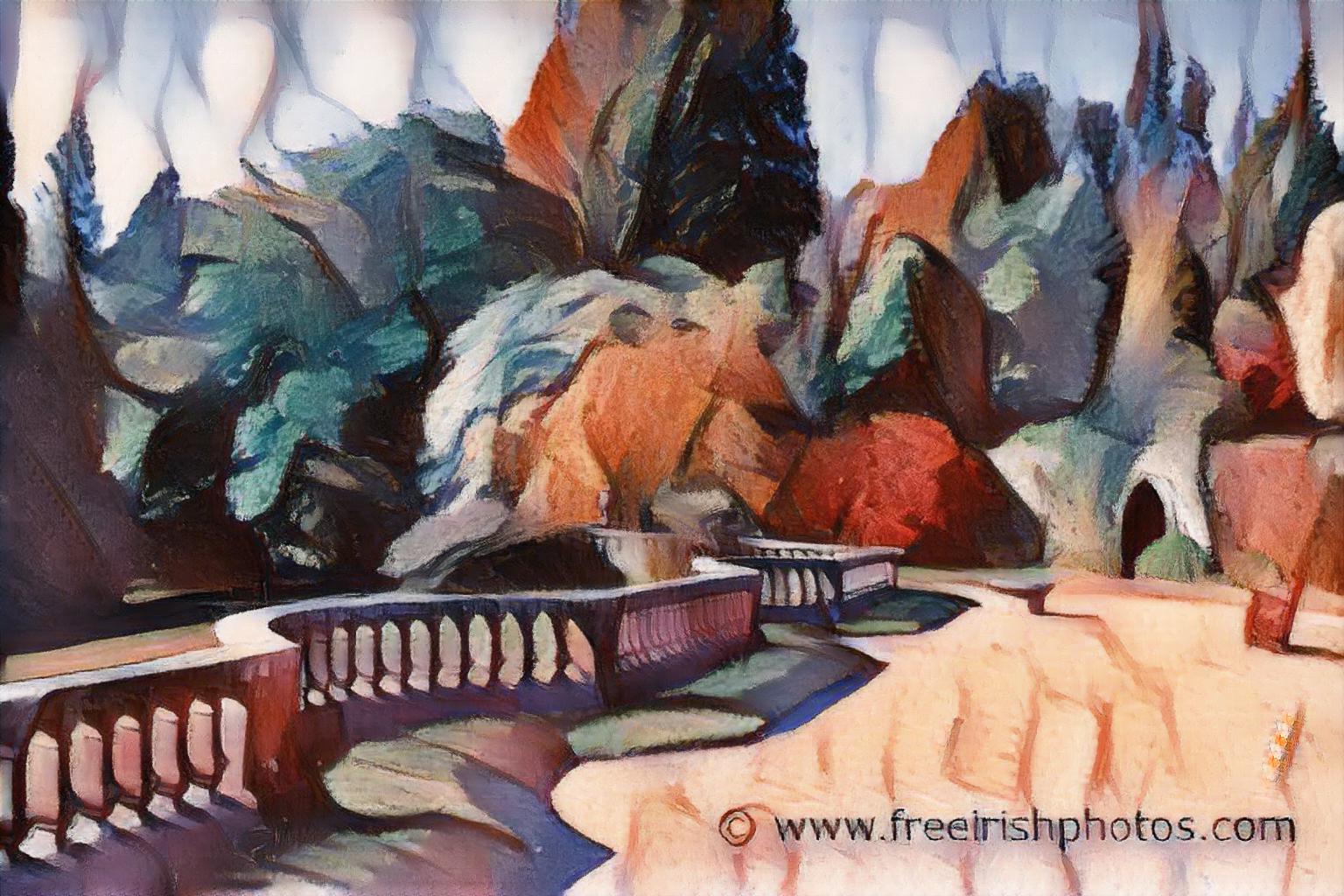}}
  \hspace{-4pt}
  \subfigure[]{\includegraphics[width=0.108\textwidth,height=0.07\textwidth]{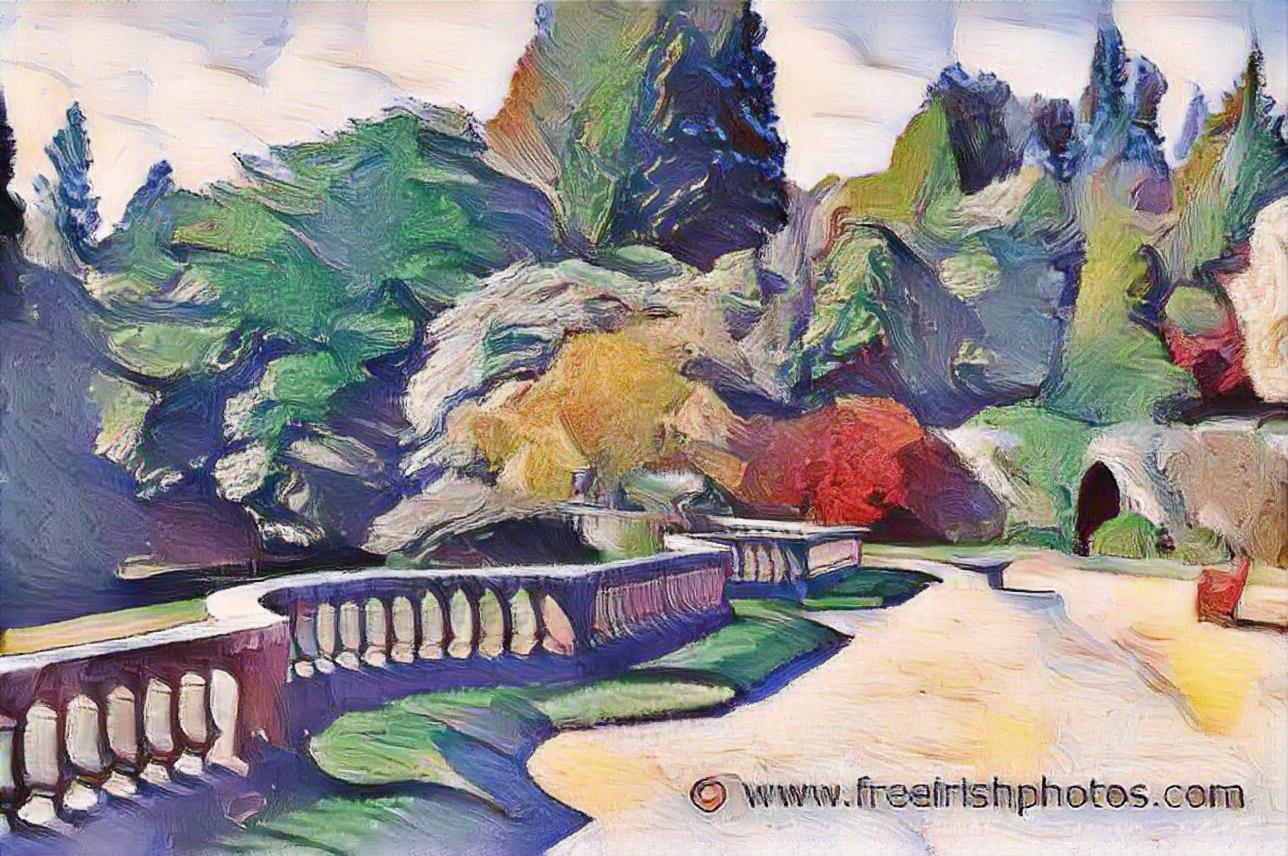}}
  \hspace{-4pt}  
  \subfigure[]{\includegraphics[width=0.108\textwidth,height=0.07\textwidth]{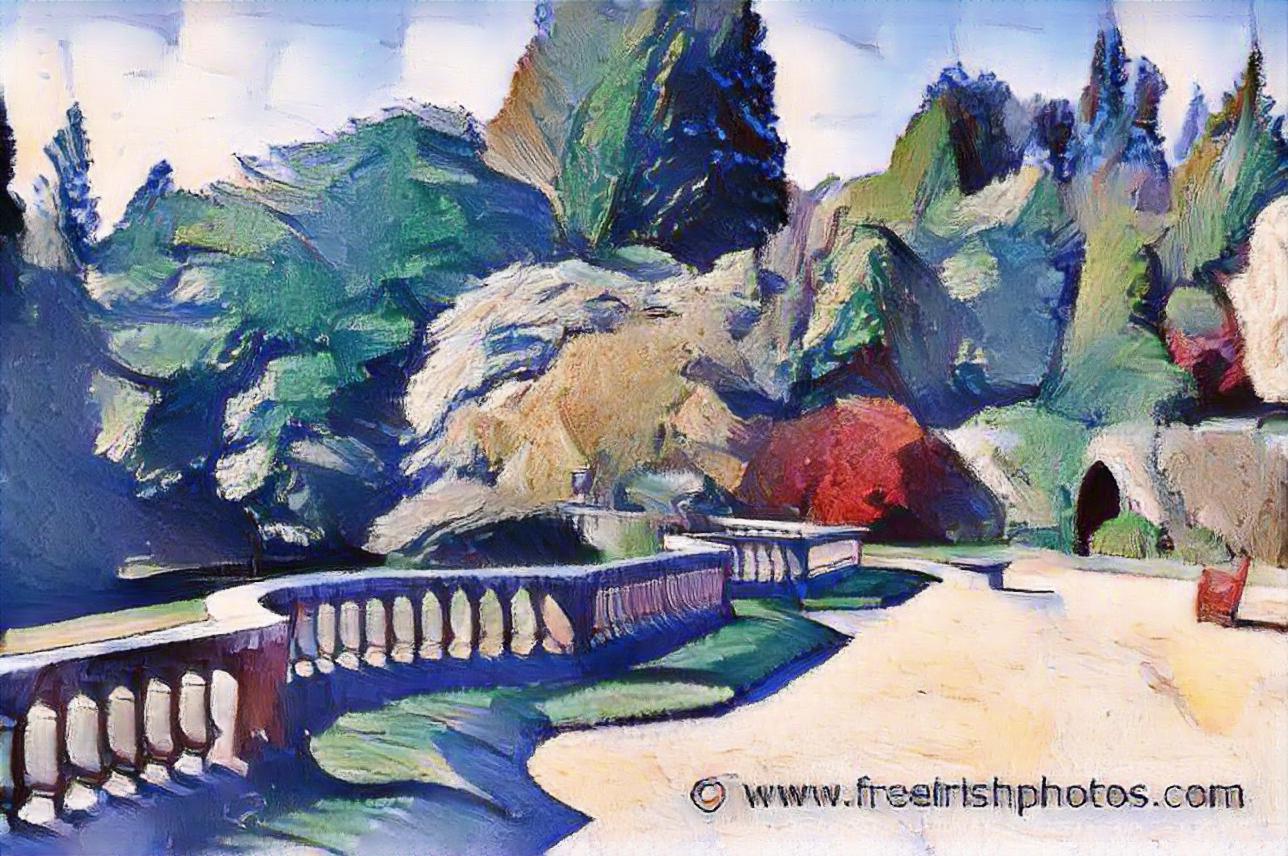}}  \\ 
\caption{\footnotesize Qualitative comparison on the effect of different K values. (a) MetaNet@2, (b) MetaNet@10, (c) MetaNet@20, (d) AdaIN@2, (e) AdaIN@10, (f) AdaIN@20, (g) DRB-GAN@2, (h)  DRB-GAN@10, (i)  DRB-GAN@20.
}
\label{fig:com}
\end{figure*}

{\noindent\bf Model Weighted Averaging.}
As shown in Figure \ref{fig:com}, our DRB-GAN model obtains obvious benefits when adding more individual style within one artistic domain, while other arbitrary style transfer method (AdaIN) focus on the single image style. Notably, averaging weights of AdaIN leads to worse performance as the number of used style images increases.

\begin{figure*}[t]
\centering
\includegraphics[width=0.96\textwidth]{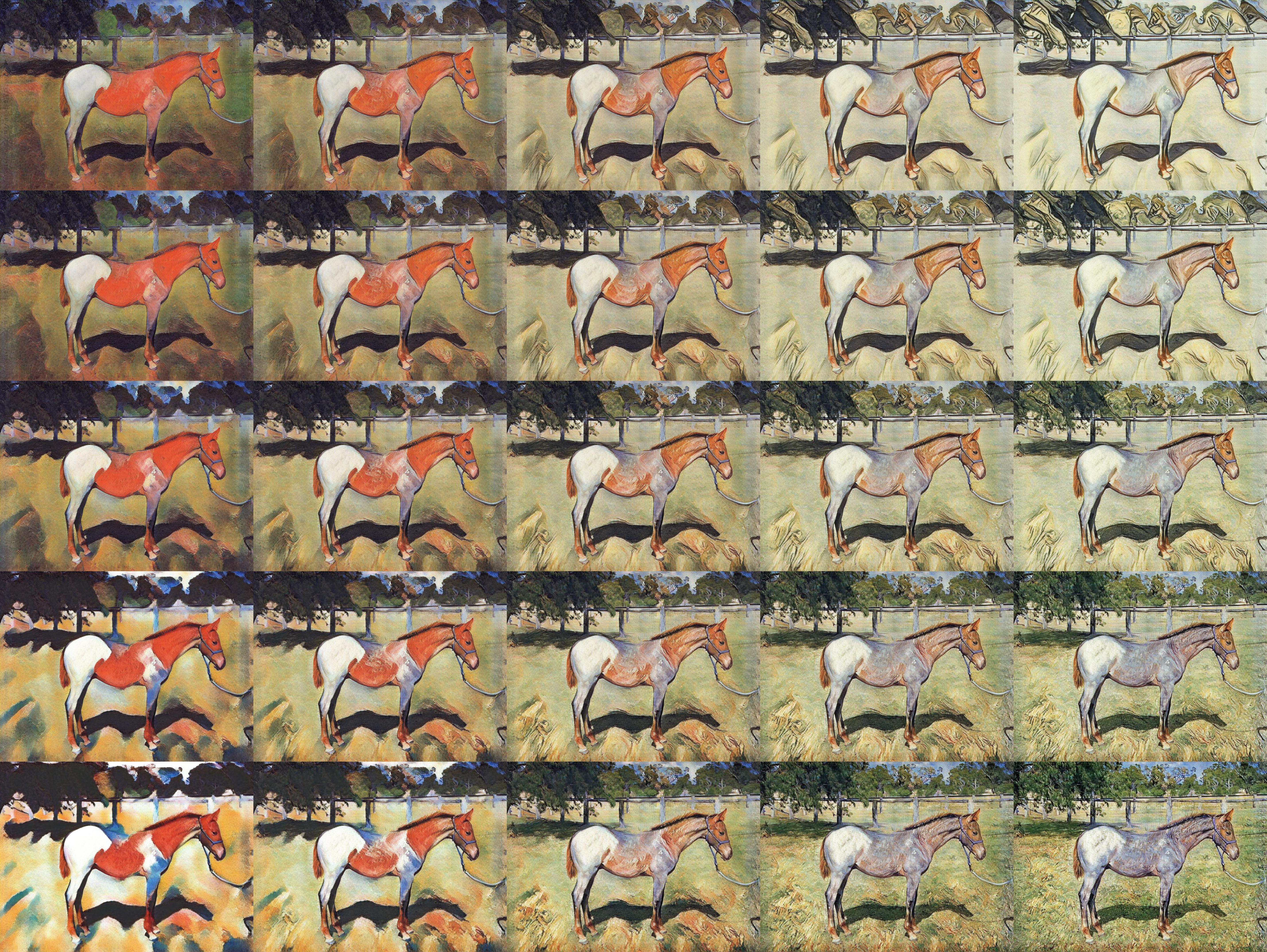}
\includegraphics[width=0.96\textwidth]{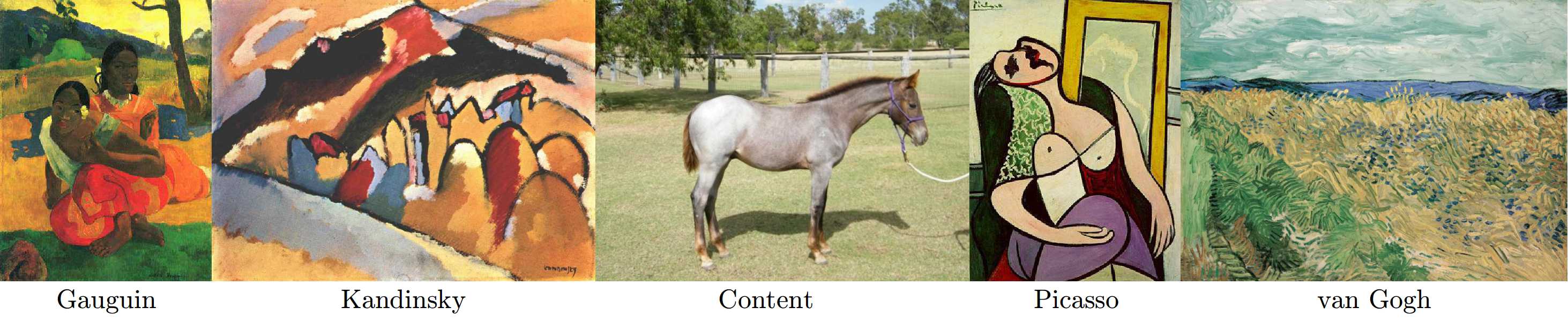}
\caption{Four-way style interpolation across multiple domains. By interpolate the generated weights, we can interpolate between arbitrary styles. The style and content images are listed in the last row for reference.}
\label{horse}
\vspace{-0.4cm} 
\end{figure*}

{\noindent\bf Four-way Style Interpolation.}
To further illustrate the manifold structure, a four-way style interpolation is shown in Figure \ref{horse}. The style interpolation is continuous and smooth, and therefore linearly combine the generated parameters is an efficient way to achieve the style interpolation.

\begin{figure*}[bh!]
\centering
\includegraphics[width=1.0\textwidth]{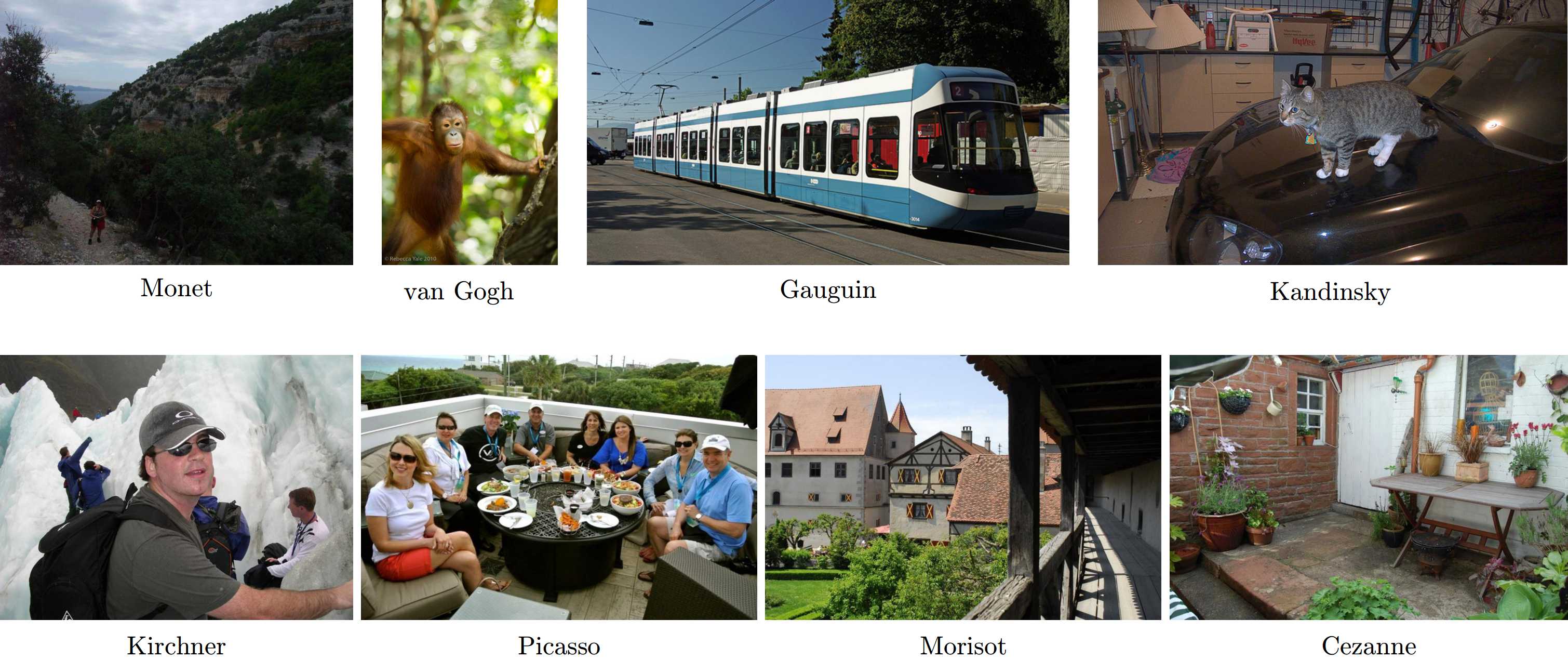}
\caption{The content images used for image style transfer. The text at the bottom indicates the target style. The smaller edge of the images will be resized to $1280$ pixel for high-resolution image generation.}
\label{full}
\vspace{-0.4cm} 
\end{figure*}

\begin{figure*}[t]
\centering
\includegraphics[width=1.0\textwidth]{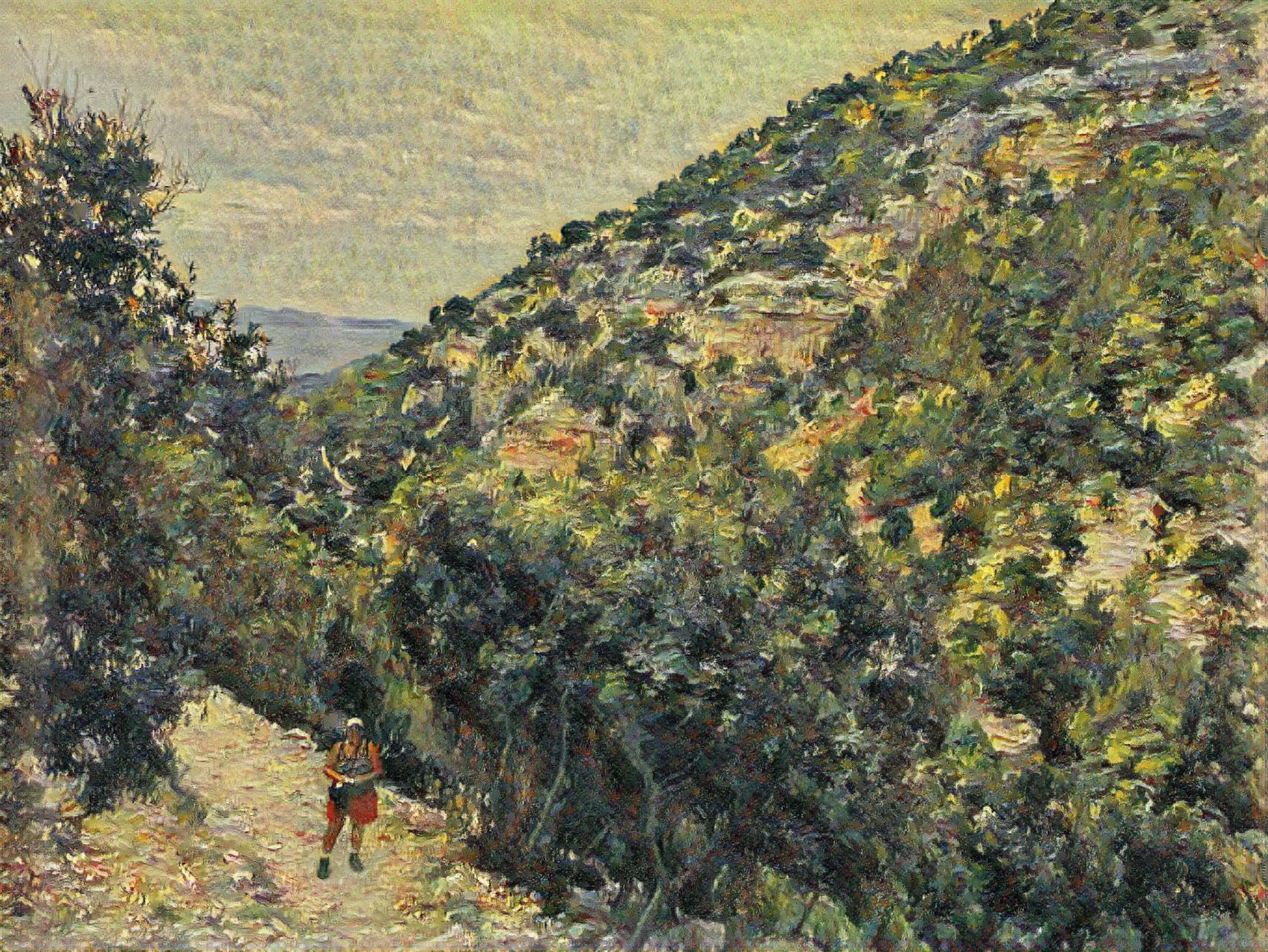}
\caption{High-resolution image generated by our model. The color variations, scale and stroke size are visible. (Monet).}
\label{full1}
\vspace{-0.4cm} 
\end{figure*}

\begin{figure*}[t]
\centering
\includegraphics[width=0.9\textwidth]{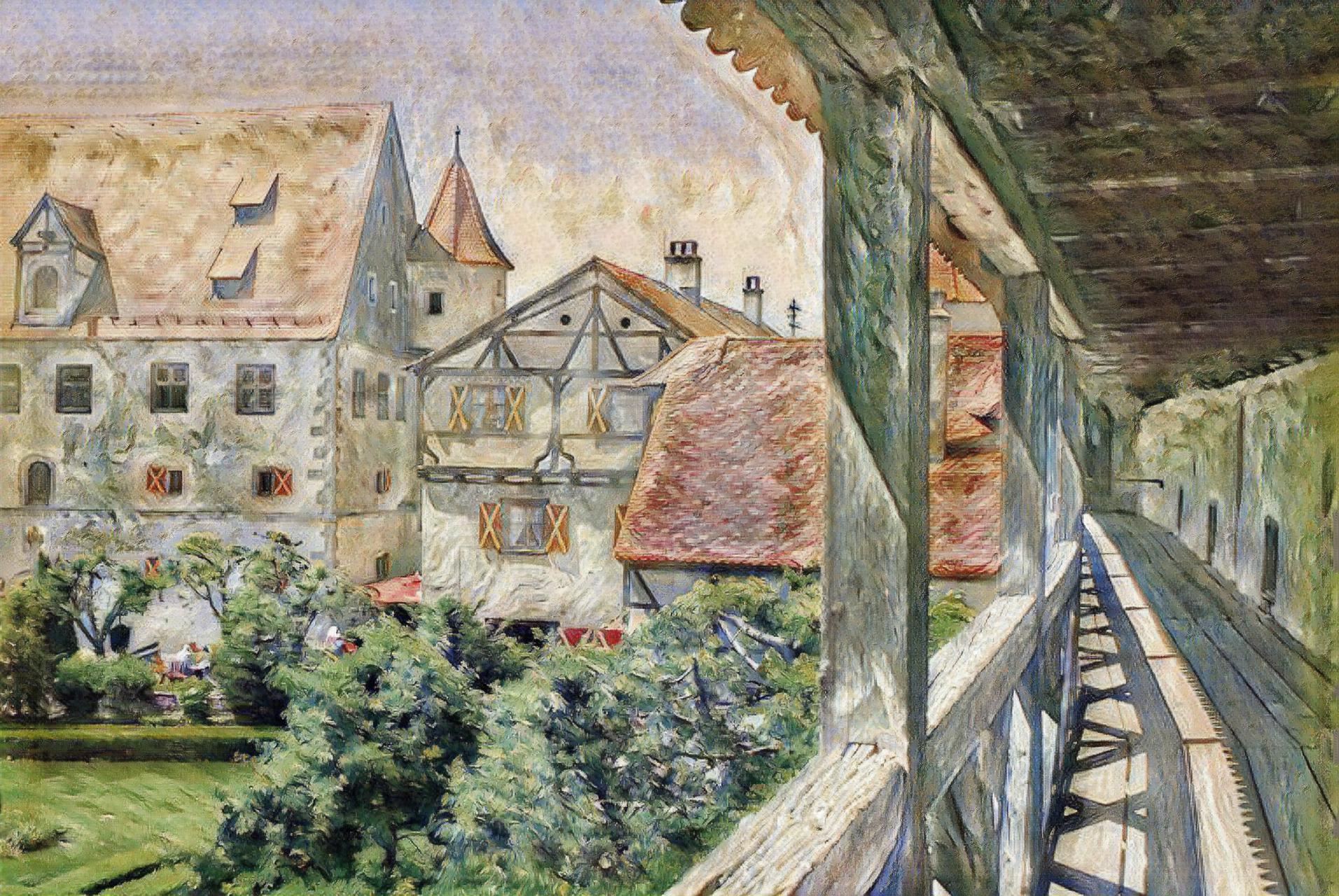}
\caption{High-resolution image generated by our model. The color variations, scale and stroke size are visible. (Morisot)}
\label{full2}
\vspace{-0.4cm} 
\end{figure*}

\begin{figure*}[t]
\centering
\includegraphics[width=0.9\textwidth]{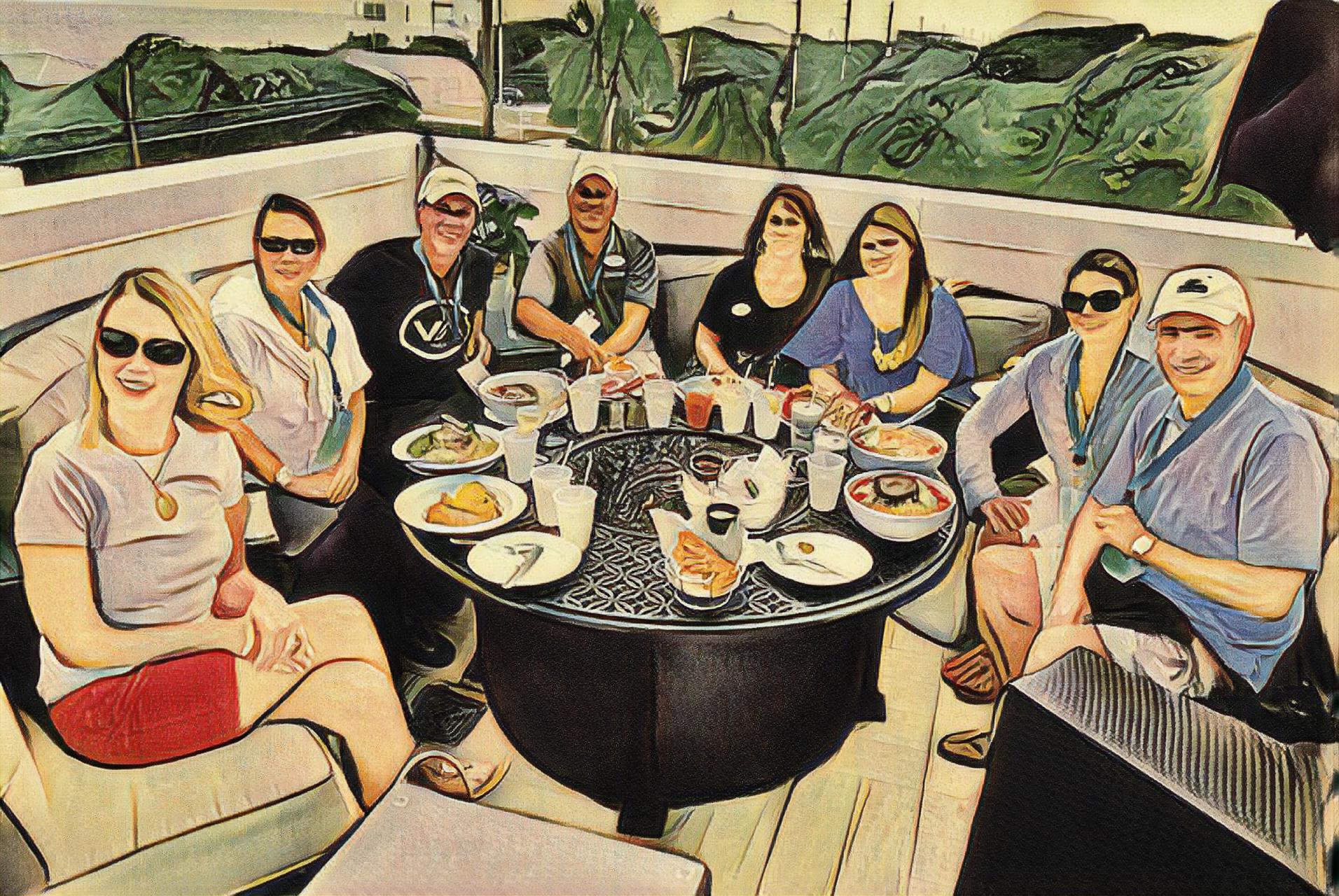}
\caption{High-resolution image generated by our model. The color variations, scale and stroke size are visible. (Picasso)}
\label{full3}
\vspace{-0.4cm} 
\end{figure*}

\begin{figure*}[t]
\centering
\includegraphics[width=0.9\textwidth]{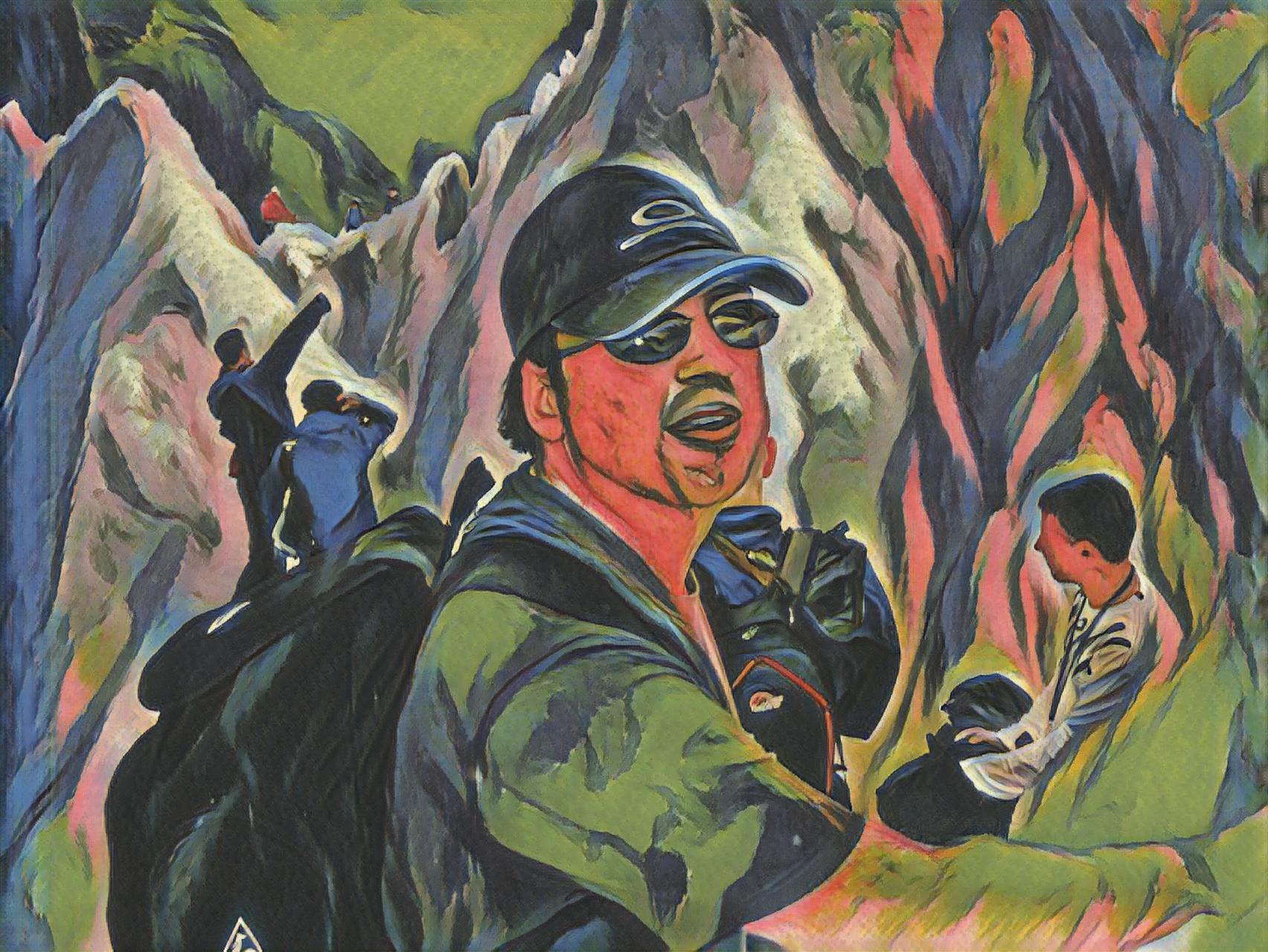}
\caption{High-resolution image generated by our model. The color variations, scale and stroke size are visible. (Kirchner)}
\label{full4}
\vspace{-0.4cm} 
\end{figure*}

\begin{figure*}[t]
\centering
\includegraphics[width=0.9\textwidth]{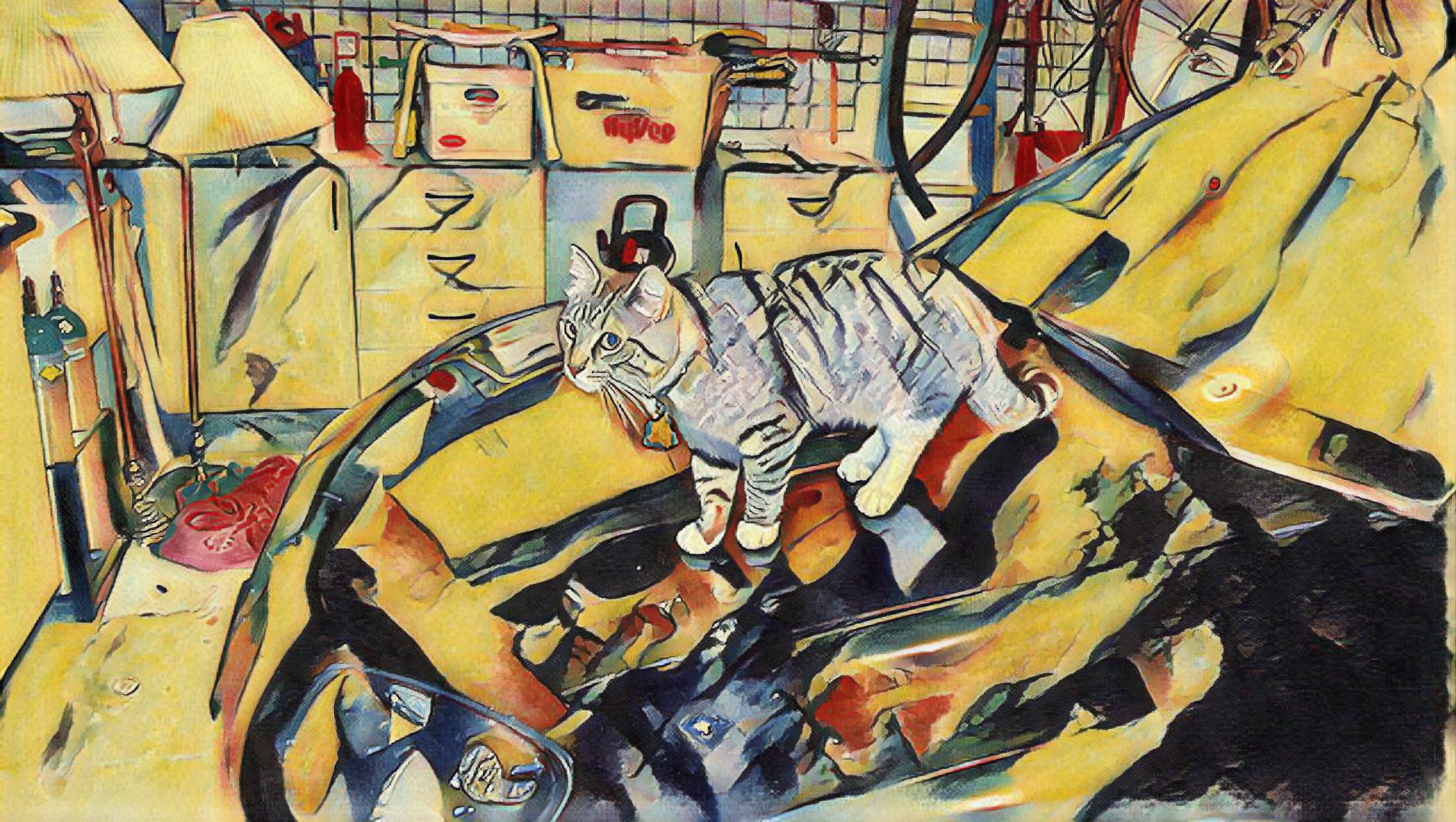}
\caption{High-resolution image generated by our model. The color variations, scale and stroke size are visible. (Kandinsky)}
\label{full8}
\vspace{-0.4cm} 
\end{figure*}

{\noindent\bf High-resolution Image Generation.}
 The proposed DRB-GAN allows us to produce high-quality stylized images in high-resolution. To obtain these synthesized images, the smaller edge of the content images (Figure \ref{full}) is resized to $1280$ and the aspect ratio is kept. The results are presented in Figure \ref{full1} - Figure \ref{full8}. As we can observe, the stylizations exhibit a lot of fine details such as homogeneous regions and smooth color transitions. More importantly, the structure similarity is well preserved. 

\begin{figure*}[ht!]
\centering
\includegraphics[width=1.0\textwidth]{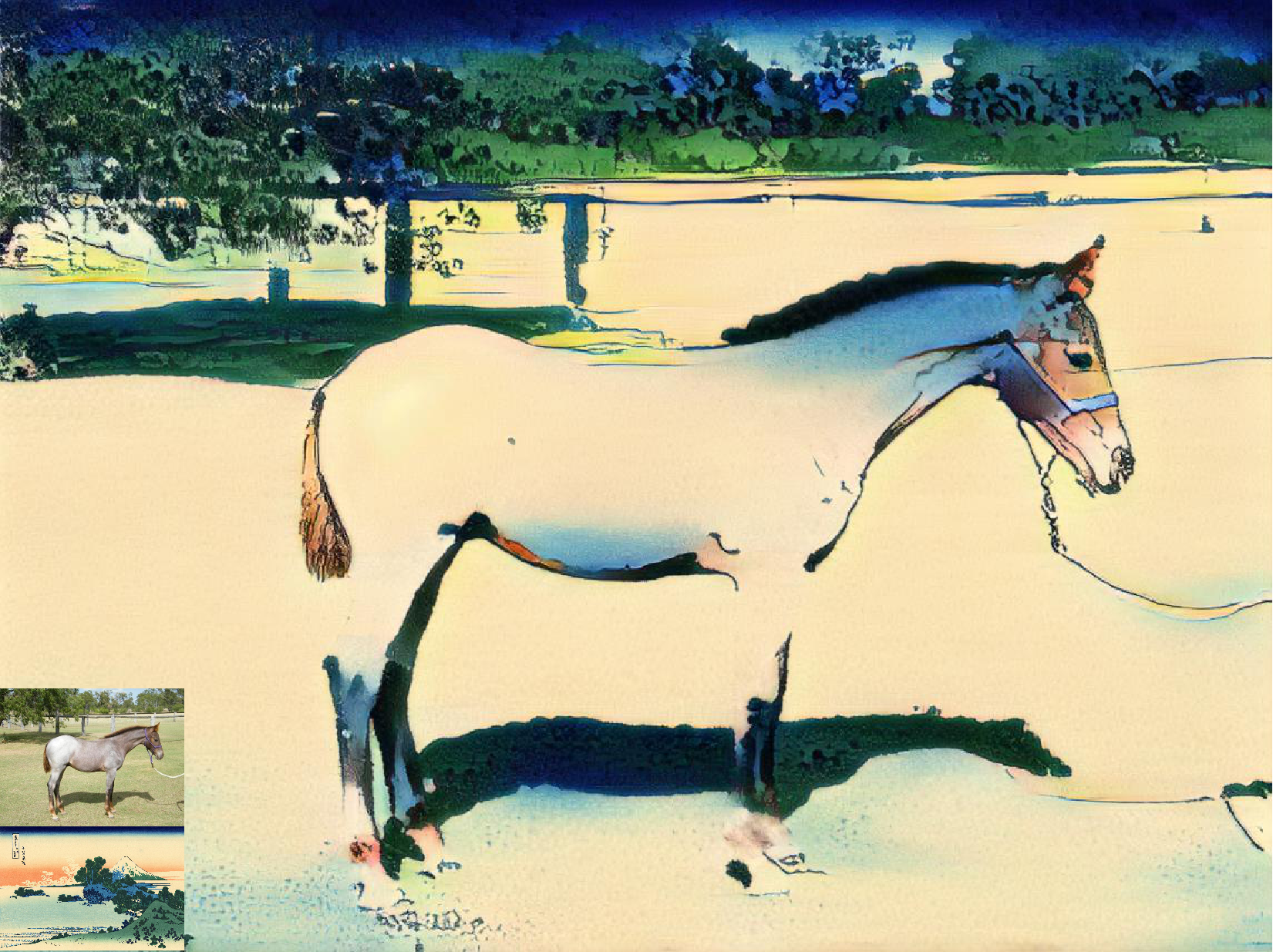}
\caption{An example of stylization in Ukiyo-e style}
\label{fig:Ux}
\vspace{-0.4cm} 
\end{figure*}

\clearpage

\clearpage

\begin{figure*}[t]
\centering
\includegraphics[width=1.0\textwidth]{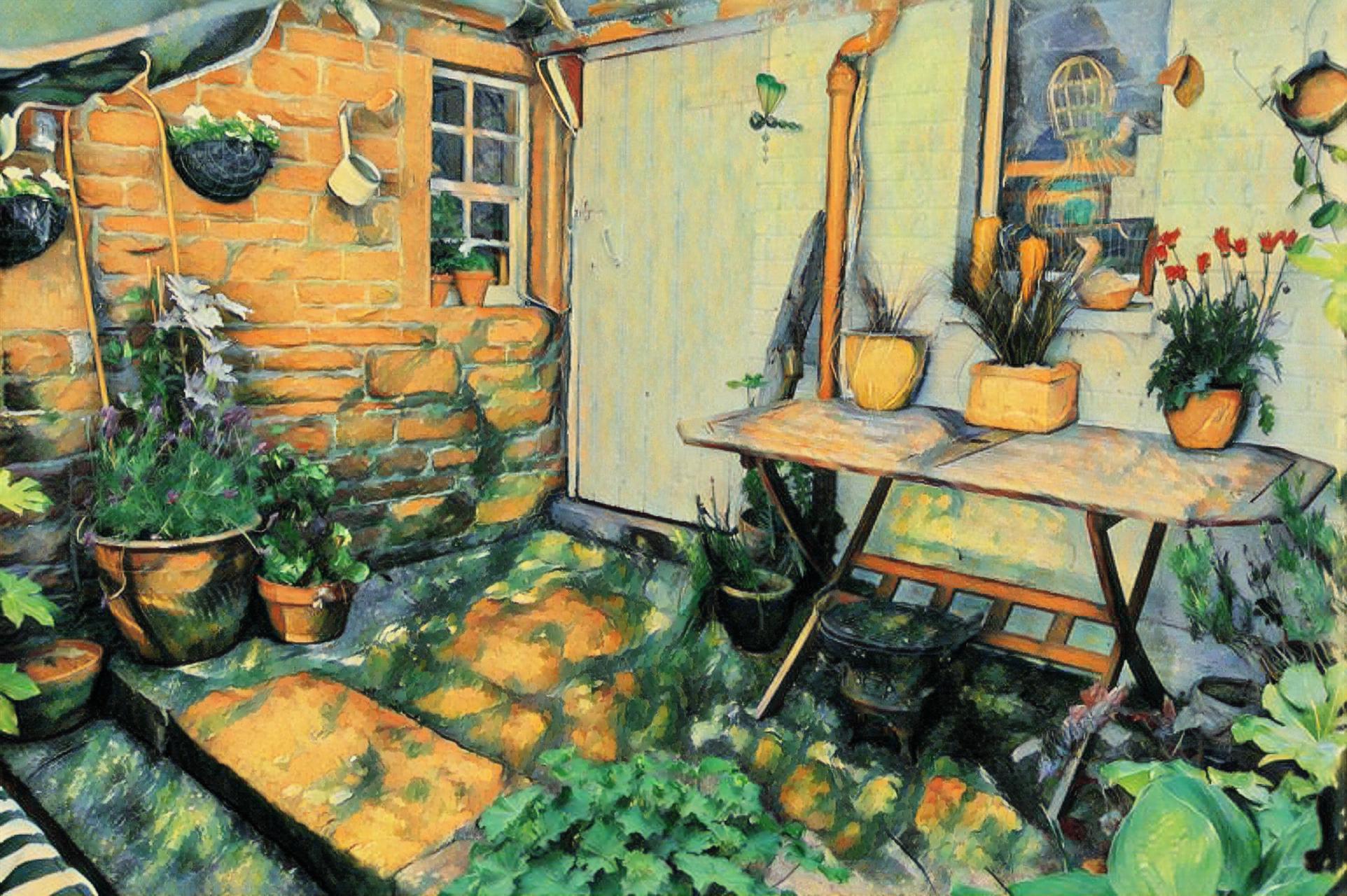}
\caption{High-resolution image generated by our model. The color variations, scale and stroke size are visible. (Cezanne)}
\label{full5}
\vspace{-0.4cm} 
\end{figure*}

\begin{figure*}[t]
\centering
\includegraphics[width=1.0\textwidth]{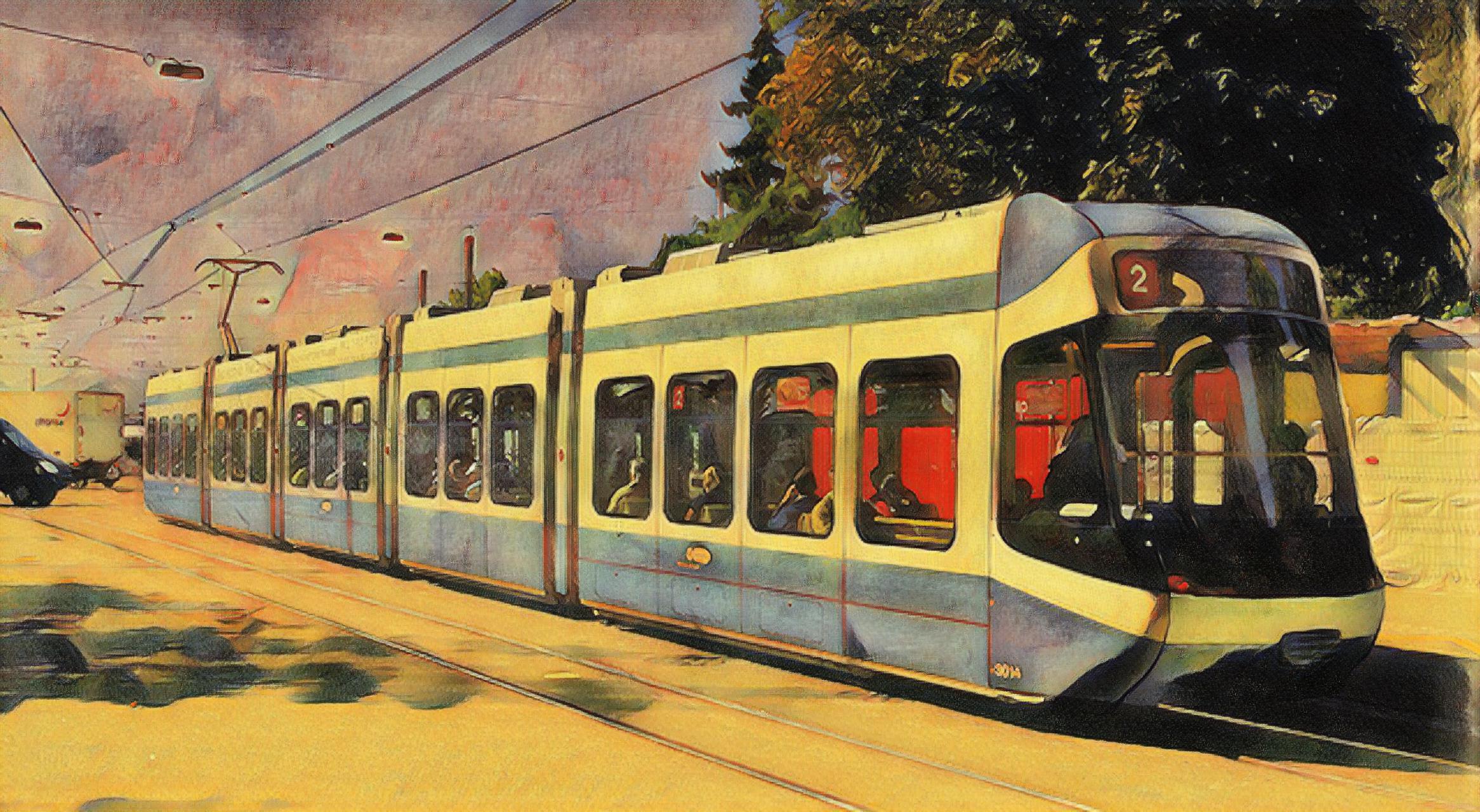}
\caption{High-resolution image generated by our model. The color variations, scale and stroke size are visible. (Gauguin)}
\label{full6}
\vspace{-0.4cm} 
\end{figure*}

\begin{figure*}[t]
\centering
\includegraphics[width=0.6\textwidth]{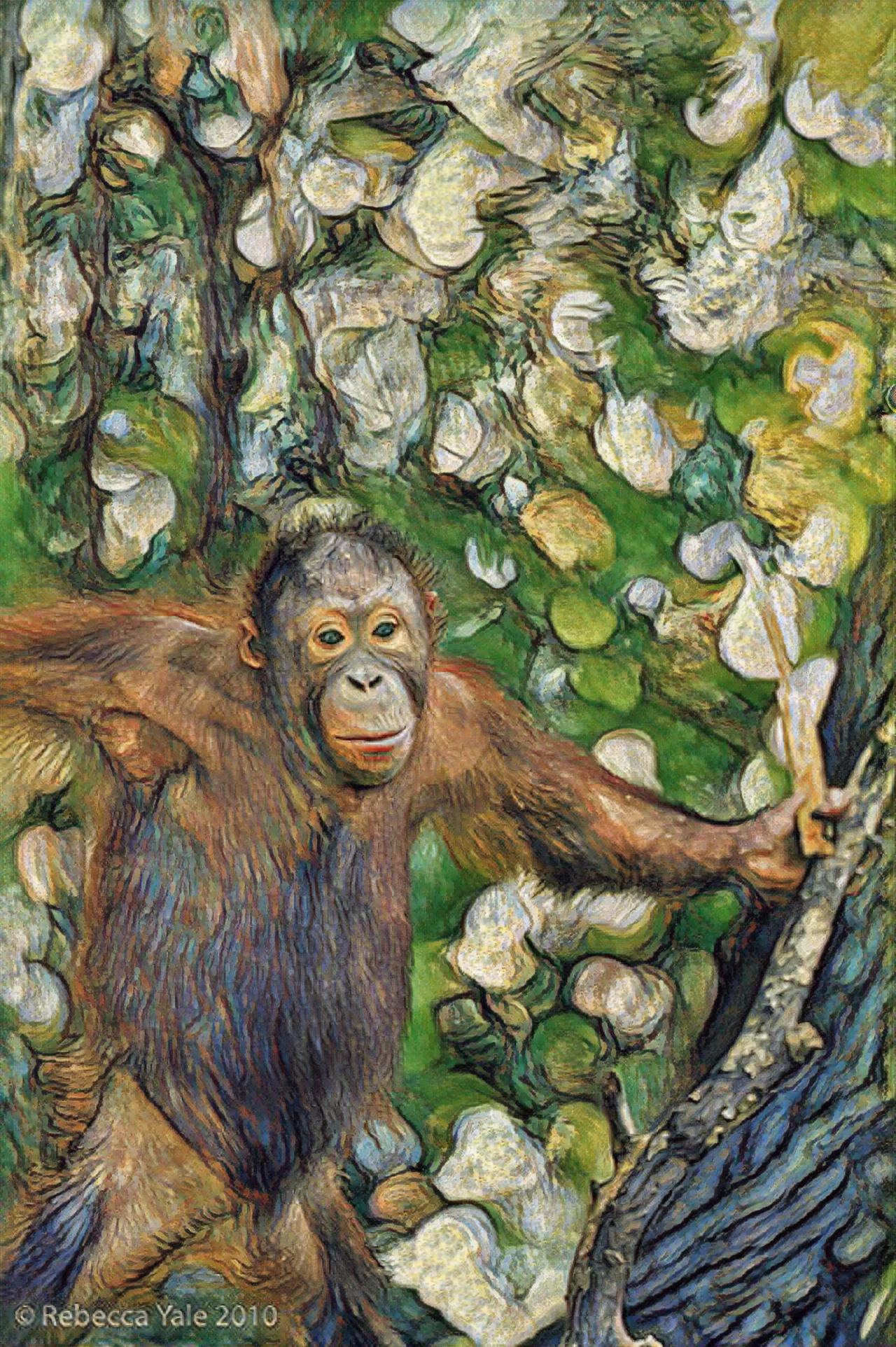}
\caption{High-resolution image generated by our model. The color variations, scale and stroke size are visible. (van Gogh)}
\label{full7}
\vspace{-0.4cm} 
\end{figure*}
\end{document}